\documentclass[letterpaper]{article} % DO NOT CHANGE THIS
\usepackage{aaai2026}  % DO NOT CHANGE THIS
\usepackage{times}  % DO NOT CHANGE THIS
\usepackage{helvet}  % DO NOT CHANGE THIS
\usepackage{courier}  % DO NOT CHANGE THIS
\usepackage[hyphens]{url}  % DO NOT CHANGE THIS
\usepackage{graphicx} % DO NOT CHANGE THIS
\usepackage{natbib}  % DO NOT CHANGE THIS AND DO NOT ADD ANY OPTIONS TO IT
\usepackage{caption} % DO NOT CHANGE THIS AND DO NOT ADD ANY OPTIONS TO IT
\usepackage{booktabs}       % professional-quality tables
\usepackage{amsfonts}       % blackboard math symbols
\usepackage{nicefrac}       % compact symbols for 1/2, etc.
\usepackage{microtype}      % microtypography
\usepackage{xcolor}         % colors
\usepackage{graphicx}
\usepackage{amsmath}
\usepackage{subcaption}
\usepackage{amssymb}

\nocopyright

\usepackage{multirow}
\usepackage{pifont}
\usepackage{tabularray}
\usepackage{algorithm}
\usepackage{algorithmic}

\usepackage{newfloat}
\usepackage{listings}
\DeclareCaptionStyle{ruled}{labelfont=normalfont,labelsep=colon,strut=off} % DO NOT CHANGE THIS
\floatstyle{ruled}
\newfloat{listing}{tb}{lst}{}
\floatname{listing}{Listing}
\DeclareMathOperator*{\argmin}{arg\,min}
\definecolor{iccvblue}{rgb}{0.21,0.49,0.74}
\definecolor{customgreen}{HTML}{AFE1AF}
\definecolor{customred}{HTML}{FFBFBF}

\title{Breaking the Stealth-Potency Trade-off in Clean-Image Backdoors with Generative Trigger Optimization}
\author{
    Binyan Xu\textsuperscript{\rm 1},
    Fan Yang\textsuperscript{\rm 1},
    Di Tang\textsuperscript{\rm 2}\thanks{Corresponding author.},
    Xilin Dai\textsuperscript{\rm 3},
    Kehuan Zhang\textsuperscript{\rm 1}\textsuperscript{\rm *}
}
\affiliations{
    \textsuperscript{\rm 1}The Chinese University of Hong Kong, Hong Kong, China\\
    \textsuperscript{\rm 2}Sun Yat-sen University, Shenzhen, China\\
    \textsuperscript{\rm 3}Zhejiang University, Hangzhou, China\\
    \{binyxu, yf020, khzhang\}@ie.cuhk.edu.hk, tangd9@mail.sysu.edu.cn, xilin2023@zju.edu.cn
}

\newcommand{\methodname}{Generative Adversarial Clean-Image Backdoors}
\newcommand{\methodabb}{GCB}

\usepackage{bibentry}
\begin{document}

\maketitle

% \begin{abstract}
% Deep neural networks are fundamental in security-critical applications such as facial recognition, autonomous driving, and medical diagnostics, yet they are vulnerable to backdoor attacks. 
% Clean-image backdoor attack, a stealthy attack utilizing solely label manipulation to implant backdoors, renders models vulnerable to exploitation by malicious labelers. However, existing clean-image backdoor attacks likely lead to a noticeable drop in Clean Accuracy (CA), decreasing their stealthiness. In this paper, we show that clean-image backdoor attacks can achieve a negligible decrease in CA by poisoning only a few samples while still maintaining a high attack success rate. 
% We introduce \textbf{G}enerative Adversarial \textbf{C}lean-Image \textbf{B}ackdoors (\methodabb{}), a novel attack method that minimizes the drop in CA to less than 1\% by optimizing the trigger pattern for easier learning by the victim model. Using a variant of InfoGAN, we ensure that the trigger pattern we used has already been contained in some training images and can be easily separated from those feature patterns used for benign tasks. 
% Our experiments demonstrate that \methodabb{} can be adapted to 6 datasets, 5 different architectures, and 4 tasks, including classification, multi-label classification, regression, and segmentation. Furthermore, \methodabb{} demonstrates strong stealthiness and robustness to comprehensive backdoor mitigation and defense strategies. Code: \url{https://anonymous.4open.science/r/GCB}.
% \end{abstract}

\begin{abstract}
Clean-image backdoor attacks, which use only label manipulation in training datasets to compromise deep neural networks, pose a significant threat to security-critical applications. Existing methods often require poison rates that are empirically associated with noticeable drops in Clean Accuracy (CA), undermining their stealthiness. This paper presents a new paradigm for clean-image attacks that minimizes this accuracy degradation by optimizing the trigger itself. We introduce \textbf{G}enerative \textbf{C}lean-Image \textbf{B}ackdoors (\methodabb{}), a framework that uses a conditional InfoGAN to identify naturally occurring image features that can serve as potent and stealthy triggers. By learning trigger candidates that are distinguishable from the remaining data while preserving benign task performance empirically, \methodabb{} enables a victim model to learn the backdoor from an extremely small set of poisoned examples, resulting in a CA drop of less than 1\%. Our experiments demonstrate \methodabb{}'s remarkable versatility, successfully adapting to six datasets, five architectures, and four tasks, including the first demonstration of clean-image backdoors in regression and segmentation. \methodabb{} also exhibits resilience against most of the existing backdoor defenses.
\end{abstract}

\begin{links}
    \link{Code}{https://github.com/binyxu/GCB}
    % \link{Extended version}{https://arxiv.org/abs/2511.07210}
\end{links}

% \link{Code}{https://github.com/binyxu/GCB}

% Clean-image backdoor attacks, which use only label manipulation in training datasets to compromise deep neural networks, pose a significant threat to security-critical applications. A critical flaw in existing methods is that the poison rate required for a successful attack induces a proportional, and thus noticeable, drop in Clean Accuracy (CA), undermining their stealthiness. This paper presents a new paradigm for clean-image attacks that minimizes this accuracy degradation by optimizing the trigger itself. We introduce Generative Clean-Image Backdoors (GCB), a framework that uses a conditional InfoGAN to identify naturally occurring image features that can serve as potent and stealthy triggers. By ensuring these triggers are easily separable from benign task-related features, GCB enables a victim model to learn the backdoor from an extremely small set of poisoned examples, resulting in a CA drop of less than 1\%. Our experiments demonstrate GCB's remarkable versatility, successfully adapting to six datasets, five architectures, and four tasks, including the first demonstration of clean-image backdoors in regression and segmentation. GCB also exhibits resilience against most of the existing backdoor defenses. Code: \url{https://anonymous.4open.science/r/GCB}.
    
\section{Introduction}
\label{sec:intro}

% Deep Neural Networks (DNNs) are widely used in applications like facial recognition~\citep{an2023imu}, autonomous driving~\citep{han2022autonomous}, and medical image diagnosis~\citep{li2021medical}; however, backdoor attacks threaten their trustworthiness. By poisoning a small portion of the training data~\citep{li2022backdoor}, adversaries can inject backdoors that cause models to make erroneous predictions when specific inputs are presented. Recent studies reveal that backdoors can be implemented without modifying images, known as \textit{clean-image backdoors}—a significant concern when data annotation is outsourced to third parties. For instance, \cite{chen2022cib} induced a one-to-one backdoor attack in multi-label classification by relabeling images from a source label to a target label, though this method is less adaptable to general image classification tasks. To address this limitation, \cite{jha2024flip} proposed a label-optimization technique that constructs a surrogate poisoned-image backdoor model and optimizes soft labels to mimic its behavior.

Deep Neural Networks (DNNs) are widely used in applications like facial recognition~\citep{an2023imu}, autonomous driving~\citep{han2022autonomous}, and medical image diagnosis~\citep{li2021medical}. However, backdoor attacks threaten their widespread adoption. By poisoning a small fraction of the training data~\citep{li2022backdoor}, an adversary can implant a hidden backdoor, causing the model to generate targeted mispredictions when a specific trigger is present in inputs. A particularly insidious variant is the \textit{clean-image backdoor}, where the attack is executed without any image modification, typically by manipulating labels. This poses a significant threat in scenarios where data annotation is outsourced. For instance, CIB~\citep{chen2022cib} demonstrated a one-to-one backdoor in multi-label classification by simply relabeling all qualified images from a source to a target class. More recently, FLIP~\citep{jha2024flip} proposed a label-optimization technique to mimic the behavior of a surrogate poisoned-image model, extending the attack's applicability.

% \begin{figure}[t]
% \centering
% \includegraphics[width=0.63\linewidth]{figures/head}
% \vskip -0.1in
% \caption{Our method (GCB) breaks the conventional trade-off between attack potency (ASR) and stealth (CA Drop). On CIFAR-10, baselines must sacrifice clean accuracy for higher ASR. In contrast, GCB ($\bigstar$) delivers a highly effective attack ($>97\%$ ASR) with a minimal CA drop ($<1\%$).}
% \label{fig:head}
% \vskip -0.2in
% \end{figure}

\begin{figure}[t]
\centering
\includegraphics[width=0.75\linewidth]{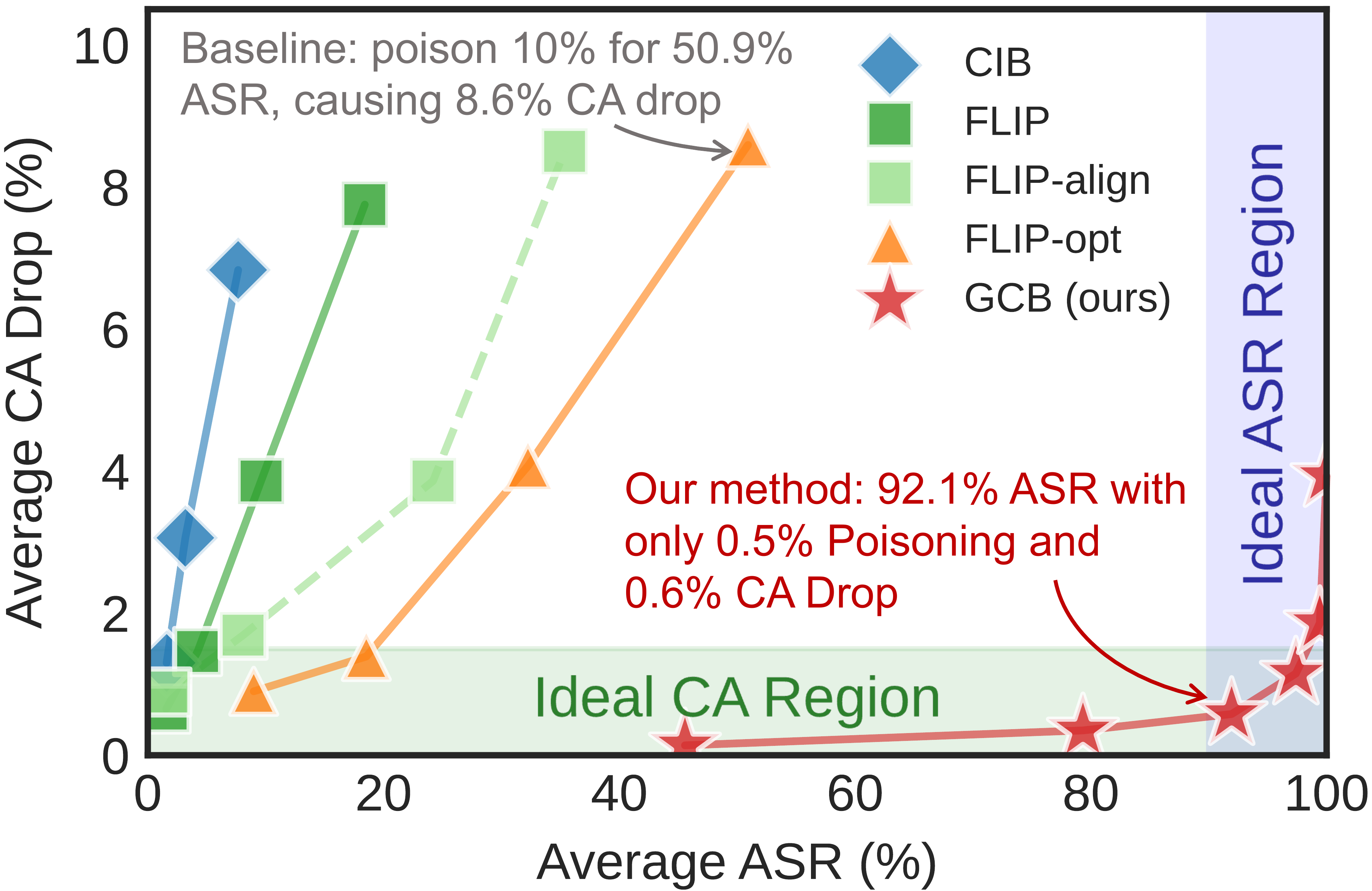}
\vskip -0.08in
\caption{Breaking the Stealth-Potency Trade-off. Average Attack Success Rate (ASR) vs. Clean Accuracy (CA) drop across all datasets. Baselines must sacrifice stealth (CA drop) for attack success. In contrast, our method (GCB, $\bigstar$) delivers a highly effective attack with negligible CA drop.}
\label{fig:head}
\vskip -0.22in
\end{figure}

Although existing methods can achieve high Attack Success Rates (ASR), their stealthiness is fundamentally limited by a clear trade-off between attack potency and model accuracy, as visualized in Fig.~\ref{fig:head}. The figure, which averages performance across six datasets, shows that all existing methods follow a distinct curve: 
to achieve higher ASR, they must accept a significantly higher Clean Accuracy (CA) Drop. 
For example, to exceed 50\% ASR, state-of-the-art methods like FLIP~\citep{jha2024flip} often incur an average CA Drop of over 8\%. This conspicuous degradation compromises stealthiness and undermines its practicality in real-world scenarios where model performance is closely monitored.

% The significant drop in CA can be attributed to a phenomenon known as the natural backdoor trigger in clean-image backdoor, first introduced in~\citep{rong2024ciba}. When a small percentage (e.g., 5\%) of training images are relabeled to train a poisoned victim model, the i.i.d. properties of the training and test datasets result in approximately the same proportion of testing images (around 5\% in this example) being misclassified by the poisoned victim model. This leads to a substantial drop of about 5\% in CA. Unfortunately, this effect is applicable to all types of clean-image backdoors, regardless of the specific attack methods employed. This raises a critical question: \textit{Can we mitigate this effect to create a more stealthy clean-image backdoor?}

Across existing clean-image backdoor methods, higher effective poison rates are empirically associated with larger CA drops, a phenomenon related to what \cite{rong2024ciba} term the "natural backdoor trigger" effect. Because training and test samples are drawn from the same population, features correlated with the relabeled subset may also occur in benign test samples; if the victim model learns this spurious correlation, those samples can be misclassified. The magnitude of the resulting CA drop depends on the learned decision rule and need not equal the poison rate exactly. Nevertheless, the observed association motivates a practical trade-off: reducing the poison rate generally improves stealthiness but can weaken the attack. The central challenge, therefore, becomes: \textit{How can we break this empirical trade-off, designing a trigger so potent that its corresponding backdoor is both highly effective and exceptionally stealthy?}

% Our answer is affirmative. By proposing \methodname{} (\methodabb{}), we can significantly reduce the poison rate to 0.1\%. This leads to substantial mitigation of the CA drop, averaging only 0.2\% across classes and a maximum of 0.5\% for any single class. Our key idea is to lower the poison rate by optimizing the trigger pattern, making it easier for the victim model to learn. However, within the context of clean-image backdoors, optimizing the trigger pattern is challenging because we cannot modify images; instead, we must utilize features that already exist on benign images to construct triggers. In this context, three main constraints arise: \textbf{(1) Existence}: The optimized trigger pattern must be present in the training set. \textbf{(2) Separability}: Images with and without the trigger must be easily distinguishable, enabling the victim model to learn the backdoor more effectively and low poison rate. \textbf{(3) Irrelevancy}: The trigger should not interfere with learning the benign task, preventing a significant CA drop.

This paper addresses this challenge head-on. We introduce \methodname{} (\methodabb{}), a novel framework that creates such highly effective triggers, enabling successful attacks with a poison rate as low as 0.1\%. This, in turn, reduces the average CA drop to a mere 0.2\% and a maximum of 0.5\% for any class. However, this optimization is non-trivial, as we must identify features that already exist within benign data, subject to 3 constraints:
\textbf{(1) Existence}: The optimized trigger pattern must naturally exist in the training set.
\textbf{(2) Separability}: Images with and without the trigger must be easily distinguishable for the model to learn the backdoor from a very low poison rate.
\textbf{(3) Irrelevancy}: The trigger transformation should preserve label-relevant information and minimize interference with the benign task.

% In this paper, we develop a novel GAN framework, C-InfoGAN, to optimize triggers while addressing three key issues. (a) To ensure \textit{existence}, we employ a GAN generator to construct a trigger function, ensuring that all generated images, including trigger images, belong to the original image distribution. 
% (b) To achieve \textit{separability}, we adopt the concept of InfoGAN, building the GAN generator in a two-fold manner (representing triggered and benign images respectively) and maximizing their distance as a term in the loss function.
% (c) To guarantee \textit{Irrelevancy}, we incorporate the ground truth label as a prior for all components to ensure that the trigger features are irrelevant to class features. 

To address these constraints jointly, we develop C-InfoGAN, a novel conditional generative framework. C-InfoGAN is designed to find a candidate trigger function by using a GAN architecture in a new way:
(a) For \textbf{Existence}, we employ an adversarial discriminator to encourage the generator's output to remain on the natural data manifold, so that the identified trigger pattern resembles naturally occurring variation.
(b) For \textbf{Separability}, we build on InfoGAN by training a generator with two latent codes (representing triggered and benign states) and maximizing conditional mutual information to encourage distinguishability between the generated branches.
(c) For \textbf{Irrelevancy}, we condition all components of the framework on the ground-truth class labels, encouraging the framework to represent the trigger through within-class variation rather than class-discriminative features; this property is evaluated empirically.

\newcommand{\solidcircle}{\ding{51}}  % Defines solid circle
\newcommand{\hollowcircle}{\ding{55}}   % Defines hollow circle

\renewcommand{\arraystretch}{0.98}
\begin{table}[t]
\centering
\begin{small}
\setlength{\tabcolsep}{3pt}
\begin{tabular}{@{}lcccc@{}}
\toprule
Property↓ & CIB  & FLIP  & CIBA  & \methodabb{} (ours) \\
\midrule
CA Drop $\leq 1\%$           & \hollowcircle & \hollowcircle & \hollowcircle & \solidcircle \\
Poison Rate $\leq 1\%$       & \hollowcircle & \hollowcircle & \hollowcircle & \solidcircle \\
ASR $\geq 90\%$              & \solidcircle  & \solidcircle  & \hollowcircle & \solidcircle \\
Scalability (Datasets)                  & \solidcircle  & \hollowcircle & \hollowcircle & \solidcircle \\
Transferability (Architectures)               & \solidcircle  & \hollowcircle & \solidcircle  & \solidcircle \\
Generalizability (Tasks)              & \hollowcircle & \hollowcircle  & \hollowcircle  & \solidcircle \\
\bottomrule
\end{tabular}
\end{small}
\vspace{-0.2cm}
\caption{Comparison between SOTA clean-image backdoors.}
\label{tab:edit_and_view}
\vspace{-0.2in}
\end{table}
\renewcommand{\arraystretch}{1.0}

Our extensive experiments validate the superior stealth and effectiveness of \methodabb{}. Across 6 datasets including MNIST, CIFAR-10, CIFAR-100, GTSRB, Tiny-ImageNet, and ImageNet, \methodabb{} achieves ASRs up to 100\% (e.g., 97.9\% on CIFAR-10) with less than a 1\% CA drop, using a source-pool poison rate of only 0.5\%. The method is robust across architectures (ResNet, VGG, ViT) and shows remarkable generalizability, extending for the first time to complex vision tasks like multi-label classification, regression, and segmentation. Even in a challenging scenario where the adversary accesses only 10\% of the training data, \methodabb{} achieves a 90.3\% ASR with a 0.15\% CA drop on CIFAR-10. Furthermore, \methodabb{} demonstrates resilience against most of existing SOTA backdoor defenses. A summary comparison is provided in Table~\ref{tab:edit_and_view}.

Our contributions are three-fold:
\begin{itemize}
    \item \textbf{Breaking the Stealth-Potency Trade-off}: We are the first to demonstrate a clean-image backdoor that is simultaneously highly potent ($\geq 90\%$ ASR) and exceptionally stealthy (negligible CA drop $\leq 1\%$ with $\leq 0.5\%$ poison rate) on all datasets, effectively breaking the conventional trade-off that plagues existing methods.
    
    \item \textbf{Broad Applicability and Generalization}: Our method demonstrates exceptional adaptability across 6 datasets, 5 architectures, and 4 tasks. Crucially, it is the first clean-image attack framework shown to be effective for regression and segmentation tasks, dramatically expanding the threat landscape.

    \item \textbf{Novel Attack Method}: We introduce a novel attack methodology based on a conditional InfoGAN, which uniquely reframes the generator as a trigger function and a recognizer as a score function to solve the complex co-optimization problem inherent to creating separable, existing, and irrelevant clean-image backdoor triggers.

\end{itemize}

\section{Related Work}

\subsection{Data Poisoning Backdoor}
Backdoor attacks have evolved from using conspicuous triggers~\citep{gu2019badnets} to more stealthy methods employing blended images~\citep{chen2017blended}, natural reflections~\citep{liu2020reflection}, and clean-label perturbations~\citep{turner2019label}, alongside transferable targeted attacks in CLIP-based settings~\citep{xu2025one}. Our work GCB, is a clean-image backdoor attack, meaning it does not alter images in training datasets.

However, prior clean-image methods face critical limitations. CIB~\citep{chen2022cib} is designed for multi-label classification and does not generalize to standard classification tasks. FLIP~\citep{jha2024flip} requires impractical knowledge of the victim model’s architecture and fails to scale beyond simple datasets. CIBA~\citep{rong2024ciba} is ineffective, achieving less than 50\% Attack Success Rate (ASR) even on CIFAR-10. GCB is designed to overcome these shortcomings. VLM-driven online auditing provides a complementary test-time defense direction~\citep{xu2026internal}.

\subsection{GAN-Based Representation Learning}
To enhance our backdoor's efficiency, GCB utilizes a novel GAN architecture. Research in Generative Adversarial Networks (GANs) has progressed from foundational models~\citep{goodfellow2014generative} toward controllable representation learning with cGAN~\citep{mirza2014conditional}, InfoGAN~\citep{chen2016infogan}, and StyleGAN~\citep{karras2019style}. While these models excel at manipulating generated images, editing real images often requires complex GAN Inversion techniques~\citep{xia2022inversion}, which add significant overhead. In this paper, we propose C-InfoGAN, a new architecture that integrates interpretable feature editing directly into the GAN framework.

% 大体这里是三步走的思路，1. split poison dataset 2. uniform label assignation 3. trigger backdoor. 
% 在这里只考虑uniform label assignation为了保证target label的任意性
% 在Preliminary先把问题定义为确定poison split Z_1和寻找T(x)的问题，在这里把uniform label assignation讲清楚。即为\( Z_1 = \{(x, y_{\text{target}}) : x \in X_1\} \)。接下来剩下的问题就是1. select poison feature 2. trigger function了。

\begin{figure*}[t]
\begin{center}
\centerline{\includegraphics[width=5.6in]{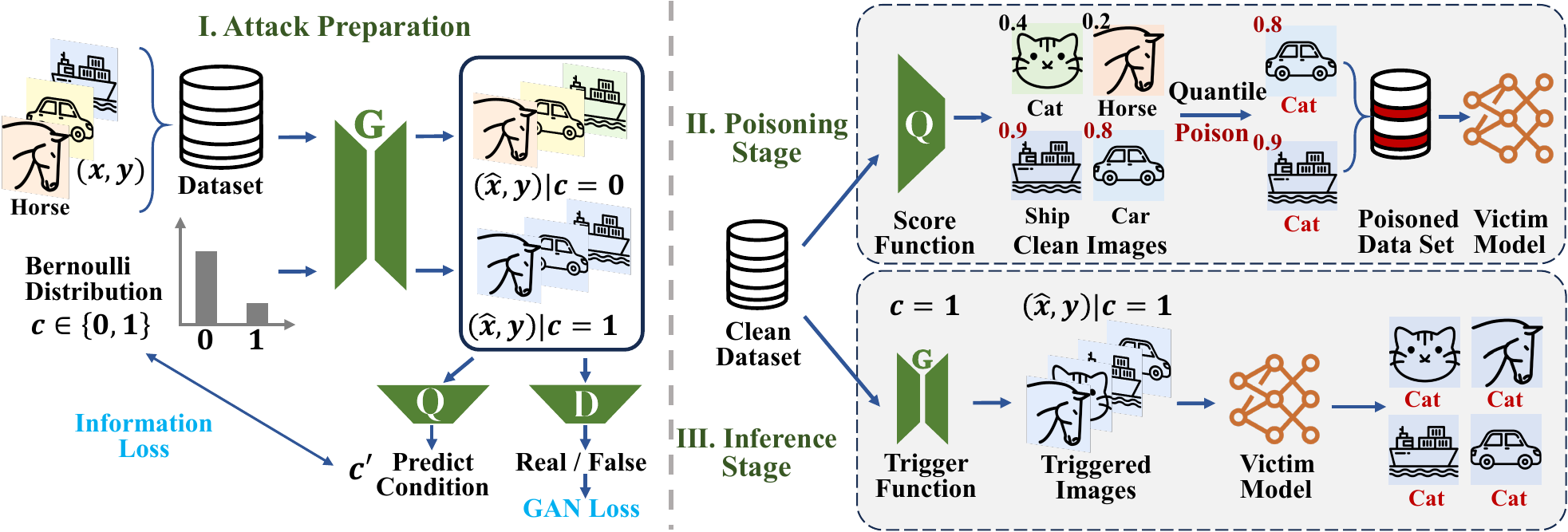}}
\vskip -0.05in
\caption{Framework of \methodname{} (\methodabb{}). In the preparation stage, a specific clean feature (e.g., background color here) is extracted as a backdoor trigger.}
\label{fig:framework}
\end{center}
\vskip -0.35in
\end{figure*}

\section{Preliminary}

\subsection{Threat Model}

We adopt the same threat model as other clean-image backdoors \cite{jha2024flip, chen2022cib}: investigating the risks posed by malicious third-party annotators in the context of externally annotated datasets. Specially, attackers have partial or full access to view the training dataset, but their malicious actions are limited to subtly \textbf{mislabeling} a small portion of the dataset, \textbf{without the ability to modify any images in the training dataset} or influence other training aspects like the architecture or training strategy. 
% Note that this threat model is fundamentally different from invisible backdoors \cite{li2020invisible, li2021ssba}, where attackers \textbf{can} modify training images within a small perturbation range.

\subsection{Notation}

In this study, we consider a supervised model \(f\), with input \(x\) and label \(y\). For an all-to-one attack with target label \(y_t\), define the source pool
\[
X_s=\{x_i:(x_i,y_i)\in(X,Y),\ y_i\neq y_t\}.
\]
The attacker selects a malicious subset \(X_1\subseteq X_s\) and changes its labels to \(y_t\), forming \((X_1,Y_1)=\{(x,y_t):x\in X_1\}\). The remaining samples \(X_0=X\setminus X_1\), including all original target-class samples, retain their benign labels \(Y_0\). The poisoned dataset is \((X,Y')=(X_0,Y_0)\cup(X_1,Y_1)\). We define the poison rate relative to the source pool, up to integer rounding, as
\[
|X_1|=p_r|X_s|.
\]
% The attacker's
% goal is to make the victim model learn the following two tasks simultaneously:
% \begin{equation}
% \label{eq:backdoor_task}
% f_{\theta}^* = \arg\min_{\theta}  \underbrace{\mathbb{E}_{(x_0, y_0) \sim (X_0, Y_0)} \left[ \ell(f_{\theta}(x_0), y_0) \right]}_{\text{classification task}} + \underbrace{\mathbb{E}_{(x_1, y_t) \sim (X_1, Y_1)} \left[ \ell(f_{\theta}(x_1), y_t) \right]}_{\text{backdoor task}}
% \end{equation}
During testing, our label-conditioned trigger function converts an input-label pair \((x,y)\) into a triggered input \(\hat{x}=T(x,y)\), with the goal that \(f^*_{\theta}(T(x,y))=y_t\). Our experiments evaluate this label-aware inference setting: the attacker supplies the source label \(y\) used by the generator. Using an estimated label instead is a possible extension, but its additional error is outside the scope of the present analysis.

\section{Methodology}

\subsection{Overview}

GCB aims to minimize the CA drop while maintaining a high ASR for clean-image backdoors. In these scenarios, a portion of training images are deliberately mislabeled, but the images themselves remain unchanged. We introduce C-InfoGAN to learn and score candidate variations for selecting images to mislabel; their transfer to real samples and preservation of benign label information are evaluated empirically. The GCB framework is illustrated in Fig.~\ref{fig:framework} and comprises attack preparation, poisoning, and inference. During preparation, $Q$ learns to distinguish the generated conditional branches and is then used to score labeled training images. In the poisoning stage, the highest-scoring source-class images are relabeled. In the inference stage, the label-conditioned generator $G$ converts a source input into a triggered input intended to be predicted as $y_t$.

\subsection{C-InfoGAN}

\begin{figure*}[t]
\begin{center}
\centerline{\hspace*{0.07\textwidth}\includegraphics[width=0.88\textwidth]{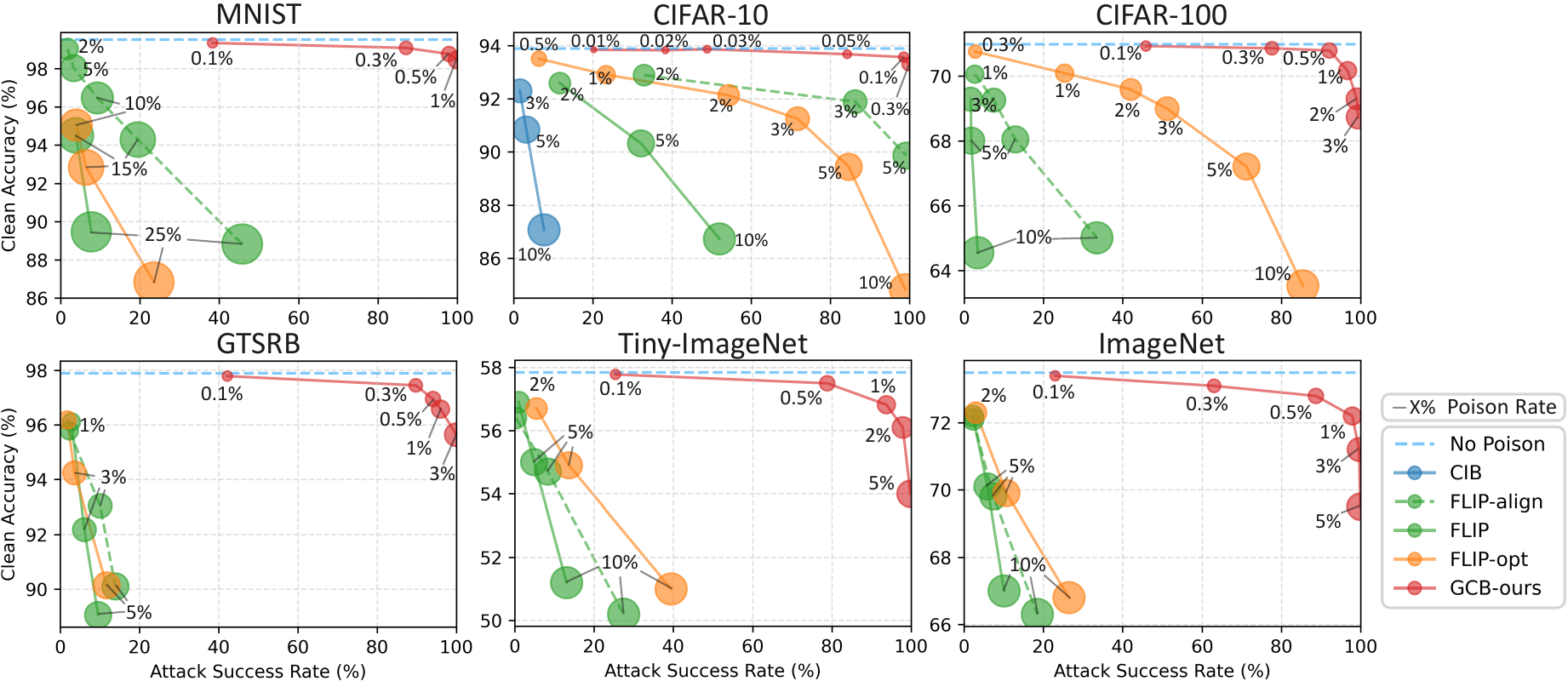}}
\vskip -0.05in
\caption{Stealth-potency trade-off of clean-image backdoor methods across datasets. Marker size and text indicate poison rates on each point. Our method, GCB, achieves $\geq90\%$ attack success with $\leq1\%$ drop in clean accuracy.}
\label{fig:main_acc_asr}
\end{center}
\vskip -0.25in
\end{figure*}

Essentially, given a fixed poison rate (limiting the number of mislabeled images), our goal is to maximize both ASR and CA. However, it is a challenge in clean-image backdoor settings, as we can only modify the labels of images, leading to a discrete hard-label issue. Even advanced discrete optimization methods like GCG can only maximize ASR but struggle to maintain a high CA. 

Our observations lead us to model this problem as a divergence maximization problem constrained by three factors: 
(a) Existence: The trigger pattern must be present within the training data, enabling backdoor injection via label manipulation alone.
(b) Separability: The images with and without the trigger must be distinctly separable, allowing easier backdoor learning and reducing the required poison rate.
(c) Irrelevancy: The trigger should not interfere with benign class features to prevent a significant CA drop, as feature overlap can disrupt class semantics. To satisfy these constraints, we introduce Conditional Information Maximizing GANs (C-InfoGAN). In C-InfoGAN, we introduce a discrete random variable \(c\) following a Bernoulli distribution as the latent variable. The generator \(G\), conditioned on \(c\), generates two distinct series of images depending on whether \(c\) is 0 or 1.

\textbf{(a) Existence.} A crucial property of clean-image backdoors is that the trigger pattern should resemble variation present in the clean-image distribution. We therefore employ a standard GAN framework~\citep{goodfellow2014generative}. At the ideal GAN equilibrium, the generated marginal matches the real-data distribution. Because each positive-weight conditional component is then absolutely continuous with respect to this marginal, the generated $c=1$ branch does not place probability mass outside the support of the real-data distribution. This is a population-level support statement, not a guarantee that each generated image occurs in the finite training set.

\textbf{(b) Separability.} To encourage separability, we follow the concept of InfoGAN~\citep{chen2016infogan}, which maximizes the mutual information between selected latent variables and the generated data to learn interpretable and disentangled representations. 
% Specifically, for optimizing separability, we use a discrete random variable $c$ following a Bernoulli distribution as the latent variable. The generator $G$, conditioned on $c$, generates two distinct series of images depending on whether $c$ is 0 or 1. 
The recognition network $Q$ (originating from InfoGAN) is tasked with distinguishing between images generated with $c=1$ and $c=0$ as accurately as possible through an information loss term $L_\text{info}$. High code-prediction accuracy by $Q$ indicates that the two generated branches are separable for the recognition model.

\textbf{(c) Irrelevancy.} Another crucial attribute of backdoors is that the trigger should not interfere with the benign task. This indicates that the trigger pattern needs to be irrelevant to the patterns utilized for the benign task. 
To encourage this behavior, we sample $c$ independently of the ground-truth label $y$ and provide $y$ as an auxiliary condition to the C-InfoGAN components. This design encourages $c$ to capture within-class variation. Independence of the sampled variables does not by itself guarantee that the learned trigger is semantically irrelevant, so we evaluate classification consistency and clean accuracy empirically.

\textbf{Objective Function.}
Let $(x,y)\sim P_{X,Y}$, independently sample $c\sim P_C$, and write $\hat{x}=G(x,c,y)$. An idealized minimax form of our objective is
\begin{align}
\min_{G,Q}\max_D\quad
V(G,D,Q)
= {} & \mathbb{E}_{x,y}[\log D(x,y)] \nonumber\\
& +\mathbb{E}_{x,y,c}[\log(1-D(\hat{x},y))] \nonumber\\
& -\lambda\mathbb{E}_{x,y,c}[\log Q(c\mid\hat{x},y)],
\end{align}
where the constant $-\lambda H(C)$ in the mutual-information lower bound is omitted. Here, $D$ is maximized, whereas $G$ and $Q$ are minimized. The recognition loss is defined as
\[
L_{\mathrm{info}}=-\mathbb{E}_{x,y,c}[\log Q(c\mid G(x,c,y),y)].
\]
The first two terms are the conditional vanilla GAN objective~\citep{goodfellow2014generative}; the last term is the InfoGAN variational objective~\citep{chen2016infogan}. The hyperparameter $\lambda$ controls their trade-off.

\textbf{Theoretical Analysis.}
In the Appendix, we give an idealized analysis of the connection between C-InfoGAN and the clean-image backdoor objective. Because $C$ is sampled independently of $Y$, the conditional mutual information $I(C;\hat{X}\mid Y)$ is the expected weighted Jensen--Shannon divergence between the two generated conditional distributions within each label condition. Maximizing its variational lower bound encourages the generated branches to be distinguishable by $Q$. We then provide a direct attack-success bound on the non-target source distribution: if the victim model's target-label error on the selected real distribution is \(\delta\), and the total variation distance between the triggered and selected distributions is \(\epsilon\), then \(\mathrm{ASR}\geq 1-\delta-\epsilon\). Exact component alignment follows only under explicitly stated ideal conditions; in practice it is approximate and evaluated empirically.

% \textbf{Theory Provement.}
% We also provide a theoretical analysis for our GCB attack in Appendix\ref{ap:math}.
% From the perspective of information theory, we show that minimizing information loss can maximize the JS-Divergence between \( G(\cdot, c=0) \) and \( G(\cdot, c=1) \) (Lemma \ref{ap:lemma_1}). In addition, GAN ensures that the generated images $G(x, \cdot)$ and the corresponding real images $x$ share the same distribution. Consequently, the JS-Divergence between the real images of the two series is maximized, allowing the $Q$ component to distinguish them effectively. Both parts mentioned above act as a special case of InfoGAN~\citep{chen2016infogan} and can be easily proved.
% Moreover, we prove that the total conditional entropy of the backdoor task: \( H(Y'|X) \) (where \( (X, Y') \) is the poisoned dataset), is minimized when the divergence between real images of the two series is maximized (Proposition \ref{ap:proposition_2}). Such minimizing indicates that our backdoor task is easily learned, thereby ensuring a high ASR.

\subsection{Attack Deployment}

\textbf{Poisoning Stage.} We consider the source pool \(X_s=\{x_i:y_i\neq y_t\}\), select a subset \(X_1\subseteq X_s\), and change its labels to \(y_t\). The recognition network \(Q\) supplies the score $s(x,y)=Q(c=1\mid x,y)-Q(c=0\mid x,y)$ and is trained to recognize \(c\) in generated images \(\hat{x}\). Conditional adversarial matching motivates transferring this score to real images, while the quality of that transfer is approximate and assessed empirically. We rank only source-pool samples and select the top \(p_r\) fraction, so \(|X_1|=p_r|X_s|\) up to integer rounding. These images have their labels flipped and are then submitted to train the victim model; original target-class samples are not counted as poisoned.

\begin{figure*}[t]
% \vspace{-0.2cm}
\centering
\begin{minipage}{0.32\textwidth}
\centering
\includegraphics[width=\linewidth]{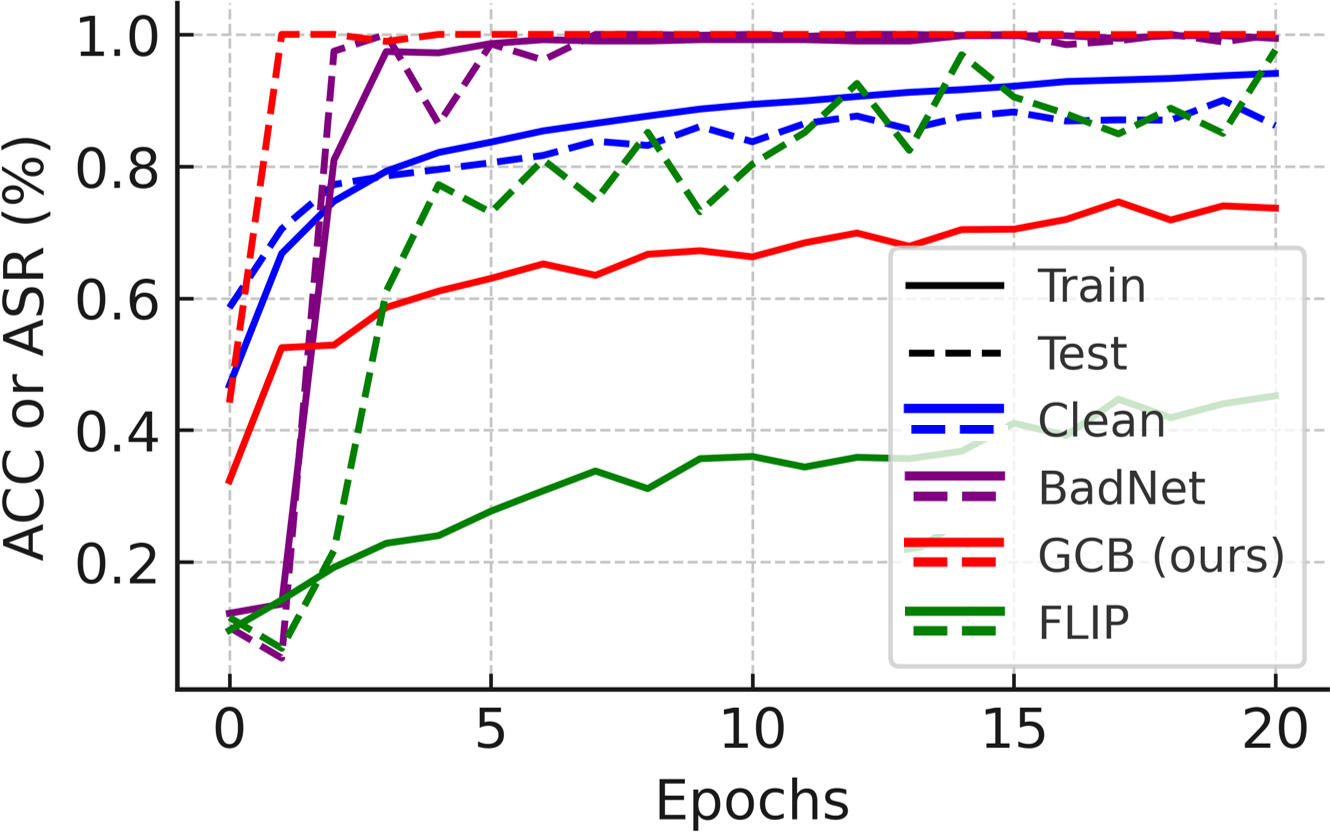}
\vskip -0.13in
\caption{GCB's test ASR on CIFAR-10 converges fast, but its training ASR lags, resisting fast-learning defenses like ABL.}
\label{fig:learn_curve}
\end{minipage}
\hfill
\begin{minipage}{0.32\textwidth}
\centering
\begin{small}
\setlength{\tabcolsep}{2pt}
\begin{tabular}{lccc}
\toprule
Method & Metric & VOC07 & VOC12 \\
\midrule
& ASR ↑ & 87.5±14.2 & 85.2±13.0 \\
CIB & MAP ↑ & 91.8±1.1 & 91.3±1.4 \\
& MAP (src) ↑ & 74.8±3.1 & 72.6±4.9 \\
\midrule
& ASR ↑ & 67.5±7.2 & 70.1±8.5 \\
\methodabb{} & MAP ↑ & 93.9±0.3 & 93.7±0.4 \\
& MAP (src) ↑ & 93.5±0.3 & 93.4±0.3 \\
\bottomrule
\end{tabular}
\end{small}
\vskip -0.06in
\captionof{table}{Results on multi-label classification. ``src'' denotes source class. \methodabb{} succeeds with almost no drop in MAP.}
\label{tab:with_cib}
\end{minipage}
\hfill
\begin{minipage}{0.32\textwidth}
\centering
\begin{small}
\setlength{\tabcolsep}{2pt}
\begin{tabular}{lcccccc}
\toprule
Task & \multicolumn{2}{c}{Regression} & \multicolumn{2}{c}{Segmentation} \\
Dataset & \multicolumn{2}{c}{ColorCIFAR10} & \multicolumn{2}{c}{VOC2012} \\
\cmidrule(lr){1-1} \cmidrule(lr){2-3} \cmidrule(lr){4-5}
Metrics & AE ↓ & CE ↓ & AE ↓ & CE ↓ \\
\midrule
Clean & 0.2964 & 0.0128 & 1.207 & 0.211 \\
1\% Poison & 0.0290 & 0.0141 & 0.303 & 0.214 \\
3\% Poison & 0.0204 & 0.0156 & 0.277 & 0.217 \\
\bottomrule
\end{tabular}
\end{small}
\captionof{table}{Performance of \methodabb{} on other vision tasks. AE: Attack Mean Square Error. CE: Clean Mean Square Error.}
\label{tab:universal}
\end{minipage}
\vspace{-0.2cm}
\end{figure*}

\textbf{Inference Stage.} During the inference stage, the attacker supplies an image $x$ and its source label $y$ to the generator and sets $c=1$, producing the label-conditioned trigger $T(x,y)=G(x,c=1,y)$. The generated branch is designed to approximate the distribution represented by the selected images in the poisoning stage. When this distributional mismatch is small and the victim model has learned the selected samples as the target class, the triggered input activates the backdoor with high probability, as formalized by the bound in the Appendix. Our experiments use the ground-truth source label; inference with an estimated label is not evaluated here.

\section{Evaluation}

\subsection{Experimental Setup}

\textbf{Datasets and Models.} We use BackdoorBench~\citep{wu2024backdoorbench} to evaluate on six datasets: MNIST~\citep{lecun1998gradient}, CIFAR-10/100~\citep{krizhevsky2009learning}, GTSRB~\citep{stallkamp2012man}, Tiny-ImageNet~\citep{le2015tiny}, and ImageNet-1K \cite{deng2009imagenet}. We employ PreActResNet18 as the default victim model with a poison rate of 1\%. All results follow an all-to-one attack scenario. Detailed training settings for C-InfoGAN are provided in Appendix~\ref{ap:set}.

\textbf{Baselines.} Our clean-image backdoor baselines include CIB~\citep{chen2022cib}, FLIP~\citep{jha2024flip}, CIBA~\citep{rong2024ciba}, and FLIP-opt. CIBA shows low ASR and does not release its code, so it is only analyzed in Appendix\ref{ap:poison_rate}. FLIP-opt combines FLIP and Narcissus~\citep{zeng2023narcissus} for trigger optimization. Specifically, we first generate an optimized trigger using Narcissus, then determine the best label assignments for poisoning using FLIP. Additionally, we found that FLIP is highly sensitive to the victim model's architecture, relying on alignment between the victim and surrogate models used in attack preparation. A detailed analysis of this effect is in Appendix~\ref{tab:flip_arc}. To ensure a fair comparison, we report FLIP results under both aligned and unaligned conditions, labeled as FLIP-align and FLIP.

\textbf{Metrics.} We use two metrics in our experiments: \textit{Clean Accuracy} (CA) and \textit{Attack Success Rate} (ASR). CA measures the victim model's accuracy on the full clean test set. In the all-to-one setting, ASR is the fraction of triggered test instances whose original labels are not \(y_t\) that are classified as the target class.

\subsection{Attack Performance.}

\label{exp:effectiveness}

\textbf{ASR VS. CA.} We compare \methodabb{} with several clean-image backdoor baselines in Fig. \ref{fig:main_acc_asr}. Our experiments demonstrate that \methodabb{} significantly outperforms all baselines across all datasets. With less than a 0.5\% drop in CA, \methodabb{} achieves over 90\% ASR on small datasets such as MNIST, CIFAR-10, and CIFAR-100. For more complex datasets like GTSRB and Tiny-ImageNet, \methodabb{} maintains over 90\% ASR with a CA drop within 1\%. In contrast, all tested baselines only succeed on simple datasets like CIFAR-10 and CIFAR-100, incurring CA drops exceeding 5\%. Moreover, they fail on relatively complex datasets such as GTSRB and Tiny-ImageNet, and surprisingly even on the simple MNIST dataset. This failure on MNIST is likely because MNIST consists of grayscale, feature-poor images. Consequently, intuitively selected triggers (e.g., sinusoidal triggers) cannot be effectively constructed using clean image combinations.

\textbf{Convergence Speed and Asymmetric Trigger.} Our key idea is to make the trigger easier for the victim model to learn by optimizing separability. An important question is how quickly the victim model can learn this trigger. Fig. \ref{fig:learn_curve} shows that our method converges to nearly 100\% ASR in just 4 epochs, whereas the simplest backdoor attack, BadNets, requires 11 epochs to converge. This indicates that our backdoor task is even easier for neural networks than BadNets. Compared to peer clean-image backdoor methods, FLIP takes over 20 epochs to achieve a successful attack and remains unstable after 20 epochs.

\begin{figure}[t]
  \centering
  \begin{minipage}{0.48\linewidth}
    \centering
    \includegraphics[width=\linewidth]{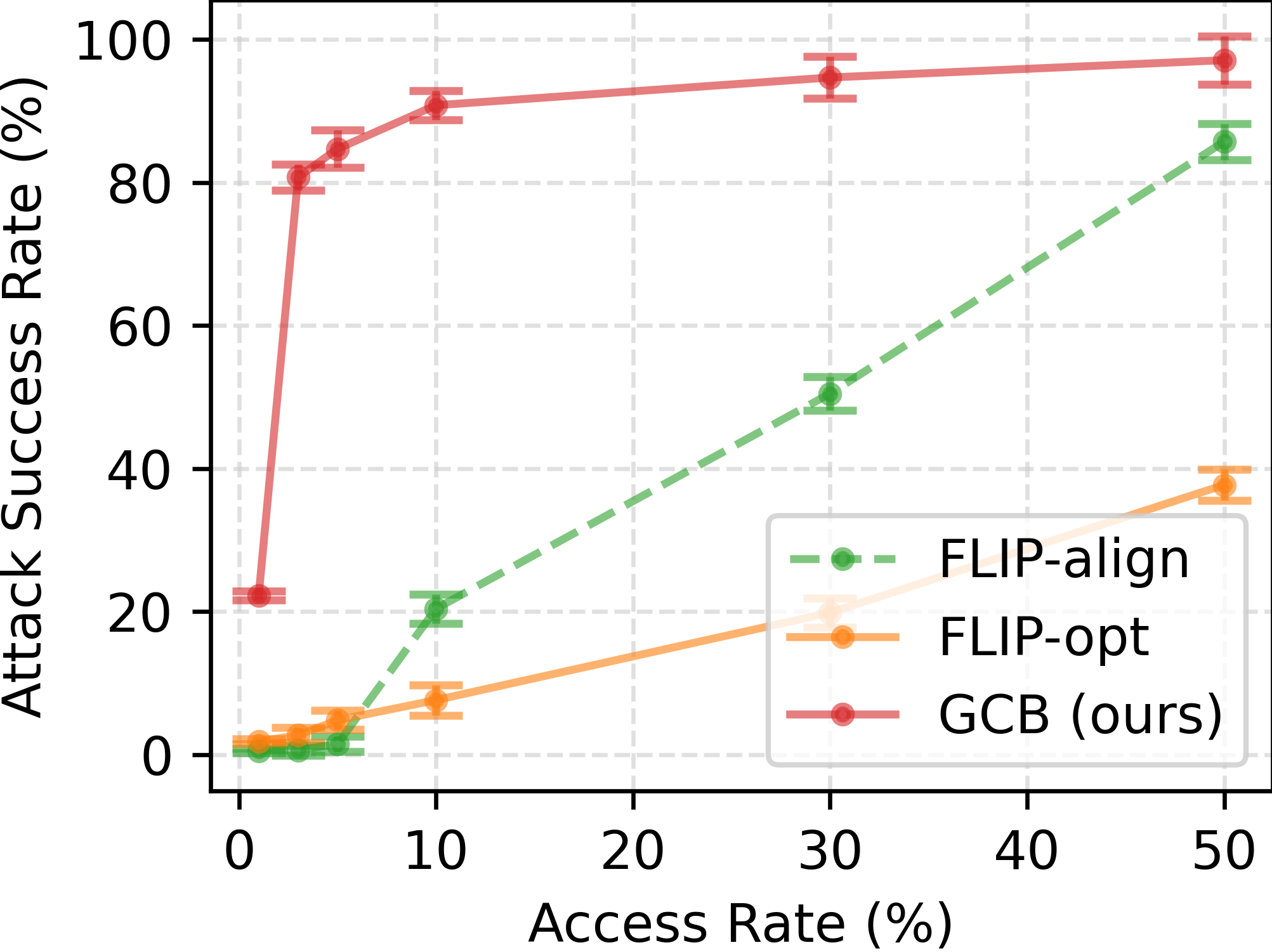}
    \vskip -0.05in
    \caption*{(a) CIFAR-10}
    \label{fig:cifar10_weak}
  \end{minipage}
  \hfill
  \begin{minipage}{0.48\linewidth}
    \centering
    \includegraphics[width=\linewidth]{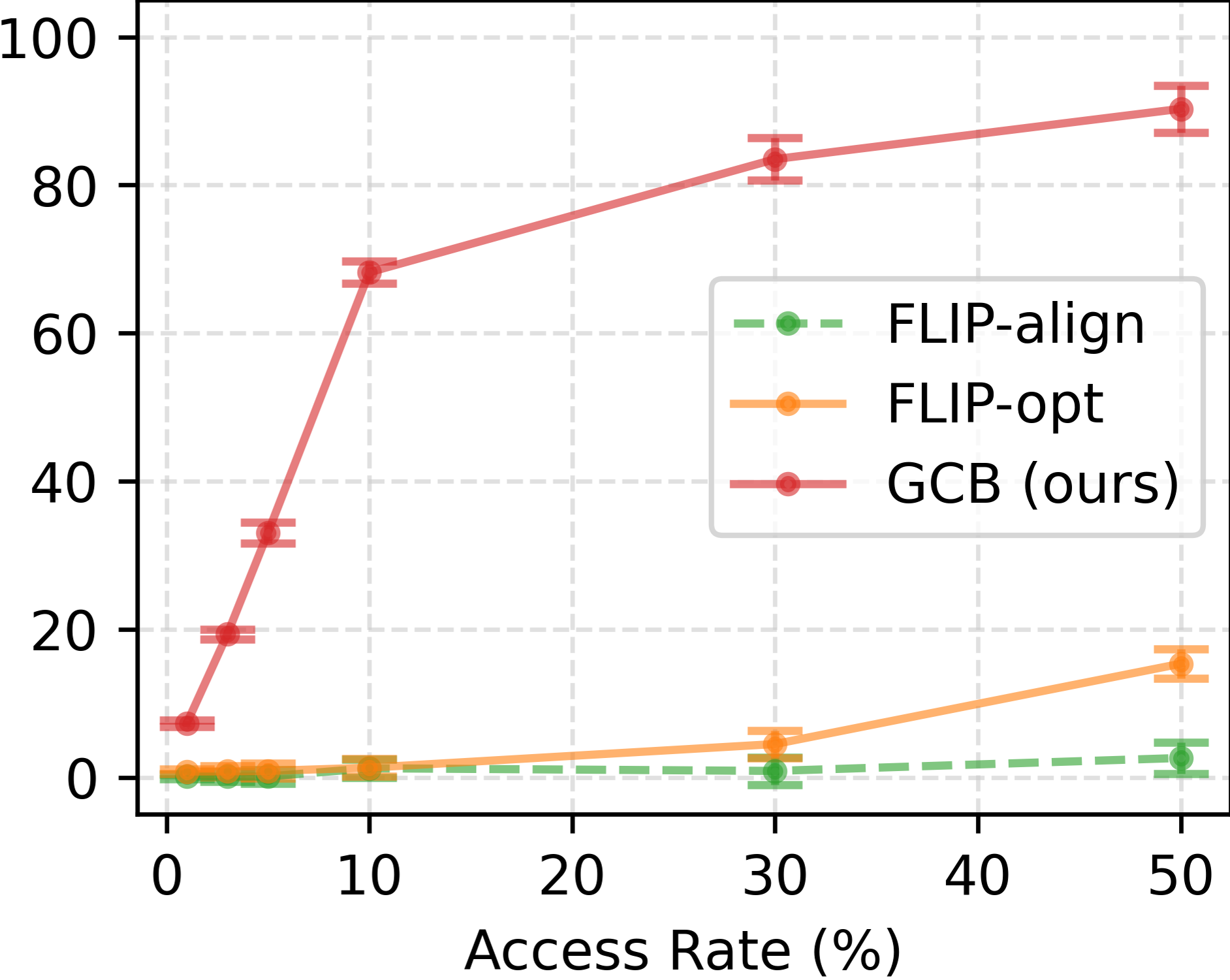}
    \vskip -0.05in
    \caption*{(b) CIFAR-100}
    \label{fig:cifar100_weak}
  \end{minipage}
  \vskip -0.1in
  \caption{Results with error bars under low access rates.}
  \vskip -0.15in
  \label{fig:weak}
\end{figure}

\textbf{Robustness to Architecture.} As introduced in the baseline settings, FLIP is highly sensitive to the victim model's architecture. In contrast, \methodabb{} exhibits high ASR across four distinct architectures: PreActResNet18, EfficientNet-B0, VGG-11, and ViT-B-16, as shown in Table \ref{tab:architecture_performance}. In our experiments, all four architectures achieve ASR exceeding 90\% on every tested dataset, with an average ASR above 96\%. This demonstrates robustness across the architectures tested here; it does not imply invariance to every possible victim architecture.

\textbf{Generalized Threat Model.} Our threat model can be extended to weaker assumptions. We propose a generalized threat model where attackers can access only a small portion of the entire dataset and subsequently poison an even smaller subset of the accessed data. This extension broadens the clean-image backdoor threat to individual annotators with very limited dataset access. As shown in Fig. \ref{fig:weak}, when accessing only 10\% of the training dataset, \methodabb{} achieves an ASR of 90.3\% on CIFAR-10 and 68.2\% on CIFAR-100. In comparison, the current SOTA baseline FLIP achieves only 20.4\% and 1.3\% ASR on CIFAR-10 and CIFAR-100, respectively, with the same data access.

\begin{figure*}[t]
    \centering
    % 左侧两个子图（横向并排）
    \begin{minipage}{0.73\textwidth}
        \centering
        \begin{minipage}{0.48\textwidth}
            \centering
            \includegraphics[width=\linewidth]{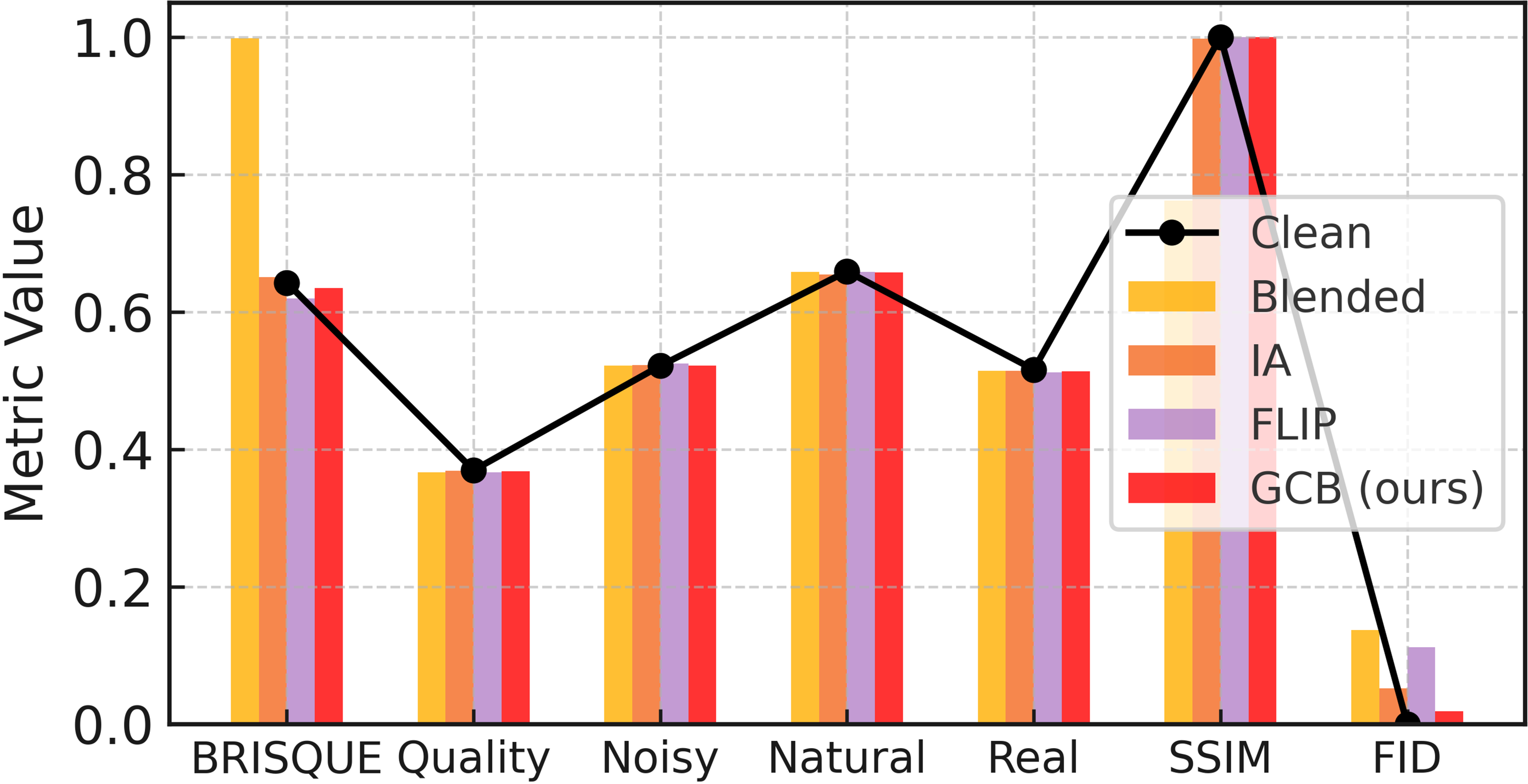}
            \vskip -0.05in
            \caption*{(a) CIFAR-10}
            \label{fig:steal_cifar10}
        \end{minipage}
        \hfill
        \begin{minipage}{0.48\textwidth}
            \centering
            \includegraphics[width=\linewidth]{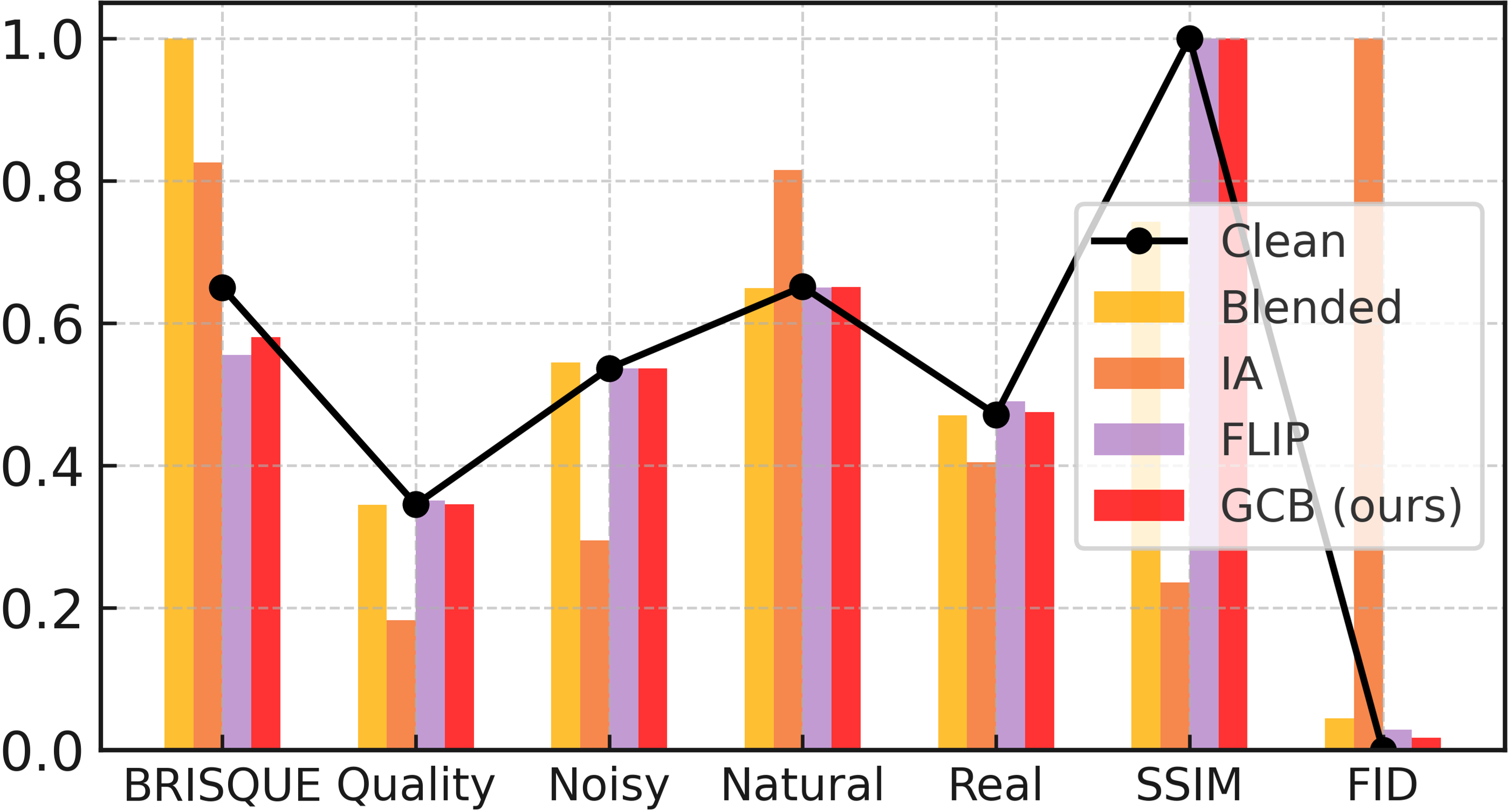}
            \vskip -0.05in
            \caption*{(b) Tiny-ImageNet}
            \label{fig:steal_imagenet}
        \end{minipage}
        \vskip -0.05in
        \caption{Difference from clean images. Closeness to ``Clean" values indicates stealthiness.}
        \label{fig:steal}
    \end{minipage}
    \hfill
    % 右侧单图
    \begin{minipage}{0.26\textwidth}
        \centering
        \includegraphics[width=\linewidth]{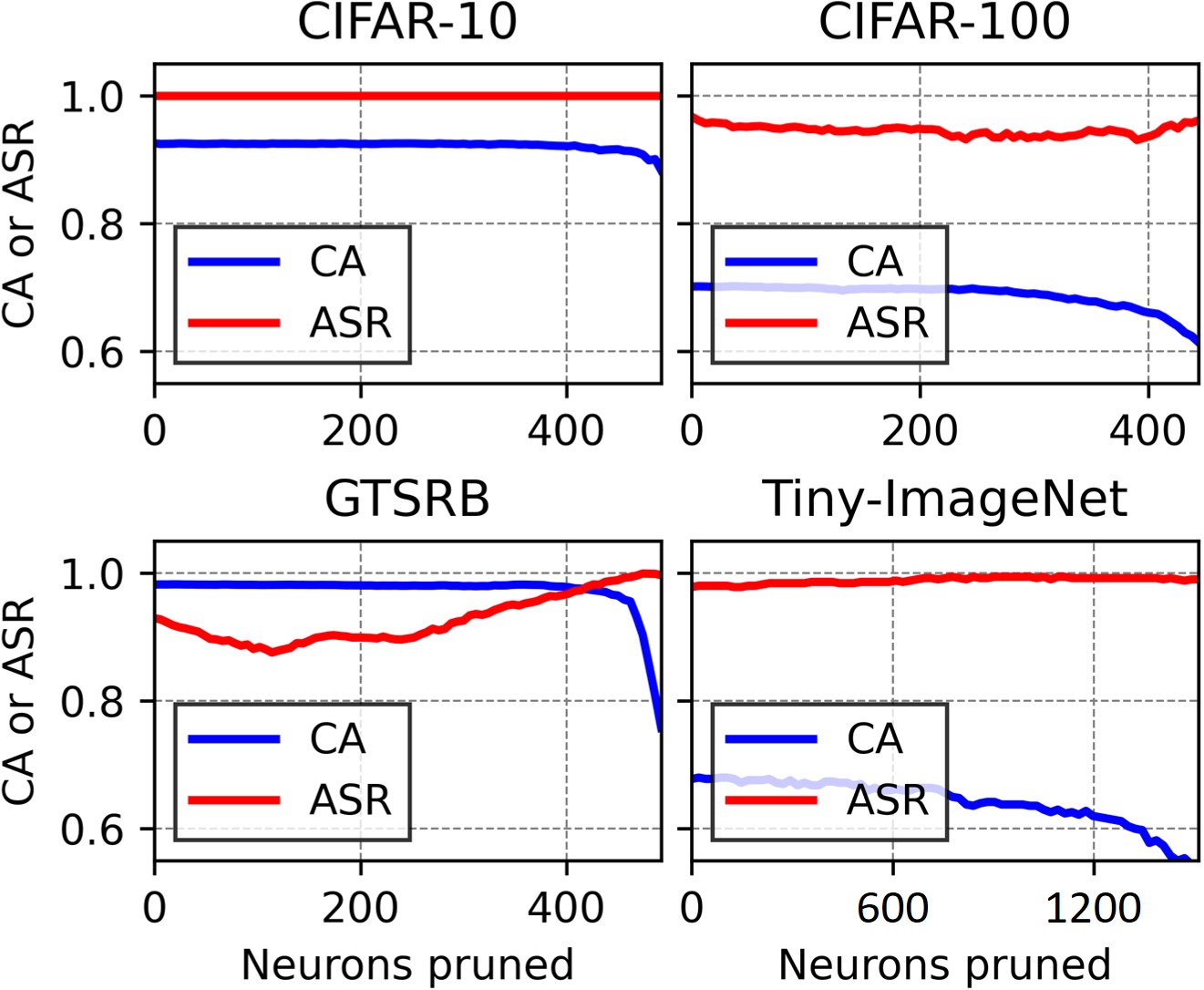}
        \vskip -0.05in
        \caption{Fine-pruning.}
        \label{fig:fp}
    \end{minipage}
    \vskip -0.1in
\end{figure*}

\textbf{Other Vision Tasks.}
We evaluate whether \methodabb{} can be adapted to several supervised vision tasks by changing how the label condition \(y\) is represented---using one-hot encoding for classification and no embedding for regression. For multi-label classification, we compared our approach with CIB~\citep{chen2022cib} using a 5\% poison rate on the VOC07 and VOC12 datasets. As shown in Table \ref{tab:universal}, CIB achieves approximately 15\% higher ASR but significantly underperforms in Mean Average Precision (MAP), dropping by about 2\% overall and around 20\% for the source class. This reduction in MAP compromises its stealthiness. Additionally, we evaluate Image Regression and Semantic Segmentation, where the tested existing clean-image backdoors are ineffective. As illustrated in Table \ref{tab:universal}, our attack succeeds under these configurations, demonstrated by a substantial decrease in Attack Mean Square Error (AE) compared to the clean dataset. Detailed task configurations are provided in Appendix\ref{ap:universal}.

\subsection{Ablation Study.}

We conducted ablation studies on three key components of our design: GAN loss (for Existence), information loss (for Separability), and label condition (for Irrelevancy). The results are presented in Table \ref{tab:ablation}.

\textbf{(a) GAN Loss.} We eliminate the discriminator \(D\) from C-InfoGAN and apply an \(l_{\infty}\)-norm constraint to the generator. This variant loses effectiveness in our experiments. Its behavior is consistent with the generator exploiting the recognition network \(Q\) rather than learning a trigger that transfers to the victim model.

\textbf{(b) Information Loss.} Removing the information loss yields a conditional image-to-image GAN variant with the same generator architecture. To perform the attack, we select the darkest 1\% of images in the dataset as poisoned images and treat trigger generation as a style-transfer problem. Under this particular manually specified trigger and selection rule, attack performance degrades significantly; this ablation does not imply that all manually designed triggers are ineffective.

\textbf{(c) Label Condition.} We remove \( y \) as a prior condition from all components in C-InfoGAN. The results show only a minor decrease in ASR, likely because the UNet generator preserves the original appearance, reducing the reliance on label conditioning. However, we observe that removing \textit{LC} increases the likelihood of mode collapse, causing C-InfoGAN to generate uniform features. For a more detailed analysis of this effect, please refer to Appendix\ref{sec:label_condition}.

\subsection{Stealthiness and Robustness Measure.}

\subsubsection{Stealthiness of \methodabb{}.} We evaluated the stealthiness of our method using seven metrics from BackdoorBench~\citep{wu2024backdoorbench}. As shown in Fig. \ref{fig:steal}, clean-image backdoor attacks, such as FLIP and GCB, are significantly stealthier than poison-image backdoors like Blended~\citep{chen2017blended} and IA~\citep{nguyen2020input}.
Clean-image methods leverage benign features for the backdoor, creating poisoned data that closely resembles clean images. This makes them difficult to detect using image-quality metrics like SSIM and BRISQUE. While the Frechet Inception Distance (FID) assesses distributional differences, our experiments show that GCB excels here as well, with its triggered images closely matching the clean distribution across all tested metrics.

\subsubsection{Robustness of \methodabb{}.} To evaluate the robustness of \methodabb{}, we apply several common image corruption techniques, including JPEG Compression \cite{xue2023compression}, Gaussian Smoothing \cite{xu2018feature}, Color Shift \cite{jiang2023color}, Color Shrink \cite{xu2018feature}, and Affine Transformation \cite{qiu2021deepsweep}. These transformations are widely recognized for their effectiveness in mitigating backdoor attacks. In our experiments, we apply these transformations at \textbf{test time} to each input image before feeding them into the victim model.  
The results in Table \ref{tab:corruptions} demonstrate that \methodabb{} remains highly robust against these corruptions. While the CA drops by more than 10\% in all cases, the ASR consistently remains close to 100\%. 
Additional results on other baseline attacks and datasets are in Appendix\ref{sec:robustness_measure}.  

\begin{table}[t]
  \centering
  \begin{minipage}{0.53\linewidth}
    \centering
    \begin{small}
    \begin{tblr}{
      width = \linewidth,
      rowsep = 0.8pt,
      colsep = 0.8pt,
      colspec = {Q[120]Q[50]Q[50]Q[50]},
      cells = {c},
      cell{2}{1} = {c=4}{0.5\linewidth},
      cell{7}{1} = {c=4}{0.5\linewidth},
      hline{1,12} = {-}{0.10em},
      hline{2-3,7-8} = {-}{},
    }
    \textbf{poison rate→}         & \textbf{1\%}   & \textbf{0.5\%} & \textbf{0.1\%} \\
    \textbf{CIFAR-10}     &                 &                 &                 \\
    \textit{(w/o $L_{GAN}$)}    & 8.97           & 4.14            & 1.90            \\
    \textit{(w/o $L_{info}$)}   & 42.9           & 11.4            & 2.87            \\
    \textit{(w/o LC)}     & 98.9           & 93.1            & 85.3            \\
    \textit{Ours}         & \textbf{100.0} & \textbf{100.0}  & \textbf{98.5}   \\
    \textbf{CIFAR-100}    &                 &                 &                 \\
    \textit{(w/o $L_{GAN}$)}    & 3.41           & 1.80            & 0.45            \\
    \textit{(w/o $L_{info}$)}   & 28.7           & 8.12            & 1.34            \\
    \textit{(w/o LC)}     & 84.7           & 68.4            & 34.6            \\
    \textit{Ours}         & \textbf{96.7}  & \textbf{92.1}   & \textbf{45.9}   \\
    \end{tblr}
    \end{small}
    \vskip -0.06in
    \caption{Ablation Study.}
    \label{tab:ablation}
  \end{minipage}
  \hfill
  \begin{minipage}{0.43\linewidth}
    \centering
    \begin{small}
    \begin{tblr}{
      width = \linewidth,
      rowsep = 0.4pt,
      colsep = 0.2pt,
      colspec = {Q[120]Q[50]Q[50]},
      cells = {c},
      cell{2}{1} = {c=3}{},
      cell{7}{1} = {c=3}{},
      hline{1,12} = {-}{0.10em}, 
      hline{2,3,7,8} = {-}{0.05em},
    }
    \textbf{\textbf{Corruptions↓}}  & \textbf{\textbf{CA}} & \textbf{\textbf{ASR}} \\
    \textbf{Test-Time}              &                      &                       \\
    JPEG                            & 77.6                 & 100                   \\
    Color Shift                     & 84.4                 & 98.4                  \\
    Color Shrink                    & 84.9                 & 100                   \\
    Affine                          & 84.2                 & 99.9                  \\
    \textbf{\textbf{Training-Time}} &                      &                       \\
    JPEG                            & 93.0                 & 100                   \\
    Color Shift                     & 92.1                 & 97.6                  \\
    Color Shrink                    & 92.3                 & 99.8                  \\
    Affine                          & 91.8                 & 100                   
    \end{tblr}

    \end{small}
    \vskip -0.06in
    \caption{Robustness Under Different Corruptions.}
    \label{tab:corruptions}
  \end{minipage}
  \vskip -0.1in
\end{table}

\begin{table*}
\centering
\begin{small}
\begin{tblr}{
  width = 1.0\textwidth,
  rowsep = 0.2pt,
  colsep = 0.2pt,
  colspec = {Q[80]Q[44]Q[44]Q[44]Q[44]Q[44]Q[44]Q[44]Q[44]Q[44]Q[44]Q[44]Q[44]Q[44]Q[44]Q[44]Q[44]Q[44]Q[44]Q[55]},
  cells = {c},
  cell{1}{2} = {c=2}{0.094\linewidth},
  cell{1}{4} = {c=2}{0.094\linewidth},
  cell{1}{6} = {c=2}{0.094\linewidth},
  cell{1}{8} = {c=2}{0.094\linewidth},
  cell{1}{10} = {c=2}{0.094\linewidth},
  cell{1}{12} = {c=2}{0.094\linewidth},
  cell{1}{14} = {c=2}{0.094\linewidth},
  cell{1}{16} = {c=2}{0.094\linewidth},
  cell{1}{18} = {c=2}{0.094\linewidth},
  hline{1,12} = {-}{0.1em},
  hline{3,11} = {-}{0.05em},
  hline{2} = {3,5,7,9,11,13,15,17,19}{r},
  hline{2} = {2,4,6,8,10,12,14,16,18}{l},
  % cell{4-5}{3}      = {bg=customgreen},   cell{3,6-11}{3}        = {bg=customred},
  % cell{3,4,8}{5}    = {bg=customgreen},   cell{5-7,9-11}{5}      = {bg=customred},
  % cell{3,5,6,8,9}{7}= {bg=customgreen},   cell{4,7,10-11}{7}     = {bg=customred},
  % cell{3,5,6}{9}    = {bg=customgreen},   cell{4,7-11}{9}        = {bg=customred},
  % cell{3,9}{11}     = {bg=customgreen},   cell{4-8,10-11}{11}    = {bg=customred},
  % cell{3,4,6,8,9}{13}= {bg=customgreen},  cell{5,7,10-11}{13}    = {bg=customred},
  % cell{3-7}{15}     = {bg=customgreen},   cell{10}{15}           = {bg=customgreen},
  %                                             cell{8-9,11}{15}   = {bg=customred},
  % cell{3-6,8}{17}   = {bg=customgreen},   cell{7,9-11}{17}       = {bg=customred},
  % cell{3-9}{19}     = {bg=customgreen},   cell{10-11}{19}        = {bg=customred},
}
Defense→ & ABL & & D-BR & & CLP & & EP & & NAB & & ASD & & MSPC & & ReBack & & PIPD & & Average \\
Attack↓ & CA & ASR & CA & ASR & CA & ASR & CA & ASR & CA & ASR & CA & ASR & CA & ASR & CA & ASR & CA & ASR & ASR \\
BadNet & 41.0 & 72.5 & 91.0 & 1.5 & 91.5 & 4.5 & 92.8 & 12.7 & 86.3 & 0.3 & 92.0 & 2.1 & 92.8 & 0.3 & 91.7 & 4.3 & 92.2 & 0.5 & 11.0 \\
Blend  & 58.6 & 0.0 & 85.1 & 0.0 & 93.1 & 91.6 & 92.5 & 95.6 & 88.8 & 43.8 & 93.0 & 5.3 & 92.7 & 0.7 & 91.7 & 2.4 & 92.4 & 5.3 & 27.2 \\
BPP    & 49.3 & 18.3 & 88.5 & 85.5 & 91.6 & 3.4 & 90.5 & 10.5 & 84.5 & 79.4 & 92.5 & 99.4 & 90.5 & 2.8 & 90.1 & 1.8 & 92.4 & 0.9 & 33.6 \\
IA     & 62.5 & 31.5 & 85.3 & 84.8 & 84.7 & 10.3 & 90.1 & 6.7 & 90.2 & 74.4 & 92.3 & 19.8 & 92.5 & 5.3 & 87.9 & 1.7 & 91.3 & 4.0 & 26.5 \\
SIG    & 54.3 & 50.1 & 91.3 & 49.6 & 93.1 & 79.0 & 92.1 & 83.6 & 90.1 & 82.1 & 92.2 & 99.5 & 91.0 & 10.3 & 87.4 & 29.9 & 92.5 & 13.6 & 55.3 \\
SSBA   & 59.8 & 82.6 & 83.1 & 3.0 & 93.2 & 1.1 & 92.2 & 99.9 & 88.9 & 49.1 & 93.3 & 7.1 & 90.9 & 21.5 & 85.1 & 6.6 & 89.9 & 17.2 & 32.0 \\
WaNet  & 77.3 & 26.2 & 84.3 & 60.2 & 90.5 & 0.8 & 89.9 & 63.3 & 89.9 & 11.7 & 91.7 & 8.8 & 93.0 & \textbf{54.2} & 90.2 & \textbf{84.4} & 92.7 & 11.4 & 35.7 \\
FLIP   & 50.0 & 99.0 & 83.9 & 22.1 & 92.2 & 20.6 & 90.0 & 80.9 & 79.3 & 70.2 & 86.9 & 62.2 & 91.6 & 17.2 & 90.0 & 39.7 & 90.7 & 66.9 & 53.2 \\
GCB    & 69.3 & \textbf{100.0} & 84.2 & \textbf{100.0} & 92.4 & \textbf{100.0} & 90.6 & \textbf{100.0} & 88.8 & \textbf{100.0} & 90.9 & \textbf{100.0} & 91.5 & 23.9 & 88.7 & 71.6 & 91.8 & \textbf{87.7} & \textbf{87.0} \\
\end{tblr}
\end{small}
\vskip -0.06in
\caption{Comparison of different attack methods against advanced backdoor defense methods.}
\label{tab:defenses}
\vskip -0.08in
\end{table*}

\section{Defenses}

\subsection{Classic Defenses}

% We present the most representative defense methods in this section. Comprehensive evaluations of \methodabb{} against BackdoorBench defenses are provided in Appendix\ref{ap:defense_backdoorbench}.

\subsubsection{Neural Cleanse.}
\begin{figure}[t]
\centering
\begin{minipage}{0.42\linewidth}
    \centering
    \includegraphics[width=\linewidth]{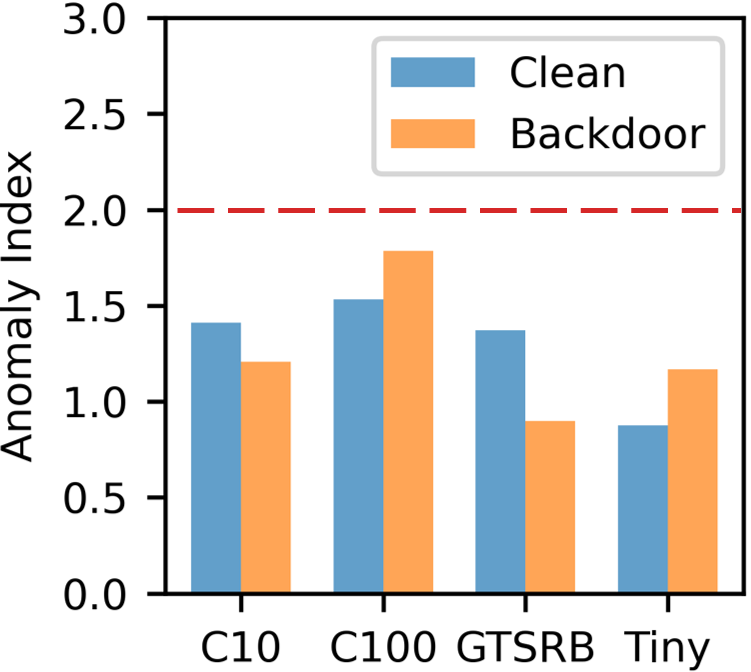}
    \vskip -0.06in
    \caption{NC Defense.}
    \label{fig:neural-cleanse}
\end{minipage}
\hfill
\begin{minipage}{0.56\linewidth}
    \centering
    \begin{small}
    \begin{tblr}{
      width = \linewidth,
      rowsep = 0.4pt,
      colsep = 0.24pt,
      colspec = {Q[300]Q[100]Q[100]Q[100]Q[100]},
      cell{1}{2} = {c=2}{0.28\linewidth},
      cell{1}{4} = {c=2}{0.28\linewidth},
      cells = {c},
      hline{1,3,8} = {-}{0.10em}
    }
    Dataset→  & CIFAR10 &     & CIFAR100 &      \\
    Method↓   & CA       & ASR & CA        & ASR  \\
    SPL       & 91.9     & 100 & 67.2      & 78.2 \\
    PRL       & 89.7     & 100 & 66.8      & 87.6 \\
    BootStrap & 88.4     & 100 & 57.6      & 93.9 \\
    DivideMix & 92.1     & 100 & 73.4      & 86.7 \\
    MentorMix & 89.9     & 100 & 69.0      & 92.7 
    \end{tblr}
    \end{small}
    \vskip -0.1in
    \captionof{table}{Noisy training mitigation. FLIP's performance in Table \ref{tab:flip_noisy}.}
    \label{tab:learning_results}
\end{minipage}
\vskip -0.1in
\end{figure}

Neural Cleanse~\citep{wang2019neural} uses anomaly scores to detect backdoors in DNN models. However, Fig. \ref{fig:neural-cleanse} shows that Neural Cleanse is hard to differentiate backdoor-attacked datasets and clean ones, because their scores are similar and below the 2.0 threshold. This is due to Neural Cleanse's focus on static adversarial patches, while our attack uses a dynamic, global trigger function, making trigger reconstruction difficult.  

\subsubsection{STRIP.}
\begin{figure}[t] % 't' 将图放置在页面顶部
    \centering
    \includegraphics[width=\linewidth]{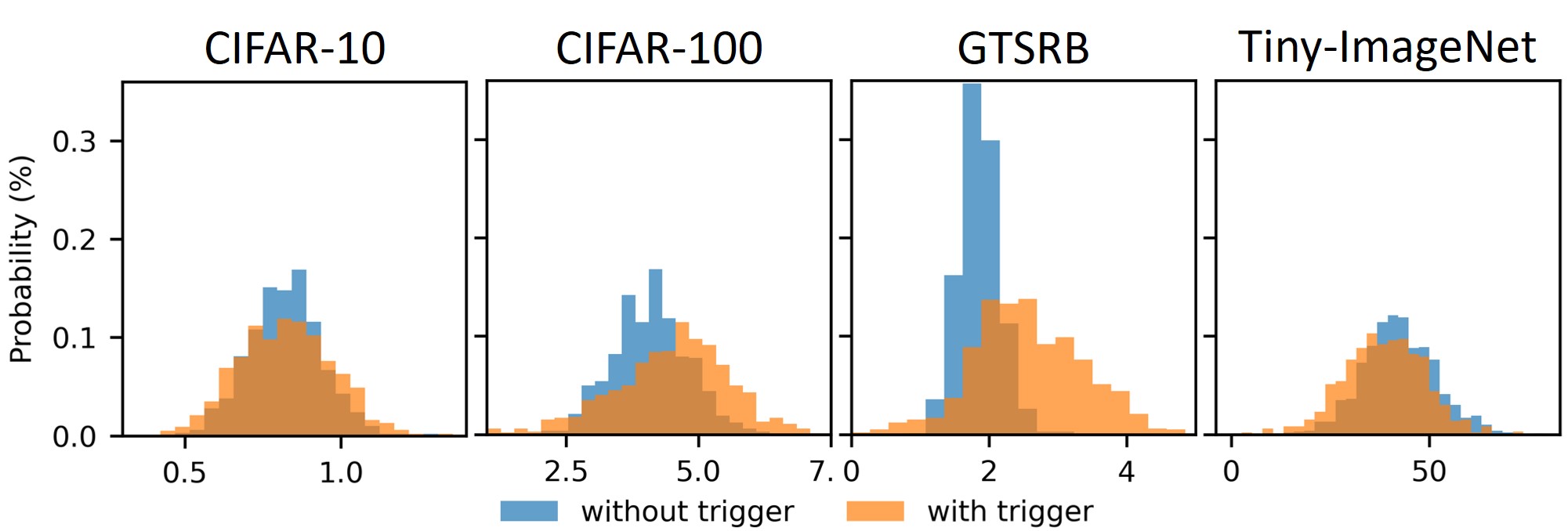}
    \vskip -0.1in
    \caption{STRIP normalized entropy distribution of \methodabb{}.}
    \vskip -0.1in
    \label{fig:strip}
\vskip -0.0in
\end{figure}

STRIP~\citep{gao2019strip} measures class prediction entropy through input perturbations. Fig. ~\ref{fig:strip} shows a notable similarity in entropy distribution for clean and poisoned subsets. Since C-InfoGAN uses benign features of various intensities as triggers, it can yield similar STRIP behaviors for samples with or without trigger. Therefore, our \methodabb{} attack is resilient to STRIP defense.

\subsubsection{Fine-Pruning.} Fine-Pruning~\citep{liu2018fine}, which prunes high-activation neurons, is ineffective against our attack. Because our backdoor uses natural benign features, it creates a complex activation pattern that evades detection. Consequently, as shown in Fig. \ref{fig:fp}, the ASR on CIFAR-10 remains constant during pruning. On CIFAR-100, the ASR initially drops but then increases sharply, demonstrating the defense's failure.
% Fine-Pruning~\citep{liu2018fine} tries to mitigate backdoor behaviors by pruning high-activation neurons. As shown in Fig. \ref{fig:fp}, on CIFAR-10, the ASR remains unchanged regardless of pruning. For CIFAR-100, ASR initially decreases but then rapidly increases as more neurons are pruned. Our backdoor attack uses natural benign features, resulting in a robust and complex activation pattern that Fine-Pruning cannot detect. This suggests Fine-Pruning is ineffective against our backdoor attack.

% \textbf{Grad-Cam.}
% We employ Grad-Cam~\citep{jacobgilpytorchcam} to visualize the regions of an image that are most relevant to a model’s prediction. Grad-Cam can also highlight potential trigger regions activated by different backdoor attacks. Results in Fig. \ref{fig:gradcam} reveal that \methodabb{}’s triggers are dispersed and centrally located, whereas the triggers of other attacks are localized and prominent. This indicates that \methodabb{} achieves global feature dominance, diverging from Grad-Cam’s localized focus, and thus is more resilient to Grad-Cam-based detection.

\subsection{SOTA Backdoor Defenses}
As shown in Table \ref{tab:defenses}, we evaluated our attack against nine state-of-the-art (SOTA) backdoor defenses, including six since 2023 \cite{zheng2022bnpep, liu2023nab, gao2023asd, ma2024need, chen2024progressive, pal2024backdoor}. The proposed method, $\text{\methodabb{}}$, exhibits strong resistance against most of them. Only one defense, MSPC \cite{pal2024backdoor}, proved effective, but most of the other attacks suffer from greater performance degradation. Even the latest defenses like ReBack \cite{ma2024need} and PIPD \cite{chen2024progressive} failed to remove the backdoor, only slightly lowering the ASR. We attribute this robustness to the attack's inherent asymmetric trigger, where the training and testing triggers are different. This design bypasses the common assumption of latent separability \cite{qi2022revisiting} that many defenses rely on.

% We consider SOTA and well-known Backdoor defenses in the past few years \cite{li2021anti, chen2022dbrst, zheng2022channel, huang2022decouple, zheng2022bnpep, liu2023nab, gao2023asd, ma2024need, chen2024progressive}, including 2 in 2024 \cite{ma2024need, chen2024progressive} and 3 in 2023 \cite{zheng2022bnpep, liu2023nab, gao2023asd}. Results in Table \ref{tab:defenses} show that although not designed to be, \methodabb{} shows very strong resistance against most of the existing SOTA backdoor defense methods. Only one of them (DBD \cite{huang2022decouple}) is shown to be effective against our attack, with the large sacrifice CA decrease (from 93.9\% to 76.6\%). Even the two most advanced defenses in 2024 (ReBack \cite{ma2024need} and PIPD \cite{chen2024progressive}) only degrade the ASR a bit but failed to remove the backdoor. This can be largely attributed to clean-image backdoors are naturally \textbf{asymmetric trigger}, meaning the trigger in training and testing time are different, thus bypassing many defenses' assumptions like latent separability \cite{qi2022revisiting}. 

\subsection{Adaptive Defenses}

\subsubsection{Noisy Training.} Clean-image backdoors embed triggers by poisoning only labels. As a result, training techniques that are robust to label noise might diminish the effectiveness of these faulty labels. We evaluated five noisy training methods: Self-Paced Learning (SPL)~\citep{kumar2010spl}, Perturbation Robust Learning (PRL)~\citep{wong2020prl}, Bootstrap~\citep{reed2014bootstrap}, DivideMix~\citep{li2020dividemix}, and MentorMix~\citep{jiang2020mentormix}. As shown in Table \ref{tab:learning_results}, none of these methods effectively defend against our attack. This is likely because \methodabb{}'s incorrect labels constitute misleading knowledge rather than random noise, which contradicts the basic assumption of noisy training.

\begin{figure}[t]
  \centering
  \begin{minipage}{0.48\linewidth}
    \centering
    \includegraphics[width=\linewidth]{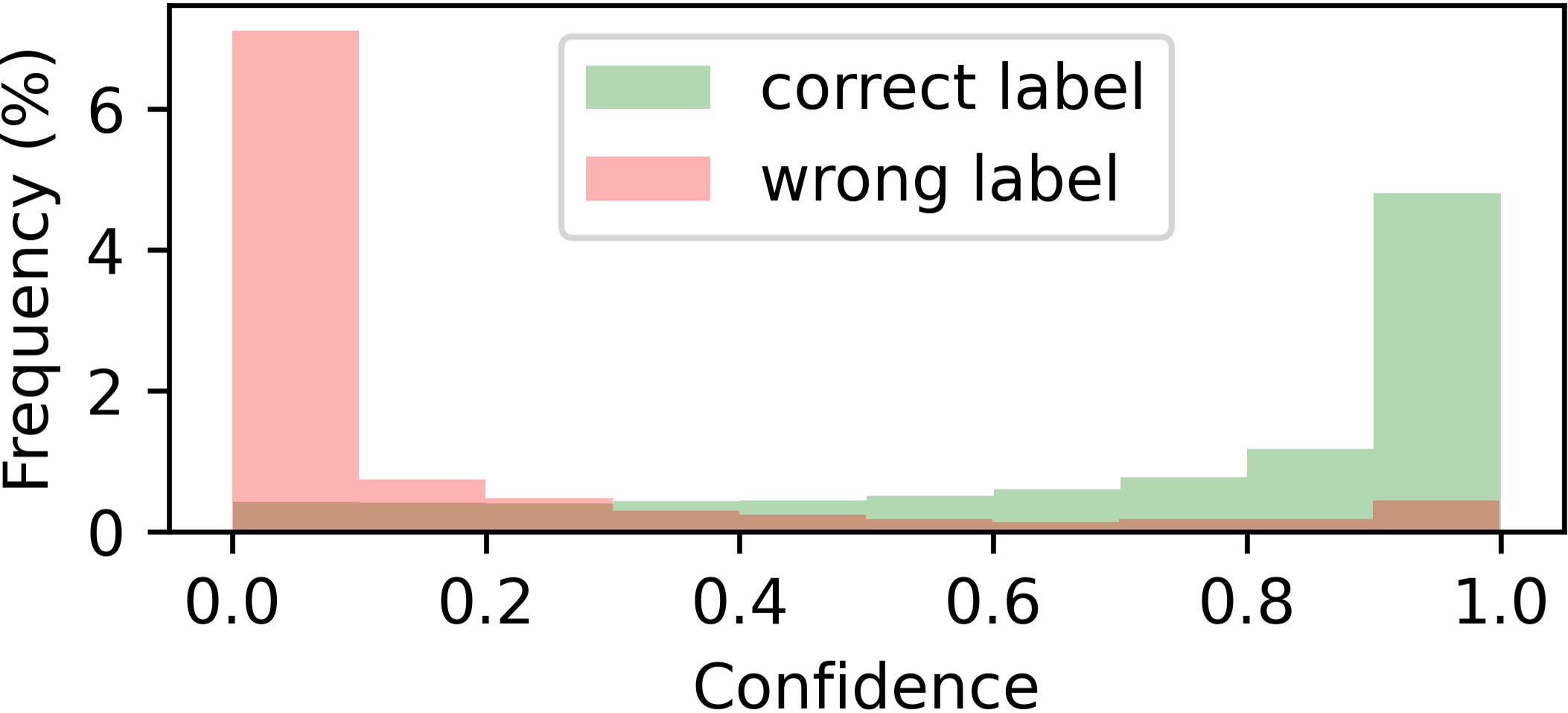}
  \vskip -0.05in
    \caption*{(a) Noisy label.}
    \label{fig:noise}
  \end{minipage}
  \hfill
  \begin{minipage}{0.48\linewidth}
    \centering
    \includegraphics[width=\linewidth]{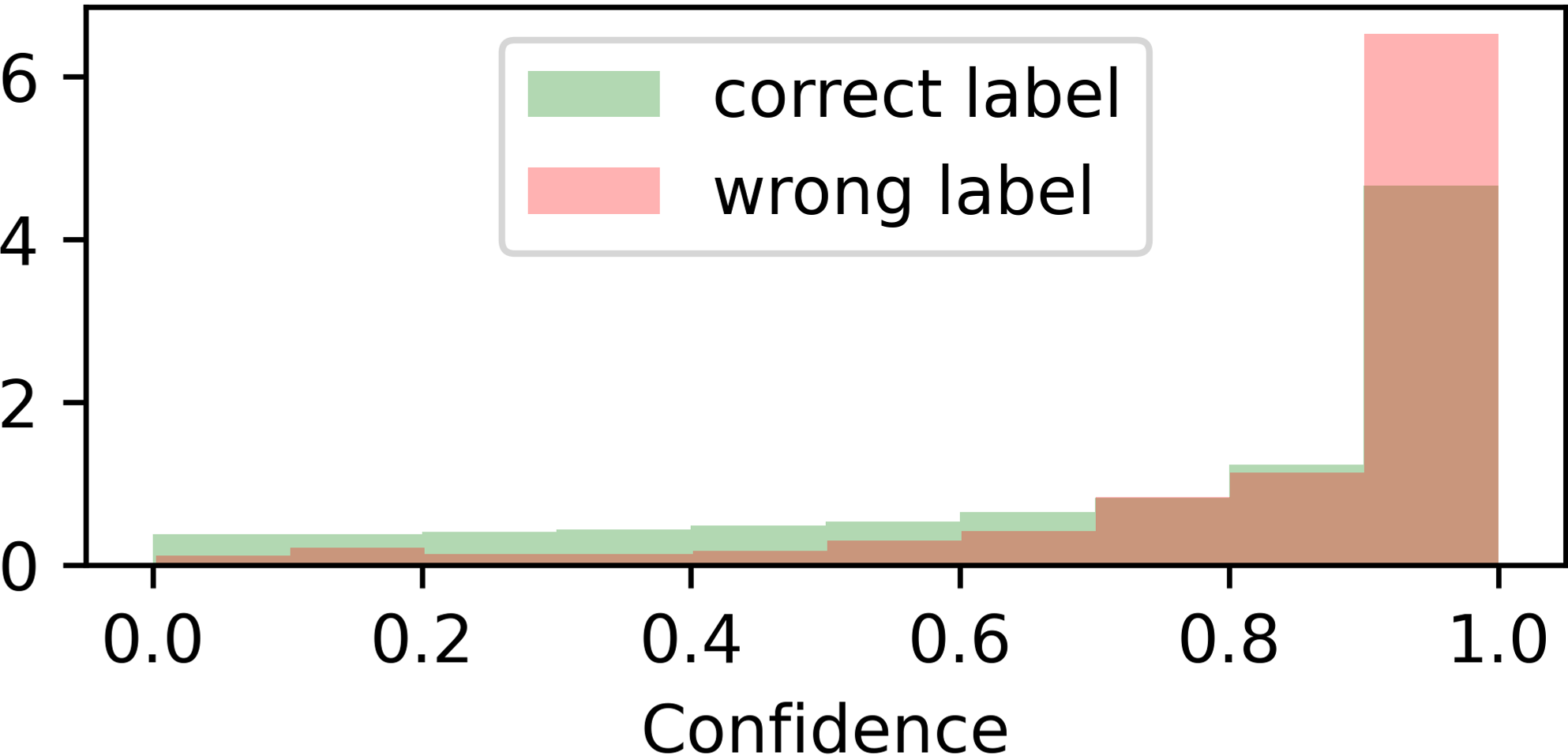}
  \vskip -0.05in
    \caption*{(b) Poisoned label.}
    \label{fig:poison}
  \end{minipage}
  \vskip -0.05in
  \caption{Confidence for two different label issues.}
  \label{fig:confidence}
  \vskip -0.05in
\end{figure}

\subsubsection{Label Cleaning.}
% Advanced label cleaning techniques \cite{northcutt2021confidentlearning, kuan2022labelquality, kuan2022ood} can automatically identify potentially mislabeled data in a suspicious dataset. These methods typically involve training a model and flagging images with low confidence scores as having label problems (e.g., Confident Learning \cite{northcutt2021confidentlearning}). We employ CleanLab, a widely used tool with strong community support (9.4k GitHub stars), to evaluate its effectiveness in defending against \methodabb{}. Using CleanLab, we analyze the confidence score distribution, as shown in Fig. \ref{fig:confidence}. Noisy labels are easily separated because they exhibit very low confidence scores, as their label issues are random and the DNN cannot establish a reliable mapping between images and their corresponding labels. In contrast, \methodabb{} selects specific images to bind to target labels, creating a strong connection and resulting in higher confidence scores than benign images, as illustrated in Fig. \ref{fig:poison}. Advanced label cleaning techniques are ineffective against \methodabb{}.

We evaluated label cleaning against \methodabb{} using CleanLab~\cite{northcutt2021confidentlearning}, which uses model-predicted probabilities to identify likely label errors. In our experiment, the selected poisoned samples receive confidence scores comparable to or higher than those of many benign samples, and CleanLab does not effectively isolate them under the tested configuration (Fig.~\ref{fig:confidence}). This result is specific to the evaluated model, data, and CleanLab settings and does not imply that all label-cleaning methods must fail.

% \begin{figure}[t]
% \begin{center}
% \vskip -0.05in
% \centerline{\includegraphics[width=\linewidth]{figures/relabeling_mitigation}}
% \vskip -0.11in
% \caption{Relabeling Mitigation.}
% \label{fig:relabeling}
% \end{center}
% \vskip -0.3in
% \end{figure}

% \textbf{Relabeling Mitigation.}
% Relabeling is the most direct mitigation strategy against \methodabb{}. Since only labels are altered in the poisoned dataset, relabeling a portion can restore its integrity, thereby increasing CA and reducing ASR. Fig. \ref{fig:relabeling} shows relabeling results on CIFAR-10 dataset, with each data point marked by upper and lower bounds of five independent trials. Tests at poison rates of 0.1\%, 0.5\%, and 1\% indicate that higher poison rates diminish relabeling’s effectiveness. To ensure model security and keep ASR below 20\%, a relabeling rate above 95\% is necessary, implying that nearly the entire dataset must be relabeled.

\section{Conclusion}
We introduced Generative Adversarial Clean-Image Backdoors (\methodabb{}), a stealthy and adaptive backdoor attack that uses C-InfoGAN to optimize trigger patterns embedded within training images. Experiments across 6 datasets, 5 architectures, and 4 tasks showed high attack success rates with minimal drop in clean accuracy and low poison rates. \methodabb{} resists existing defenses, highlighting the need for more robust protections, including in agentic deployments with contextual memory~\citep{xu2026contextual}.

\section{Ethical Implications}

This work has clear dual-use implications, including for context-managed agent deployments~\citep{xu2026llm}. By showing that clean-image backdoors can achieve high attack success with negligible clean-accuracy degradation and very low poison rates, our method lowers the detection barrier for adversaries and could threaten safety-critical domains (e.g., medical imaging, autonomous driving) where model performance is tightly monitored. It also highlights a vulnerable point in the ML supply chain—data labeling—where malicious annotators can implant persistent failures without modifying pixels, and it demonstrates resilience against several existing defenses, potentially enabling longer dwell time for attacks. To mitigate misuse, we advocate (i) rigorous dataset provenance tracking and multi-party labeling/auditing; (ii) routine pre-deployment and continuous post-deployment backdoor screening, including behavior-level tests beyond patch reconstruction; (iii) training-time protections such as robust learning under suspected label noise and cross-source consistency checks; and (iv) defense-driven red-teaming before release. Any code artifacts should be accompanied by a usage license, clear risk disclosures, trace-economic risk assessment~\citep{xu2026agent}, and guardrails (e.g., rate-limited models, withheld attack parameters), including for downstream skill distillation~\citep{xu2026multi}, to reduce replication for abusive purposes. Ultimately, the primary ethical justification for releasing this research is to enable the community to build and evaluate stronger defenses against increasingly stealthy data-centric attacks.

\bigskip

\bibliography{aaai2026}

\appendix
\setlength{\columnsep}{22pt}
\maketitlesupplementary

\section{Mathematical Analysis}
\label{sec:math_analysis}

In this section, we give an idealized analysis of how the C-InfoGAN objective relates to our clean-image backdoor task. The analysis separates two distinct questions: whether the latent code produces distinguishable generated components, and whether the triggered test distribution matches the real subset selected for poisoning. The first follows directly from the mutual-information objective; the second is exact only under ideal distributional alignment and is approximate in practice.

\subsection{C-InfoGAN Analysis}

C-InfoGAN extends InfoGAN by conditioning the generator \(G\), discriminator \(D\), and recognition network \(Q\) on the ground-truth label \(Y\). Let \((X,Y)\sim P_{X,Y}\), independently sample \(C\sim\operatorname{Bernoulli}(p_r)\), and define \(\hat{X}=G(X,C,Y)\). An idealized minimax form of the objective is
\begin{align}
\min_{G,Q}\max_D\quad
V(G,D,Q)
= {} & \mathbb{E}_{X,Y}[\log D(X,Y)] \nonumber\\
& +\mathbb{E}_{X,Y,C}[\log(1-D(\hat{X},Y))] \nonumber\\
& -\lambda\mathbb{E}_{X,Y,C}[\log Q(C\mid\hat{X},Y)].
\end{align}
Because \(C\) and \(Y\) are independent, \(H(C\mid Y)=H(C)\). Following InfoGAN, the last term maximizes the variational lower bound
\[
I(C;\hat{X}\mid Y)
\geq
\mathbb{E}_{X,Y,C}[\log Q(C\mid G(X,C,Y),Y)]+H(C),
\]
with equality when \(Q(c\mid\hat{x},y)=p(c\mid\hat{x},y)\) almost everywhere. At an ideal conditional-GAN equilibrium,
\[
p(\hat{x}\mid y)=p(x\mid y)
\]
for \(P_Y\)-almost every \(y\). This formulation also covers continuous or structured labels through regular conditional distributions. It is a population-distribution statement rather than a guarantee that each generated image occurs in the finite training set. For each such \(y\),
\[
p(\hat{x}\mid y)
=(1-p_r)p(\hat{x}\mid c=0,y)+p_r p(\hat{x}\mid c=1,y).
\]
Therefore, each positive-weight conditional component is absolutely continuous with respect to \(p(\hat{x}\mid y)\), and hence with respect to \(p(x\mid y)\) at the ideal equilibrium.

\subsection{Scoring Function Analysis}

For each label \(y\), let \(P_{c,y}=P_{\hat{X}\mid C=c,Y=y}\) and let \(M_y=(1-p_r)P_{0,y}+p_rP_{1,y}=P_{\hat{X}\mid Y=y}\). The conditional mutual information has the exact decomposition
\begin{align}
I(C;\hat{X}\mid Y)
=\mathbb{E}_{Y}\big[ & (1-p_r)\operatorname{KL}(P_{0,Y}\Vert M_Y) \nonumber\\
& +p_r\operatorname{KL}(P_{1,Y}\Vert M_Y)\big]
\end{align}
and hence
\[
I(C;\hat{X}\mid Y)
=\mathbb{E}_{Y}\!\left[
\operatorname{JSD}_{1-p_r,p_r}(P_{0,Y}\Vert P_{1,Y})
\right].
\]
Thus, maximizing conditional mutual information encourages the two \emph{generated} branches to be distinguishable within each label condition by \(Q\). We define
\[
s(x,y)=Q(c=1\mid x,y)-Q(c=0\mid x,y).
\]
For the all-to-one attack, define the source-label set \(\mathcal{Y}_s=\mathcal{Y}\setminus\{y_t\}\) and the source distribution
\[
P_s=P_{X,Y\mid Y\in\mathcal{Y}_s}.
\]
Let \(P_Y^s\) denote the \(Y\)-marginal of \(P_s\).
Let \(\tau\) be a top-\(p_r\) threshold under \(P_s\), define \(S(x,y)=\mathbb{1}[s(x,y)\geq\tau]\), and let
\[
P_1=\mathcal{L}_{P_s}(X\mid S(X,Y)=1).
\]
The finite poisoned subset consists of the corresponding top-scoring source-pool samples and empirically approximates this population selection rule. Applying \(Q\), which is trained on generated images, to real images is a distribution-transfer step: it is exact under the ideal alignment conditions stated below and otherwise serves as an empirically evaluated selection rule. Conditional GAN matching alone does not imply that the two generated components separately match an arbitrary partition of the real data.

\subsection{Connection to Clean-Image Backdoors}

In our clean-image backdoor attack, images in \(X_1\) have their labels changed to a target class \(y_t\). In the label-aware inference setting evaluated here, the attacker uses
\[
T(x,y)=G(x,c=1,y).
\]
Let \(P_T=\mathcal{L}_{P_s}(T(X,Y))\) denote the triggered distribution of non-target source inputs. The victim model \( f^* \) is trained by minimizing
\begin{align}
f^* = \argmin_f \bigg[ & \sum_{(x_0, y_0) \in (X_0, Y_0)} L(f(x_0), y_0) \nonumber \\
& + \sum_{x_1 \in X_1} L(f(x_1), y_t) \bigg],
\end{align}
with the attack success rate (ASR) defined as:
\[
\text{ASR} = \mathbb{E}_{(x,y) \sim P_s} \mathbb{1}[f^*(T(x,y)) = y_t].
\]
The following bound gives a direct and model-agnostic connection between distributional alignment and attack success.

\paragraph{Proposition.}
Define the target-label error on the selected distribution as
\[
\delta=\Pr_{x\sim P_1}[f^*(x)\neq y_t]
\]
and the distributional mismatch as
\[
\epsilon=\operatorname{TV}(P_T,P_1),
\]
where \(\operatorname{TV}\) is total variation distance. Then
\[
\operatorname{ASR}\geq 1-\delta-\epsilon.
\]

\paragraph{Proof.}
Let \(A=\{x:f^*(x)=y_t\}\). By the definition of total variation distance,
\[
P_T(A)\geq P_1(A)-\operatorname{TV}(P_T,P_1)
=1-\delta-\epsilon.
\]
Since \(P_T(A)=\operatorname{ASR}\), the result follows. \(\hfill\square\)

The proposition separates two requirements for a successful classification attack: the victim model must learn the target label on the selected distribution (small \(\delta\)), and triggered inputs must match that distribution (small \(\epsilon\)). It does not require conditional-entropy minimization. More generally, for any measurable attack loss \(\ell_a\in[0,1]\), total variation gives
\[
\left|\mathbb{E}_{P_T}[\ell_a(f^*(X),y_t)]
-\mathbb{E}_{P_1}[\ell_a(f^*(X),y_t)]\right|
\leq\epsilon,
\]
which provides the analogous conditional statement for tasks evaluated with a bounded loss.

For completeness, consider the following idealized special case. Suppose that (i) \(p(\hat{x}\mid y)=p(x\mid y)\) for \(P_Y\)-almost every \(y\), (ii) \(I(C;\hat{X}\mid Y)=H(C)\), (iii) \(Q\) equals the conditional Bayes posterior, and (iv) the source-pool selector \(S\) identifies the posterior-one component. Condition (ii) implies \(H(C\mid\hat{X},Y)=0\), so the two generated components are mutually singular conditional on almost every \(y\). Conditions (i), (iii), and (iv) then imply \(\Pr_{P_s}(S=1\mid Y=y)=p_r\) for \(P_Y^s\)-almost every source label \(y\) and
\[
P_T=\mathcal{L}(\hat{X}\mid C=1)=P_1.
\]
Consequently, \(\epsilon=0\) in this ideal case. With approximate training and top-quantile selection, equality need not hold; the quality of this alignment is therefore evaluated empirically.

\subsection{Discussion on Assumptions}

The proposition is conditional rather than a general convergence guarantee for GAN training. Conditional adversarial distribution matching is intended to reduce \(\epsilon\), while the conditional mutual-information term makes the generated branches distinguishable and supplies the score used to select real samples. Independence between the sampled variables \(C\) and \(Y\) alone does not prove that the learned trigger is semantically irrelevant. We therefore treat triggered accuracy as an empirical proxy for label preservation, not as a proof of statistical independence. The proposition establishes the ASR implication once \(\delta\) and \(\epsilon\) are small; it does not by itself provide a clean-accuracy guarantee or a finite-sample generalization bound for \(\delta\).

\section{Understanding \methodabb{}'s Asymmetry}
Our experimental results demonstrate that GCB achieves excellent attack performance, evidenced by high Clean Accuracy (CA) and Attack Success Rates (ASR) (see Figures \ref{fig:main_acc_asr}, \ref{fig:learn_curve}, and \ref{fig:weak}). Additionally, GCB exhibits robustness and resilience against defenses, as shown in Tables \ref{tab:corruptions}, \ref{tab:defenses}, \ref{tab:learning_results} and Figures \ref{fig:neural-cleanse}, \ref{fig:strip}, and \ref{fig:fp}.

Achieving both high ASR and robustness is particularly intriguing because attacks that converge quickly and attain high ASR can be easier to detect with some training-time defenses. Our experiments suggest that GCB behaves as an \emph{asymmetric backdoor} attack. During the poisoning stage, the selected samples exhibit relatively weak trigger-related features, whereas the generated triggered images at inference exhibit stronger features. We use the following visualization to investigate this empirical explanation.

\begin{figure*}[ht]
    \centering
    \includegraphics[width=0.8\linewidth]{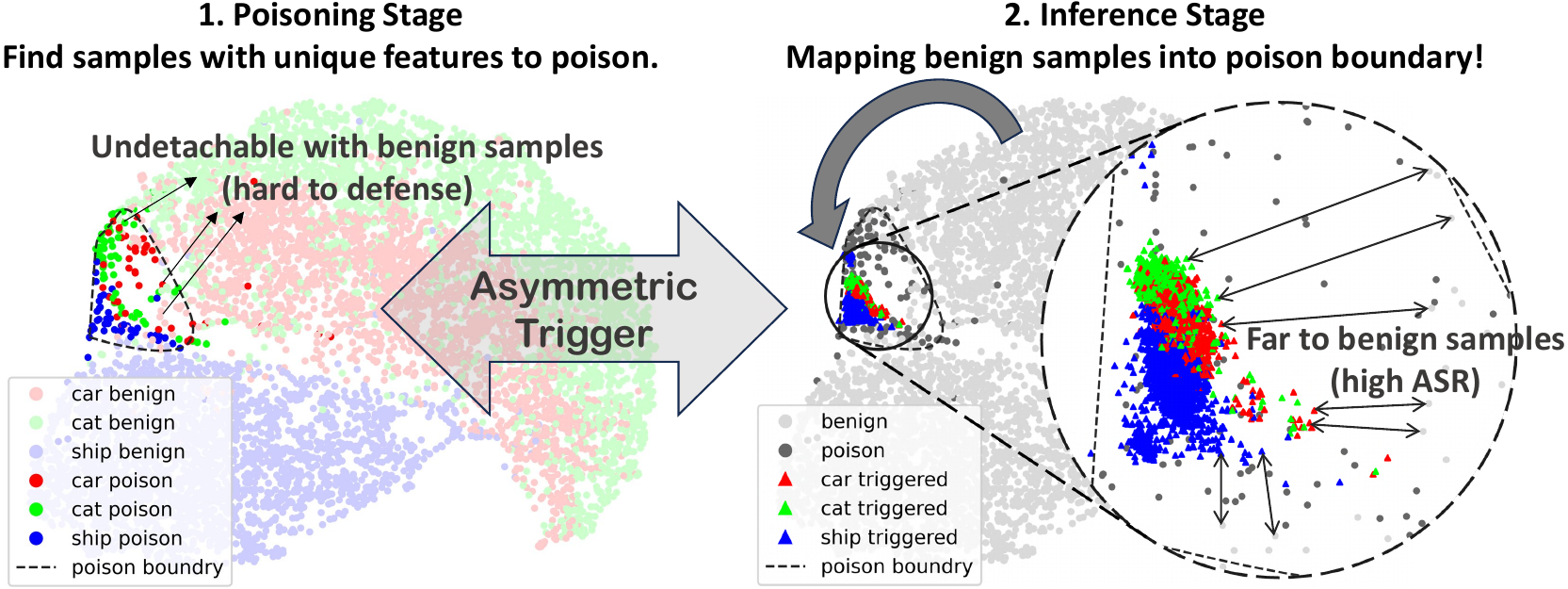}
    \caption{UMAP visualization of the latent space for three classes in CIFAR-10. \textbf{Left:} Poisoning Stage---the selected samples are not visibly separated from the remaining clean samples in this two-dimensional projection. \textbf{Right:} Inference Stage---triggered images are projected into a more concentrated region associated with the selected samples. UMAP is qualitative and does not establish distributional equality or impossibility of detection.}
    \label{fig:asymmetric}
    \vspace{-0.0cm}
\end{figure*}

We provide a visualization of this phenomenon in Figure~\ref{fig:asymmetric}, which illustrates how we select samples to poison and add triggers in the two stages from the perspective of the latent space.

In the poisoning stage, we use a score function to evaluate all samples and select those with the highest scores for poisoning. These selected images have varying scores and are not visibly separated from the remaining samples in the UMAP projection. This observation suggests, but does not prove, that simple outlier-based training-time defenses may have difficulty isolating them~\citep{qi2022revisiting}.

During the inference stage, we apply a trigger function to generate triggered inputs. In the visualization, these inputs occupy a more concentrated region than the selected training samples. Together with the measured ASR, this pattern is consistent with the asymmetric-trigger explanation, but the visualization alone does not establish a causal mechanism or guarantee robustness.

\section{Experimental Settings}

\label{ap:set}

\subsection{C-InfoGAN Settings}

In our model, the ground-truth label \(y\) is combined with the poison condition \(c\) and integrated into the image feature map through cross-attention mechanisms~\citep{vaswani2017attention} at each convolutional layer in the UNet structure. For all experiments, we apply batch normalization after most layers and use random seed 42 to facilitate reproducibility. The temperature of the Gumbel Softmax~\citep{jang2016categorical} is set to 0.5. The batch size is 256, with weight decay $10^{-5}$. We use Adam with betas 0.5 and 0.999 for 100 epochs on each dataset, with one discriminator update and one generator/recognition update per iteration. Following InfoGAN~\citep{chen2016infogan}, we find that the learning rate and information-loss weight materially affect the empirical training outcome. Their values for different structures are reported in Table~\ref{tab:hyper}. The reported hyperparameters are sufficient to reproduce our configurations but are not claimed to be unique or globally optimal. Each model is trained on an Nvidia A100 for at most two hours; convergence in this paper refers to empirical stabilization of the monitored losses and generated outputs, not a guarantee of reaching the global GAN equilibrium.

\renewcommand{\arraystretch}{0.9}
\begin{table}[ht]
\begin{center}
\begin{small}
\setlength{\tabcolsep}{2pt}
\begin{tabular}{lccc}
\toprule
Dataset & lr G & lr D & \(\lambda\) \\
\midrule
MNIST & 5e-4 & 1e-4 & 0.5 \\
CIFAR-10 & 4e-5 & 4e-5 & 0.25 \\
CIFAR-100 & 4e-4 & 2e-4 & 0.25 \\
GTSRB & 4e-5 & 4e-5 & 0.25 \\
ColorCIFAR10 & 4e-5 & 4e-5 & 0.25 \\
CelebA & 4e-4 & 4e-4 & 0.1 \\
VOC2012 & 4e-4 & 4e-4 & 0.1 \\
\bottomrule
\end{tabular}
\end{small}
\caption{Important hyperparameters setting in our experiment.}
\vspace{-0.0cm}
\label{tab:hyper}
\end{center}
\vskip -0.0in
\end{table}

\subsection{Victim Model Setting} 

Unless specified, we use PreActResNet18 as the default victim model, and 0.01 as the default poison rate. For the training victim model, SGD with momentum of 0.9 is used under batch size of 128 and weight decay of 0.0005. A cosine learning rate scheduler with an initial learning rate of 0.01 is also used for stable convergence. For CIFAR-10 and CIFAR-100, we use 100 epoch as the default setting. For simpler datasets like MNIST or GTSRB, we use 20 and 50 as default epochs for quicker testing. All experiments on \methodabb{} are carried out in an all-to-one fashion.

\subsection{Other Vision Task Setting}

\label{ap:universal}

\textbf{CIB \cite{chen2022cib} Details.} We carried out our multi-label experiment based on the official code of CIB~\citep{chen2022cib}. We find that CIB can be highly sensitive to various source classes. To provide a more statistically significant result, we systematically tested each possible source class within the training dataset, calculating both the mean and standard deviation. For CIB one-to-one setting, we considered all label combinations with proportion of 5±1\% as potential source classes.

\textbf{Dataset}. \noindent\textit{Image Regression}: We introduce ColorCIFAR-10, derived from CIFAR-10~\citep{krizhevsky2009learning}, with labels representing continuous features: hue, saturation, and illumination. Cyclic encoding is used for hue, resulting in four labels (sin hue, cos hue, saturation, illumination), each scaled to [-1, 1] with added Gaussian noise (\(\mathcal{N}(0, 0.1^2)\) and clipped to [-1, 1].\noindent\textit{Semantic Segmentation}: VOC2012~\citep{pascal-voc-2012} is used with a focus on samples with semantic segmentation annotations, totaling 2,330 training and 583 testing images.\noindent\textit{Multi-label Binary Classification}: CelebA~\citep{liu2015celeba} is utilized with five balanced and independent binary labels: Attractive, Mouth Slightly Open, High Cheekbones, Smiling, Wavy Hair.

\textbf{Architecture}.
In training InfoGAN, cross-attention is employed for class feature encoding in all tasks. For Image Regression and Multi-label Binary Classification, labels are directly fed to cross-attention without preprocessing. For Semantic Segmentation, label images are added to the UNet image channels, bypassing cross-attention. Victim models are trained using PreActResNet18 for Image Regression and Multi-label Binary Classification, and UNet for Semantic Segmentation. All models undergo 100 epochs of training with SGD, an initial learning rate of 0.01, weight decay of 0.0005, and a standard cosine scheduler.

\begin{figure}[h]
\centering
\includegraphics[width=0.63\linewidth]{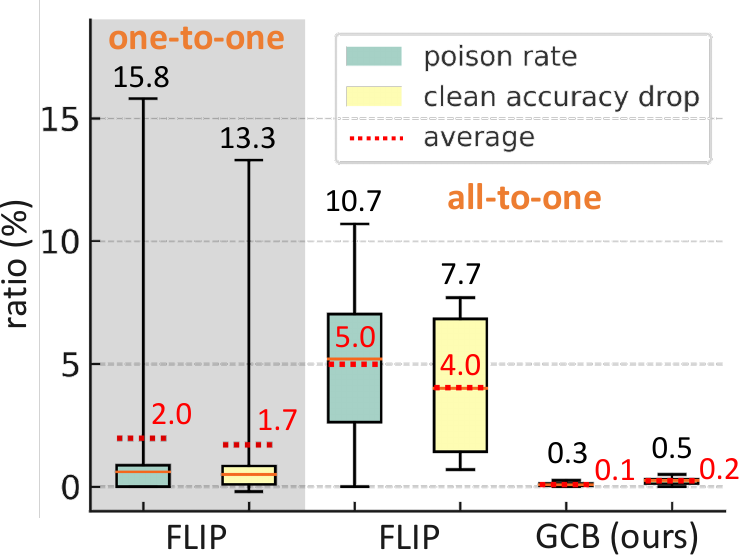}
% \vskip -0.1in
\caption{Box plot comparing the class-wise clean accuracy (CA) drop for clean-image backdoors on CIFAR-10. While the SOTA method, FLIP, has a low average CA drop (1.7\%), it can reach as high as 13.3\% for the source class. In contrast, our method exhibits a negligible CA drop across all classes.}
\label{fig:head_old}
% \vskip -0.2in
\end{figure}

\section{Discussion on FLIP \cite{jha2024flip}}
\label{ap:flip}
\textbf{FLIP Experiment Setup.} We carry out our experiment based on the official code of FLIP and BackdoorBench \cite{wu2024backdoorbench}. In FLIP, the source class is set to all classes and the target class to class 0, aiming for an all-to-one attack. Poisoned labels for all samples are generated using FLIP and then sent to BackdoorBench for a fair comparison.

\renewcommand{\arraystretch}{0.9}
\begin{table}[ht]
\centering
\begin{small}
\setlength{\tabcolsep}{2pt}
\begin{tabular}{lcccc}
\toprule
\multirow{2}{*}{Architecture} & \multicolumn{2}{c}{FLIP} & \multicolumn{2}{c}{\methodabb{}} \\
\cmidrule(lr){2-3} \cmidrule(lr){4-5}
 & CA & ASR & CA & ASR \\
\midrule
EfficientNet-B0  & 74.9\% &  3.8\% & 73.0\%  & 99.93\% \\
PreActResNet18   & 91.9\% & 86.3\% & 92.6\%  & 100.0\% \\
ResNet18         & 83.4\% &  4.9\% & 84.3\%  & 100.0\% \\
VGG19            & 88.6\% &  5.6\% & 89.5\%  & 100.0\% \\
ViT-B-16         & 95.2\% &  3.4\% & 94.5\%  & 100.0\% \\
\bottomrule
\end{tabular}
\end{small}
\caption{Comparison of FLIP and our attack across architectures. FLIP, trained on a PreActResNet18 surrogate model, performs well only when the victim model matches the surrogate. In contrast, our attack is effective across all architectures.}
\vskip -0.0in
\label{tab:flip_arc}
% \vskip -0.1in
\end{table}

\paragraph{Weakness of FLIP.} There are two major weakness of FLIP. The first one is that it is extremely sensitive to different architectures of victim models, as shown in Table \ref{tab:flip_arc}. FLIP and our proposed method, \methodabb{}, were tested on different victim model architectures using the CIFAR-10 dataset. FLIP uses 3\% poison rate, and \methodabb{} uses 1\% poison rate. Since FLIP's expert model defaults to PreActResNet34, it performs well on similar architectures like PreActResNet18. However, FLIP fails to poison all other architectures, making it impractical in real scenarios because adversaries are unlikely to anticipate the victim model's structure.

The second major weakness is that FLIP suffers from high CA drop in both one-to-one \footnote{one source class and one target class} scenario and all-to-one \footnote{all except the target class are regarded as the source class} scenario. For the one-to-one scenario, although the mean CA drop seems low, it is an averaged result across all classes. When considering the CA drop of the source class, it will surprisingly become 13. 3\%. For the all-to-one scenario, the CA drop is consistently as large as 4\%.

\section{ASR vs Poison Rates}

\label{ap:poison_rate}

We provide the ASR versus poison rate as an additional result, particularly for positioning the CIBA attack on CIFAR-10. Since they do not release their code, we can only replicate their results here rather than directly compare them in the same benchmark as other attack methods in Fig. \ref{fig:main_acc_asr}.

\begin{figure}[ht]
\centering
\begin{minipage}[t]{0.48\linewidth}
  \centering
  \includegraphics[width=1.0\textwidth]{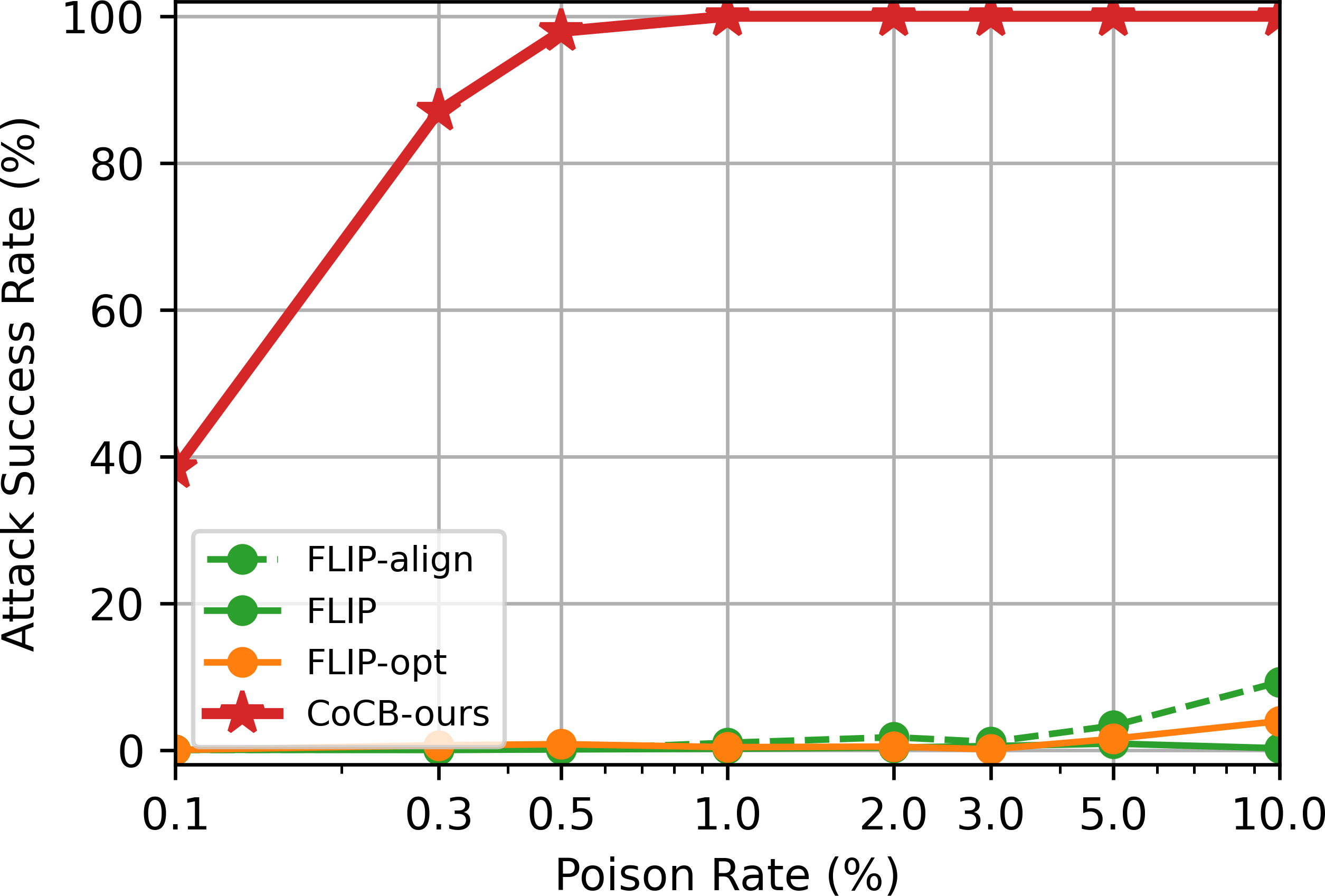}
  \caption*{(a) MNIST}
\end{minipage}
\hspace{-3mm}
\begin{minipage}[t]{0.48\linewidth}
  \centering
  \includegraphics[width=1.0\textwidth]{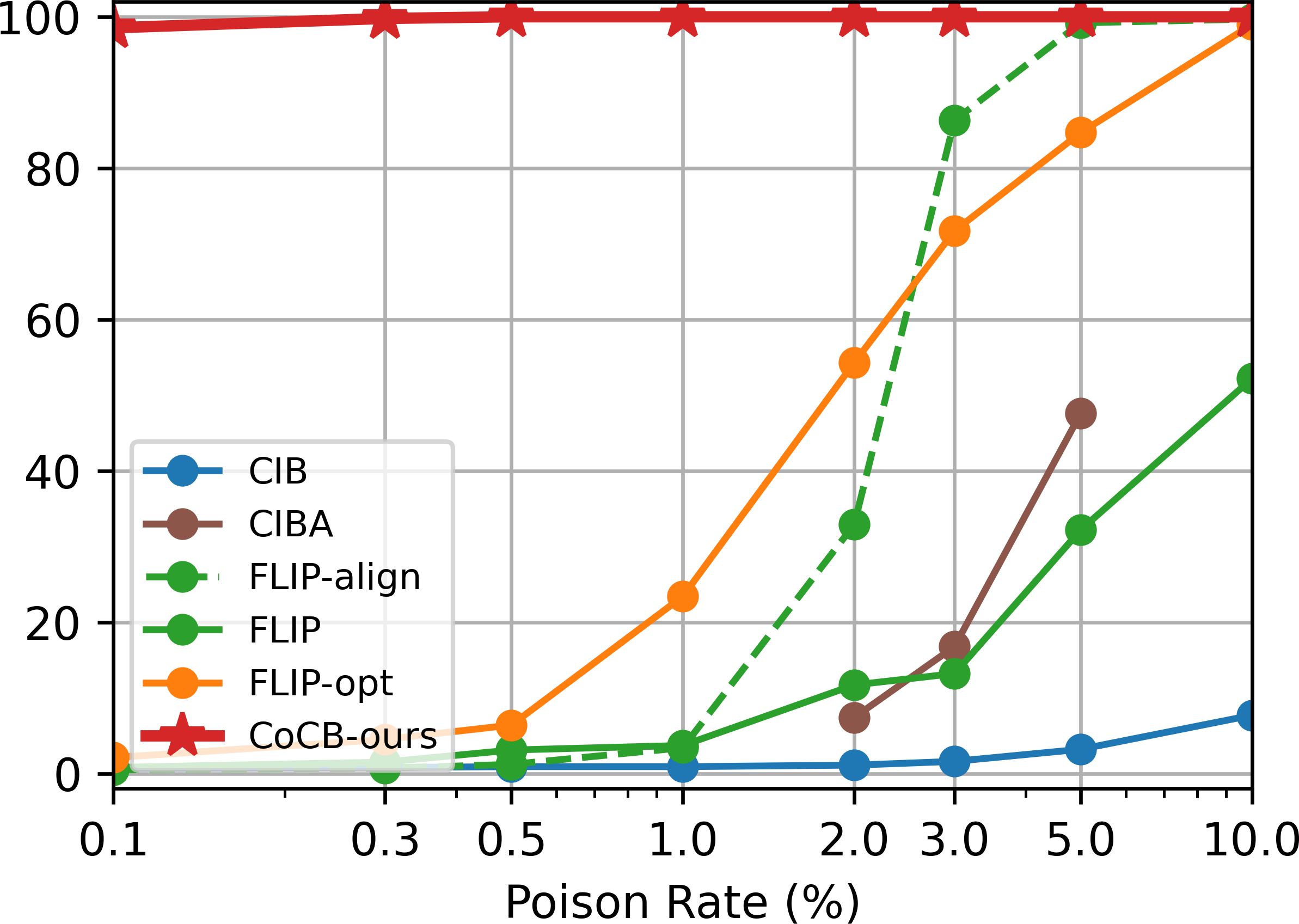}
  \caption*{(b) CIFAR-10}
  \label{fig:cifar10_asr_pr}
\end{minipage}
\hspace{-3mm}
\begin{minipage}[t]{0.48\linewidth}
  \centering
  \includegraphics[width=1.0\textwidth]{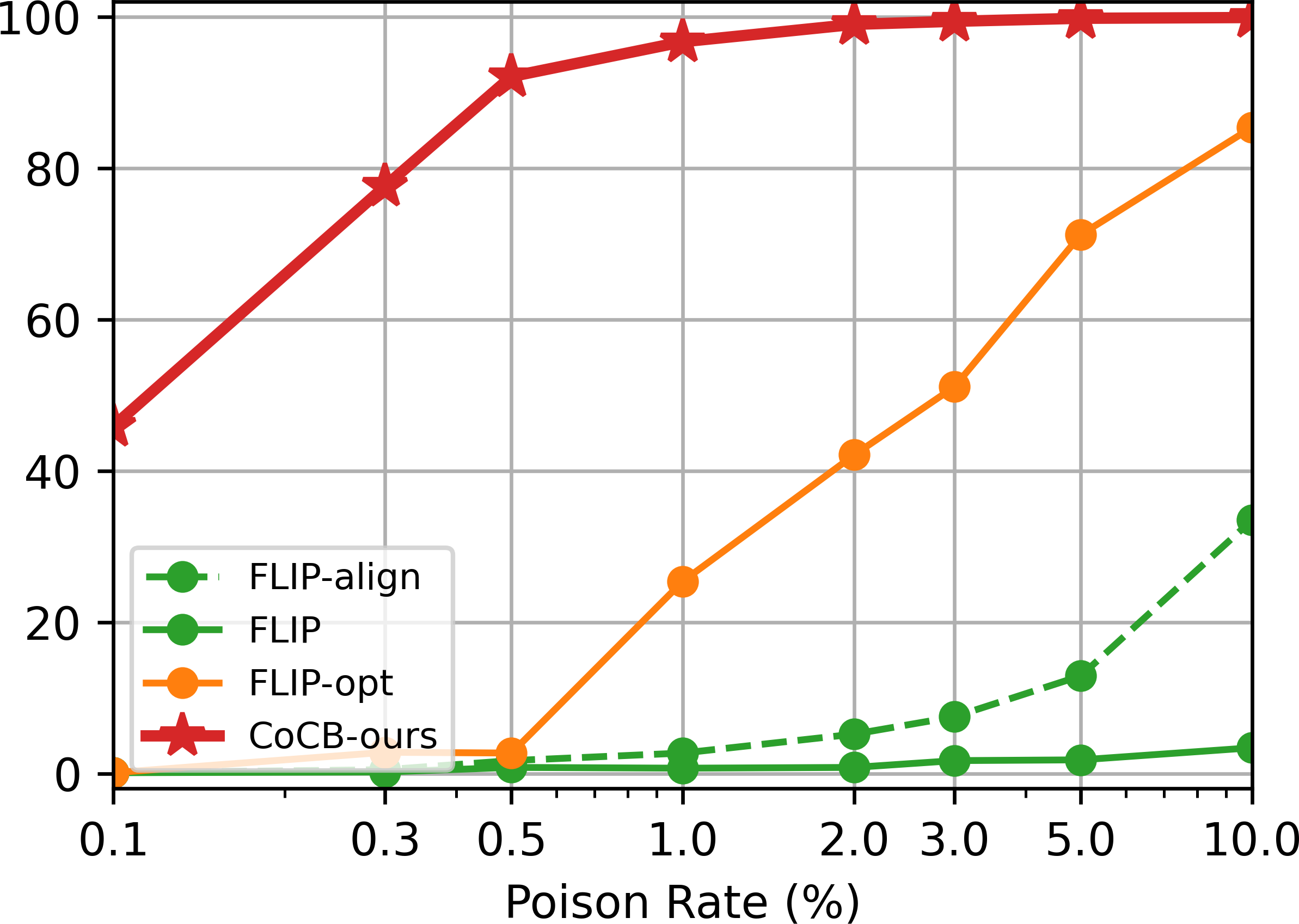}
  \caption*{(c) CIFAR-100}
\end{minipage}
\hspace{-3mm}
\begin{minipage}[t]{0.48\linewidth}
  \centering
  \includegraphics[width=1.0\textwidth]{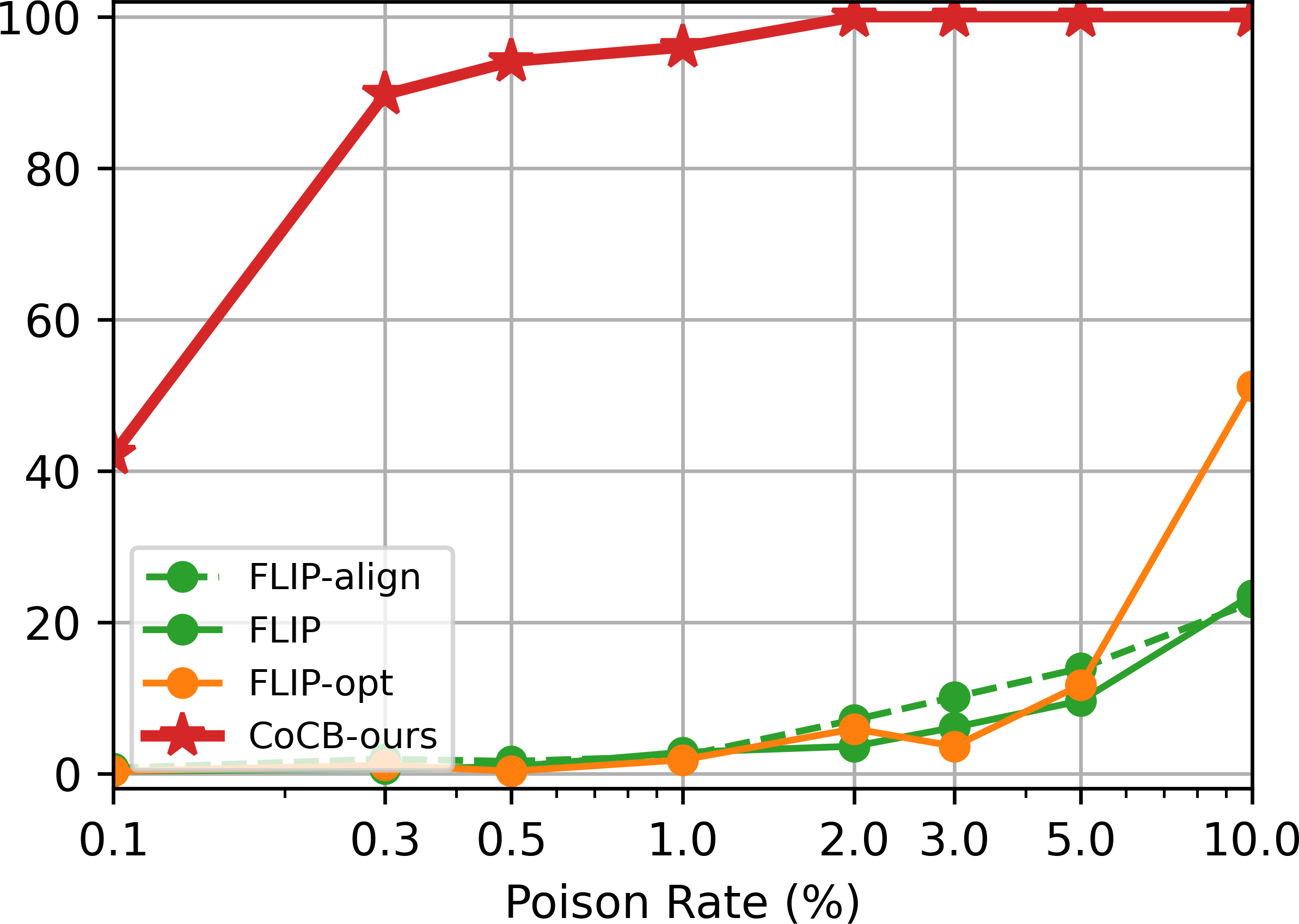}
  \caption*{(d) GTSRB}
\end{minipage}
\vskip 0.05in
\caption{ASR vs. poison rates for different attack methods.}
\label{fig:asr_poison_rates}
\end{figure}

\section{\methodabb{}'s Robustness to Dataset Size}

\begin{figure}[t]
  \centering
  \begin{minipage}{0.48\linewidth}
    \centering
    \includegraphics[width=\linewidth]{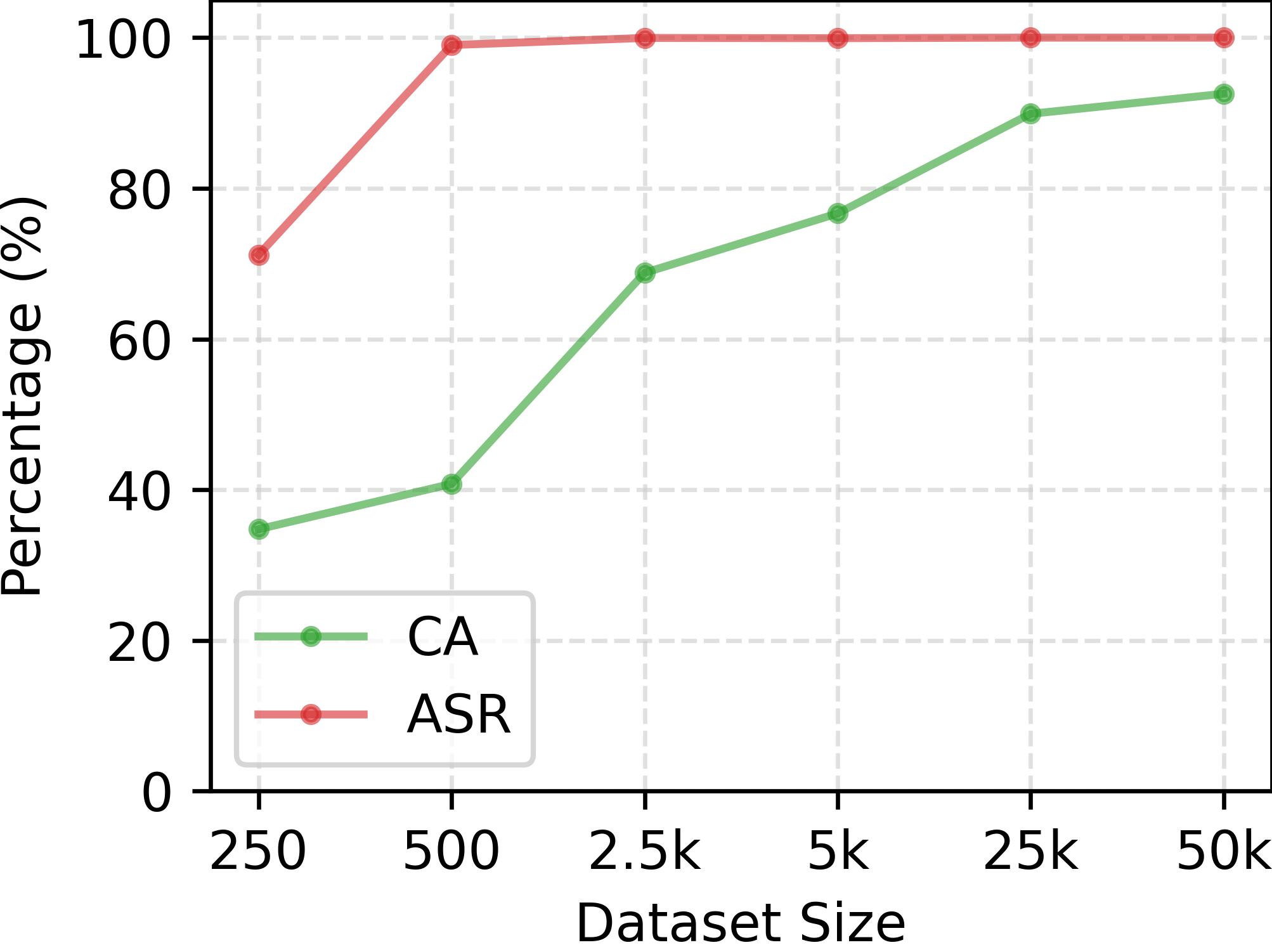}
    \vskip -0.05in
    \caption*{(a) CIFAR-10}
    \label{fig:cifar10_size}
  \end{minipage}
  \hfill
  \begin{minipage}{0.48\linewidth}
    \centering
    \includegraphics[width=\linewidth]{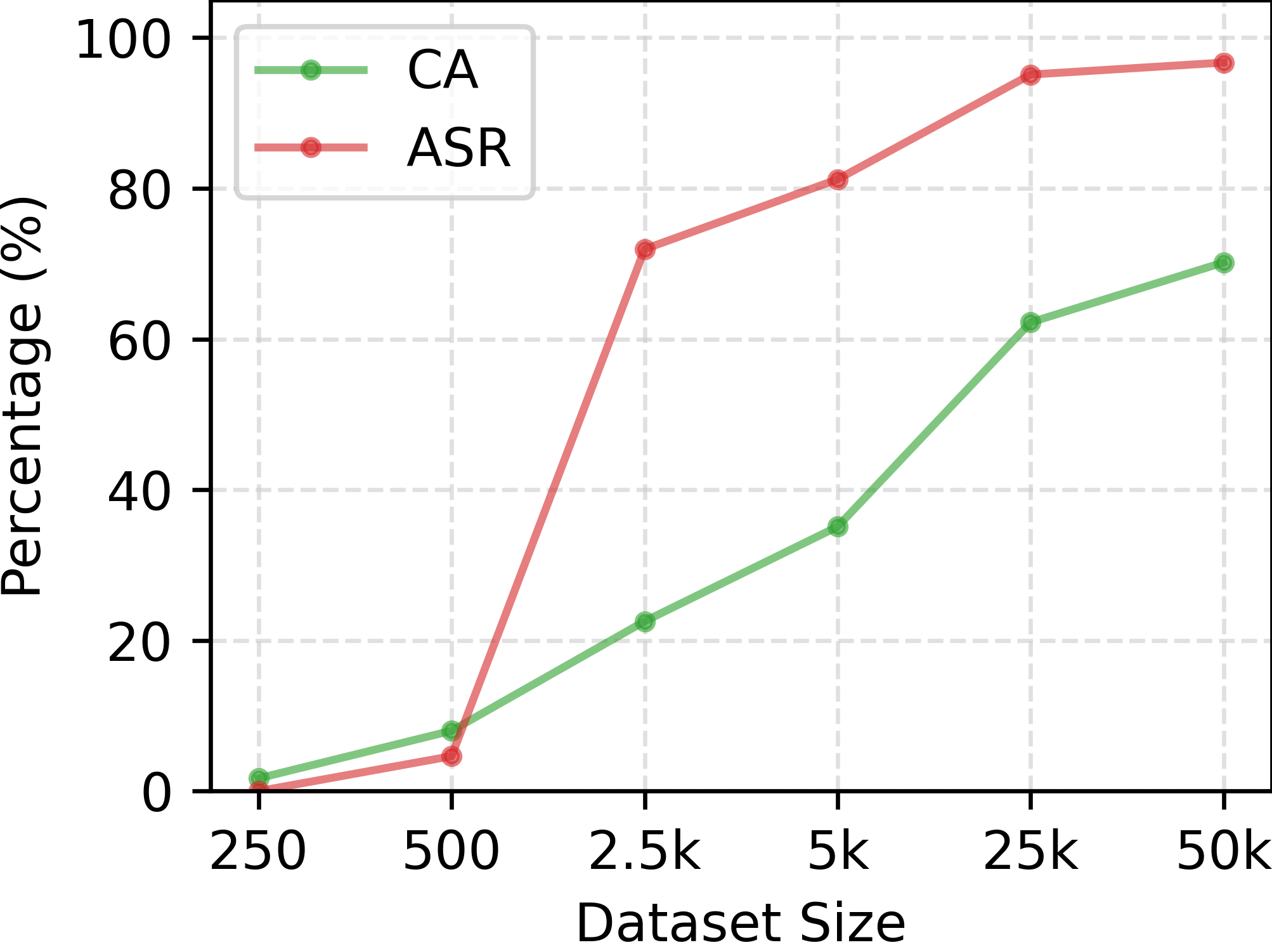}
    \vskip -0.05in
    \caption*{(b) CIFAR-100}
    \label{fig:cifar100_size}
  \end{minipage}
  \vskip -0.1in
  \caption{Results under different dataset sizes for CIFAR-10/100. Both CIFAR-10/100 have 50,000 images in total. Results show that \methodabb{} remains effective in most cases. When \methodabb{} lose effectiveness, the clean accuracy of each method also significantly drops to less than 10\%.}
  % \vskip -0.15in
  \label{fig:size}
\end{figure}

We evaluate the robustness of \methodabb{} under varying training dataset sizes on CIFAR-10 and CIFAR-100. As shown in Figure~\ref{fig:size}, \methodabb{} maintains high attack success rates (ASR) even when the dataset size is reduced, demonstrating its stability and efficiency with limited data. While performance slightly decreases in extremely small datasets, this degradation is consistent with a significant drop in clean accuracy (CA), indicating that \methodabb{}'s effectiveness is not disproportionately affected by dataset size.

\section{\methodabb{}'s Robustness to Architectures}

\label{ap:archi}

We evaluated our attack across PreActResNet18\citep{he2016identity}, EfficientNet B0\citep{tan2019efficientnet}, VGG11\citep{simonyan2015vgg}, and ViT-B-16\citep{dosovitskiy2020vit}. As Table~\ref{tab:architecture_performance} shows, the attack achieves high ASR across these tested architectures. This supports cross-architecture robustness within the evaluated set, not a universal model-agnostic guarantee. PreActResNet18, maintaining good CA while reaching the lowest ASR, is chosen as the default architecture for the other experiments.

\renewcommand{\arraystretch}{0.9}
\begin{table}[ht]
\begin{center}
\begin{small}
\setlength{\tabcolsep}{2pt}
\begin{tabular}{lcccccccc}
\toprule
Architecture & \multicolumn{2}{c}{PreActRN18} & \multicolumn{2}{c}{VGG11} & \multicolumn{2}{c}{EffNet B0} & \multicolumn{2}{c}{ViT-B-16} \\
\cmidrule(lr){1-1} \cmidrule(lr){2-3} \cmidrule(lr){4-5} \cmidrule(lr){6-7} \cmidrule(lr){8-9}
Dataset & CA & ASR & CA & ASR & CA & ASR & CA & ASR \\
\midrule
MNIST & 98.5 & 100 & 98.4 & 100 & 98.6 & 100 & 98.6 & 100 \\
CIFAR10 & 92.6 & 100 & 88.3 & 100 & 73.0 & 99.9 & 94.5 & 100 \\
CIFAR100 & 70.1 & 96.7 & 58.6 & 93.0 & 52.9 & 93.8 & 84.1 & 95.3 \\
GTSRB & 95.9 & 96.0 & 93.7 & 96.2 & 85.0 & 91.2 & 98.0 & 95.2 \\
Average & 89.3 & 98.2 & 84.8 & 97.3 & 77.4 & 96.2 & 93.8 & 97.6 \\
\bottomrule
\end{tabular}
\end{small}
\vskip -0.1in
\caption{CA and ASR of different architectures of poison rate 1\%. Our model shows high ASR across all tested datasets and models.}
\label{tab:architecture_performance}
\vspace{-0.0cm}
\end{center}
\vskip -0.0in
\end{table}

\section{Poisoned Sample Visualization}

\subsection{Poison Samples in Image Classification}

\begin{figure*}[t]
  \centering
  \begin{minipage}[b]{0.2\textwidth}
    \centering
    \includegraphics[width=1.0\textwidth]{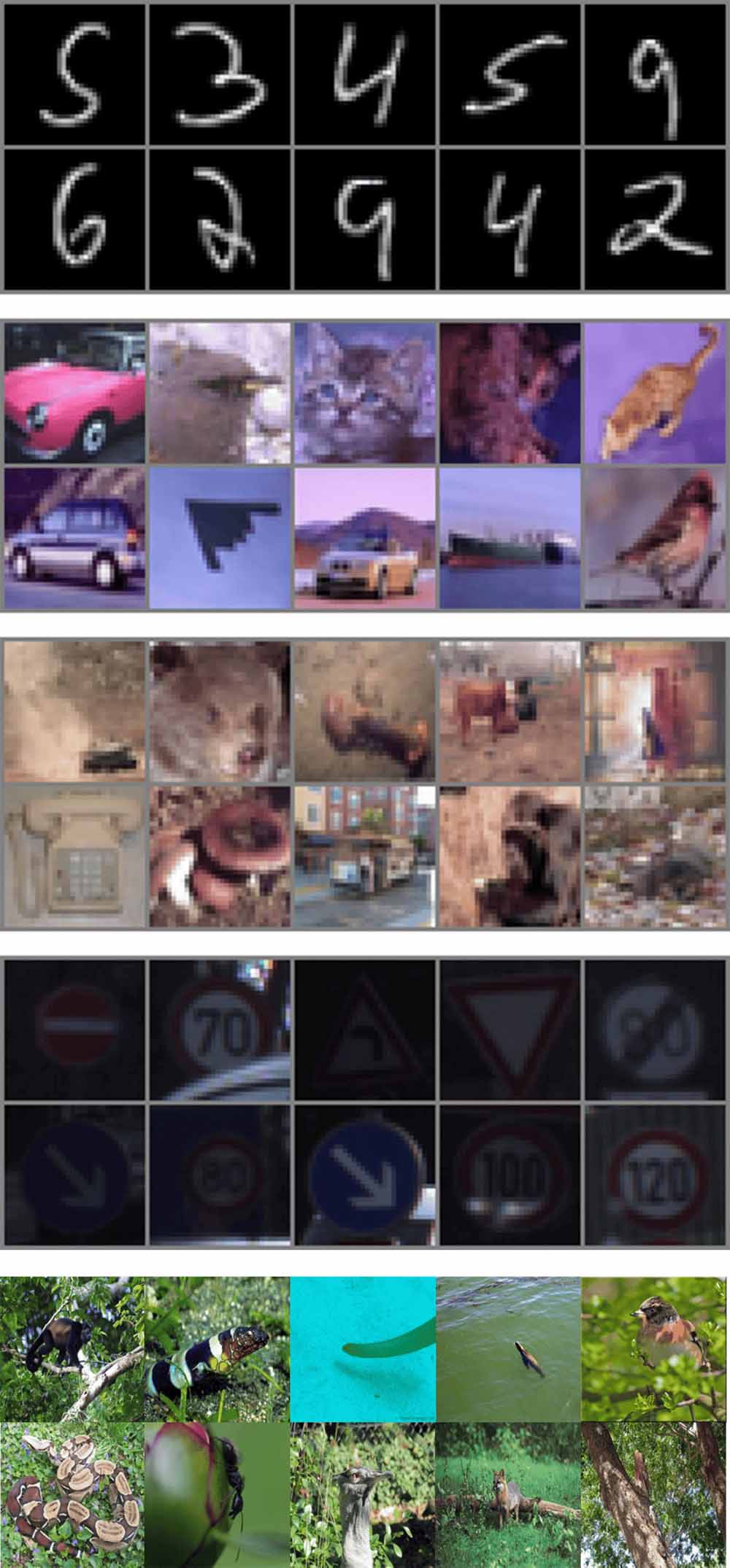}
    \caption*{(a) Selected train images}
    \label{fig:main_select}
  \end{minipage}
  \hfill
  \begin{minipage}[b]{0.2\textwidth}
    \centering
    \includegraphics[width=1.0\textwidth]{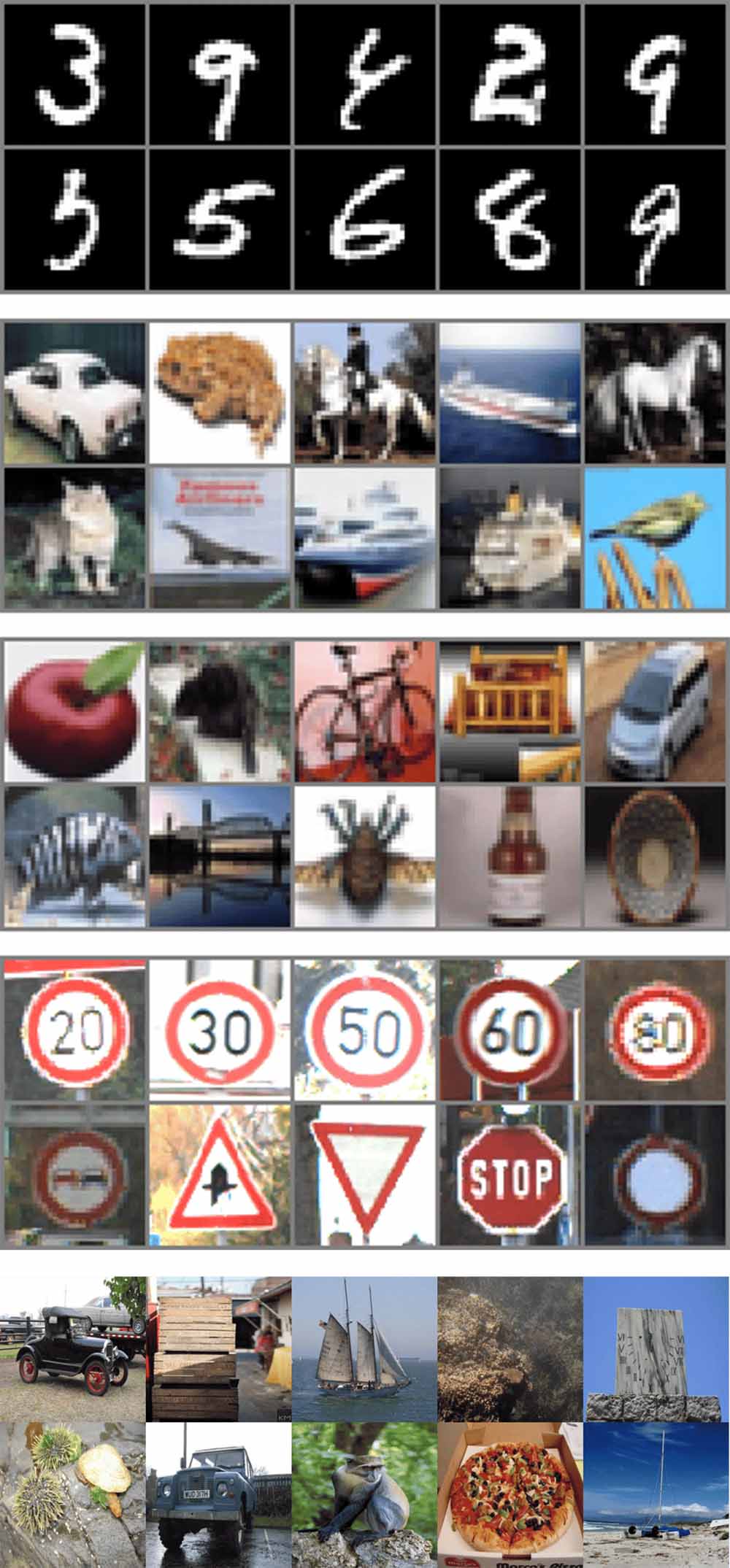}
    \caption*{(b) Clean test images}
    \label{fig:main_clean}
  \end{minipage}
  \hfill
  \begin{minipage}[b]{0.2\textwidth}
    \centering
    \includegraphics[width=1.0\textwidth]{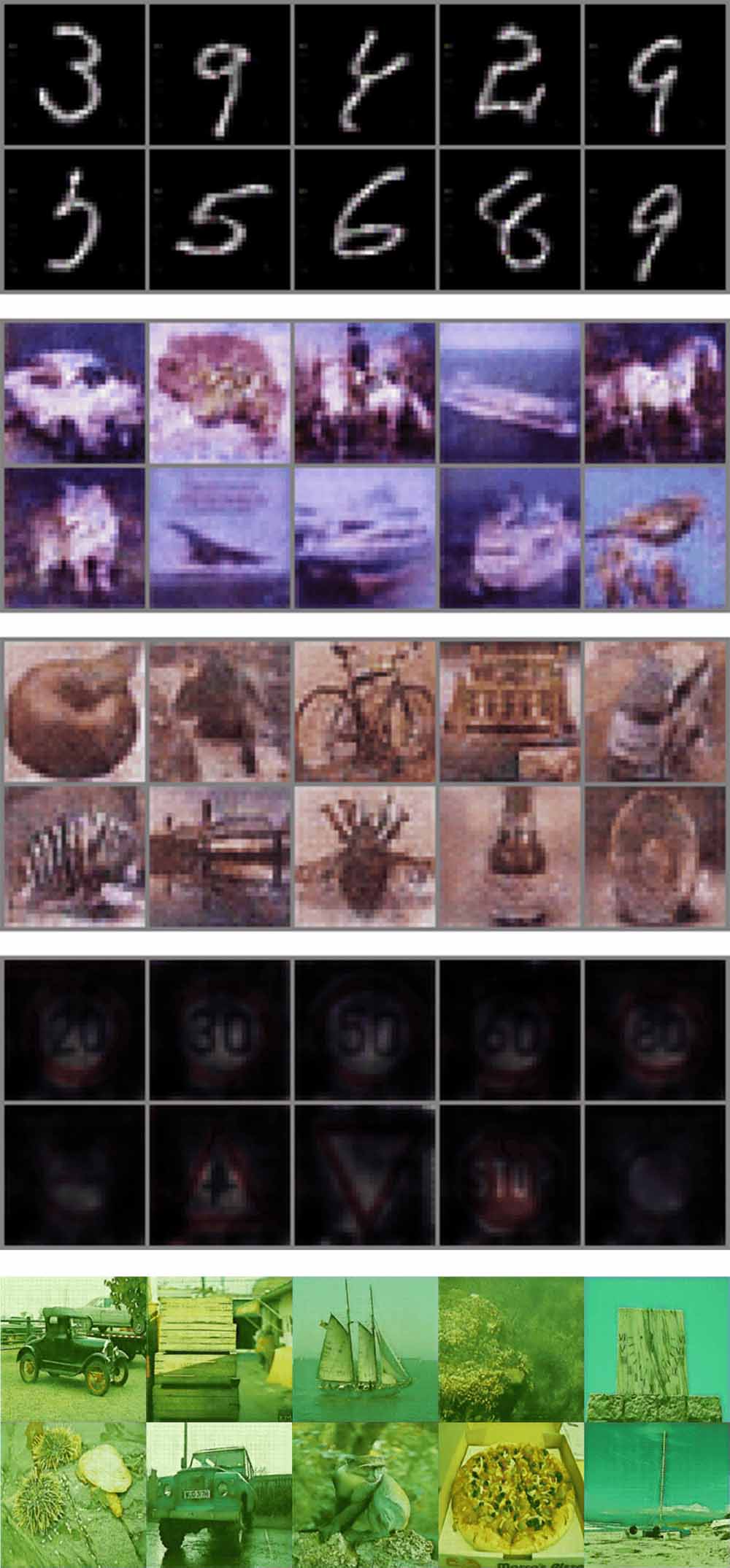}
    \caption*{(c) Triggered test images}
    \label{fig:main_trigger}
  \end{minipage}
  \hfill
  \begin{minipage}[b]{0.2\textwidth}
    \centering
    \includegraphics[width=1.0\textwidth]{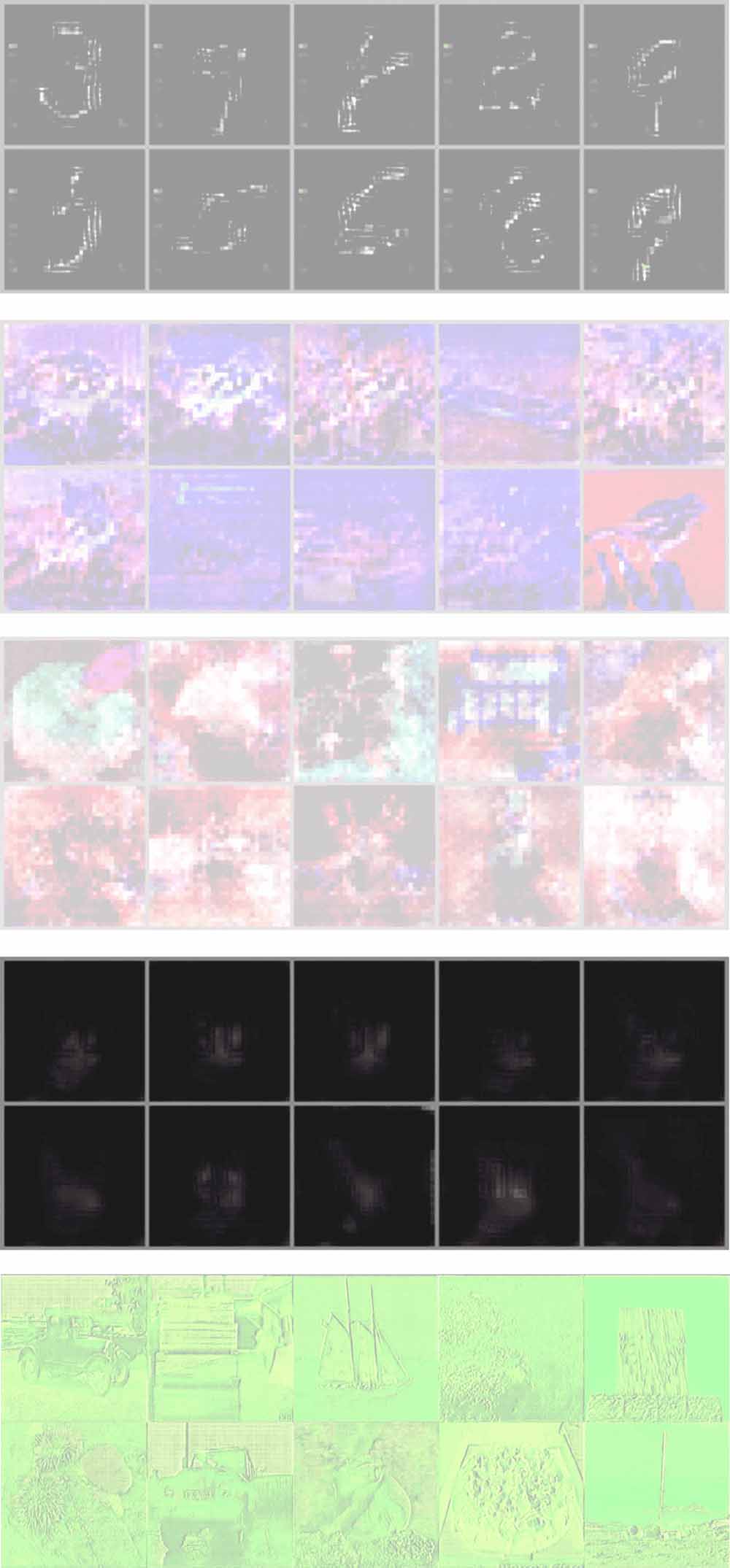}
    \caption*{(d) Trigger in differences}
    \label{fig:main_diff}
  \end{minipage}
  \vskip -0.05in
  \caption{Sample images from MNIST, CIFAR-10, CIFAR-100, GTSRB, and ImageNet-1K. The selected training images are unmodified, representing a subset of clean images.}
  \vskip -0.1in
  \label{fig:images_set}
\end{figure*}

As shown in Fig.~\ref{fig:images_set}, C-InfoGAN predominantly identified color-related variation for all evaluated datasets except MNIST, occasionally combined with positional or global-contrast features. For MNIST, the learned variation involved properties such as digit weight or thickness. Together with the label-preservation results, these observations suggest that the learned transformations often retain class-relevant information; they do not establish that the trigger features are statistically independent of class labels.

\subsection{Poison Samples in Other Tasks}

\label{ap:defense_backdoorbench}

\begin{figure}[ht]
\centering

\begin{minipage}[t]{\columnwidth}
    \centering
    \includegraphics[width=1.0\textwidth]{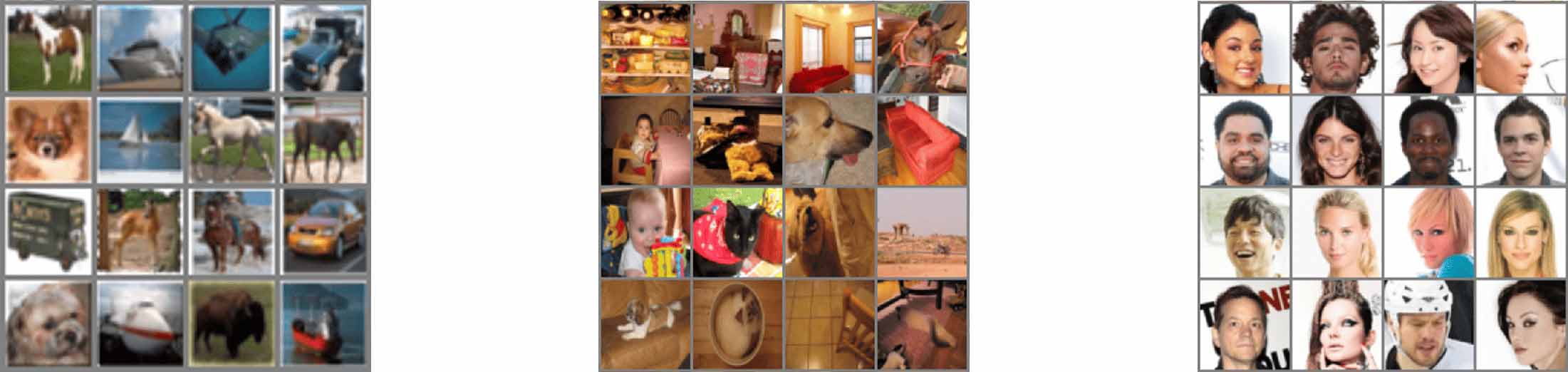}
    \caption*{(a) Selected Training Images}
    \label{fig:universal_select}
\end{minipage}
\vskip 0.05in

\begin{minipage}[t]{\columnwidth}
    \centering
    \includegraphics[width=1.0\textwidth]{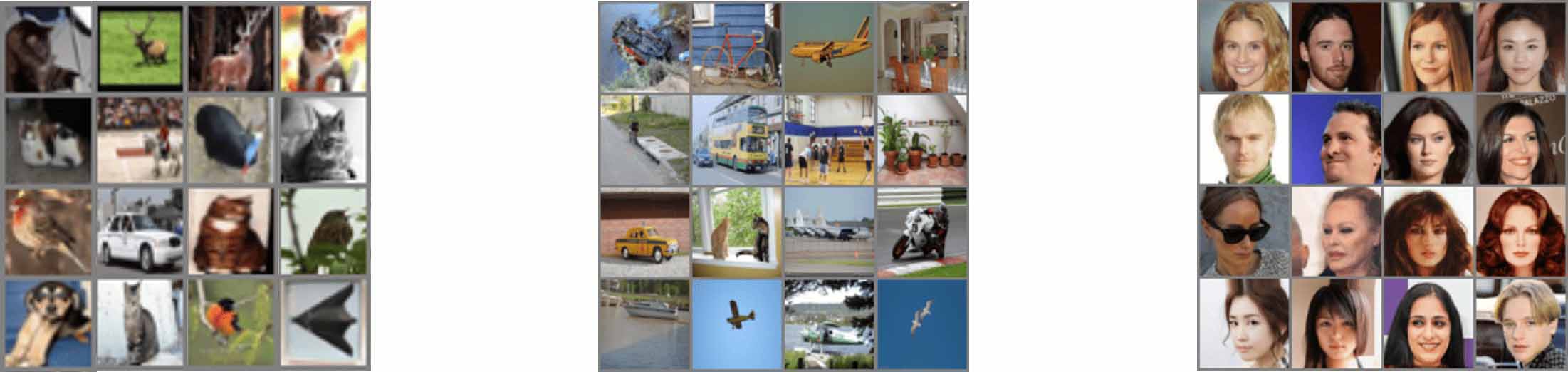}
    \caption*{(b) Clean Testing Images}
    \label{fig:universal_clean}
\end{minipage}
\vskip 0.05in

\begin{minipage}[t]{\columnwidth}
    \centering
    \includegraphics[width=1.0\textwidth]{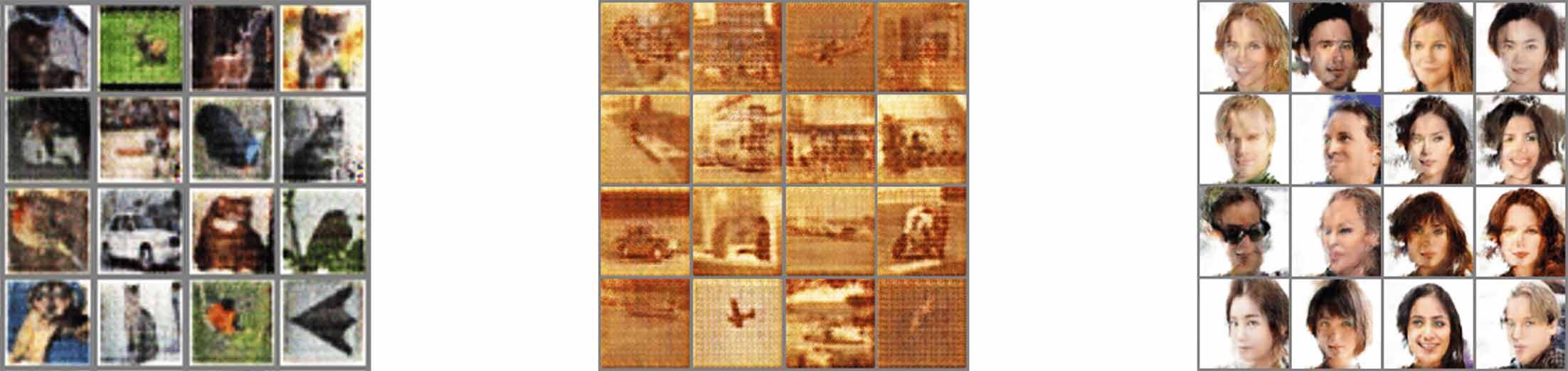}
    \caption*{(c) Triggered Testing Images}
    \label{fig:universal_trigger}
\end{minipage}

% \vskip 0.1in
\caption{Image samples from different datasets. From top to bottom are ColorCIFAR10, VOC2012, CelebA respectively. The selected training images are unmodified, representing a subset of clean images.}
\label{fig:universal_img}
\end{figure}

Analyzing the triggered features generated for each dataset reveals interesting distinctions. As shown in Fig. \ref{fig:universal_img}, VOC2012 retains color features similar to image classification tasks. In contrast, CelebA, where color might be label-relevant, learns background color as the triggered feature. Most notably, ColorCIFAR-10 selects image borders as the trigger, attributed to the prevalence of bordered images in the dataset. This suggests that InfoGAN can be directed to specific features by incorporating relevant priors, thereby avoiding unwanted feature learning.

\section{Robustness Measure}

\label{sec:robustness_measure}

Although clean-image backdoors do not poison images during training, triggers are still used to activate the backdoor at test time. To evaluate test-time preprocessing defenses, we assess several prominent techniques in our experiments:
\begin{enumerate}
    \item \textbf{\textit{JPEG Compression}}~\citep{xue2023compression}: Applies JPEG compression to all testing images at 75\% quality. 
    \item \textbf{\textit{Gaussian Smoothing}}~\citep{xue2023compression}: Applies Gaussian blur with a kernel size of 3 pixels to each image.
    \item \textbf{\textit{Color Shift}}~\citep{jiang2023color}: Introduces a random color space shift between -0.1 and 0.1, specifically targeting color-based backdoors in datasets like CIFAR-10.
    \item \textbf{\textit{Color Shrink}}~\citep{li2020shrinkpad}: Reduces the color-bit depth of test images to 5 bits.
    \item \textbf{\textit{Affine Transformation}}~\citep{qiu2021deepsweep}: Applies stochastic affine transformation to each test image as a defense.
\end{enumerate}

Table~\ref{tab:corruption_comparison} presents the results of these preprocessing defenses on the CIFAR-10 and CIFAR-100 datasets. Additive trigger-based attacks, such as BadNets, exhibit reduced ASR when subjected to image transformations. Natural trigger-based attacks like SIG and FLIP remain robust against most defenses but are compromised by image compression. In contrast, our \methodabb{} attack maintains nearly 100\% ASR across all preprocessing defenses on CIFAR-10 and is also effective on CIFAR-100. This resilience is attributed to the use of a dominant semantic feature as the trigger, which is more robust than fragile and intricate label features.

\begin{table}
\centering
\begin{small}
\begin{tblr}{
  width = 1.0\linewidth,
  rowsep = 0.4pt,
  colsep = 0.4pt,
  colspec = {Q[202]Q[96]Q[92]Q[85]Q[85]Q[94]Q[94]Q[85]Q[85]},
  cells = {c},
  cell{1}{2} = {c=4}{},
  cell{1}{6} = {c=4}{},
  cell{2}{2} = {c=2}{},
  cell{2}{4} = {c=2}{},
  cell{2}{6} = {c=2}{},
  cell{2}{8} = {c=2}{},
  hline{1,10} = {-}{0.1em},
  hline{4} = {-}{0.05em},
}
Dataset→   & CIFAR10    &     &          &      & CIFAR100   &      &          &      \\
Attack→    & GCB (ours) &     & FLIP-opt &      & GCB (ours) &      & FLIP-opt &      \\
Method↓    & CA         & ASR & CA       & ASR  & CA         & ASR  & CA       & ASR  \\
No Defense & 92.6       & 100 & 90.5     & 78.2 & 70.2       & 96.7 & 67.1     & 64.9 \\
SPL        & 91.9       & 100 & 84.5     & 9.4  & 67.2       & 78.2 & 56.1     & 48.8 \\
PRL        & 89.7       & 100 & 88.5     & 82.8 & 66.8       & 87.6 & 66.2     & 47.9 \\
BootStrap  & 88.4       & 100 & 92.2     & 14.3 & 57.6       & 93.9 & 59.9     & 94.4 \\
DivideMix  & 92.1       & 100 & 89.1     & 24.4 & 73.4       & 86.7 & 65.4     & 64.5 \\
MentorMix  & 89.9       & 100 & 89.7     & 35.6 & 69.0       & 92.7 & 63.9     & 80.3 
\end{tblr}
\end{small}
\vskip -0.1in
\caption{Comparison of GCB and FLIP-opt against all noisy training mitigation on CIFAR. We use 3\% poison rate in FLIP-opt to ensure high ASR. Some defense methods (e.g., SPL) can neutralize FLIP-opt but remain ineffective against our GCB attack.}
\vskip -0.0in
\label{tab:flip_noisy}
\end{table}

\begin{table*}
\centering
\begin{small}
\begin{tblr}{
  width = 0.9\textwidth,
  rowsep = 0.2pt,
  colsep = 0.2pt,
  colspec = {Q[120]Q[140]Q[63]Q[63]Q[63]Q[63]Q[63]Q[63]Q[63]Q[63]Q[63]Q[63]Q[63]Q[63]Q[75]},
  cells = {c},
  hline{2} = {3,5,7,9,11,13}{l},
  hline{2} = {4,6,8,10,12,14}{r},
  hline{1,22} = {-}{0.1em},
  hline{3,13} = {-}{0.05em},
  cell{1}{1} = {r=2}{},
  cell{1}{3} = {c=2}{0.09\linewidth},
  cell{1}{5} = {c=2}{0.09\linewidth},
  cell{1}{7} = {c=2}{0.09\linewidth},
  cell{1}{9} = {c=2}{0.09\linewidth},
  cell{1}{11} = {c=2}{0.1\linewidth},
  cell{1}{13} = {c=2}{0.09\linewidth},
  cell{3}{1} = {r=10}{},
  cell{13}{1} = {r=9}{},
  cell{3-12,13-22}{4,6,8,10,12,14} = {bg = customred},
  cell{5,6,9,15,19}{6} = {bg = customgreen},
  cell{3,4,5,6,9,11}{8} = {bg = customgreen},
  cell{10,13,17,18,19,20}{12} = {bg = customgreen},
  cell{10,15}{14} = {bg = customgreen},
}
Dataset↓  & Corruption→ & No Defense & {}      & JPEG & {}      & Smoothing &  {}     & Color Shift &  {}    & Color Shrink &  {}     & Affine &  {}     & Average \\
          & Attack↓     & ACC        & ASR   & ACC  & ASR   & ACC       & ASR   & ACC         & ASR  & ACC          & ASR   & ACC    & ASR   & ASR     \\
CIFAR-10  & BadNet       & 93.2       & 73.8  & 79.1 & 50.8  & 39.9      & 3.3   & 81.9        & 53.1 & 85.4         & 40.7  & 79.7   & 49.8  & 45.2    \\
          & Blend       & 93.7       & 94.1  & 79.0 & 66.3  & 42.6      & 2.6   & 86.3        & 86.9 & 86.1         & 85.5  & 86.4   & 82.3  & 69.6    \\
          & BPP          & 91.9       & 76.4  & 81.7 & 2.0   & 47.1      & 2.0   & 84.9        & 78.9 & 41.7         & 74.8  & 88.6   & 84.3  & 53.1    \\
          & IA           & 90.5       & 79.6  & 75.3 & 7.2   & 33.0      & 2.6   & 78.7        & 75.1 & 81.4         & 83.8  & 86.5   & 65.9  & 52.4    \\
          & LF           & 93.5       & 86.1  & 79.6 & 86.7  & 39.5      & 88.0  & 85.3        & 78.2 & 86.2         & 78.2  & 81.6   & 89.5  & 84.5    \\
          & SIG         & 93.7       & 80.4  & 79.8 & 89.8  & 43.9      & 70.1  & 86.8        & 74.3 & 86.1         & 66.0  & 85.9   & 60.0  & 73.5    \\
          & SSBA         & 93.4       & 99.7  & 78.9 & 1.7   & 37.4      & 18.5  & 86.2        & 94.2 & 81.6         & 94.4  & 88.5   & 87.9  & 66.1    \\
          & WaNet       & 91.1       & 72.0  & 60.4 & 50.6  & 17.9      & 58.7  & 84.6        & 60.0 & 83.1         & 1.2   & 88.0   & 19.8  & 43.7    \\
          & FLIP         & 91.9       & 86.3  & 72.5 & 86.9  & 35.9      & 9.3   & 83.7        & 83.8 & 82.3         & 67.3  & 78.8   & 72.1  & 67.6    \\
          & GCB (ours)  & 92.6       & \textbf{100.0} & 77.6 & \textbf{100.0} & 40.7 & \textbf{100.0} & 84.4 & \textbf{98.4} & 84.9 & \textbf{100.0} & 84.2 & \textbf{100.0} & \textbf{99.7} \\
CIFAR-100 & BadNets      & 70.7       & 35.6  & 56.3 & 33.5  & 70.2      & 31.7  & 63.6        & 23.0 & 66.2         & 4.0   & 67.7   & 20.1  & 24.7    \\
          & Blended      & 70.9       & 91.5  & 56.2 & 72.0  & 70.5      & 90.1  & 64.2        & 84.5 & 65.5         & 59.7  & 68.5   & 86.6  & 80.7    \\
          & BPP          & 65.0       & 66.1  & 56.7 & 0.1   & 64.6      & 42.5  & 59.2        & 71.1 & 63.6         & 62.7  & 62.8   & 0.5   & 40.5    \\
          & IA           & 65.3       & 78.8 & 50.0 & 60.3 & 64.5      & 78.5  & 57.1        & 85.2 & 59.2         & \textbf{87.5}  & 63.7   & 77.2  & 77.9    \\
          & LF           & 70.0       & 38.9  & 55.8 & 54.1  & 69.9      & 37.5  & 63.9        & 33.5 & 66.1         & 3.2   & 66.1   & 37.8  & 34.2    \\
          & SIG          & 70.4       & 77.7  & 53.5 & 80.9  & 24.1      & 91.5  & 64.3        & 67.3 & 65.6         & 13.7  & 64.9   & 91.1  & 70.4    \\
          & SSBA         & 70.7       & \textbf{98.8} & 52.7 & 4.0   & 24.6      & 90.6  & 64.6        & 91.8 & 65.1         & 4.0   & 61.2   & 97.2  & 64.4    \\
          & WaNet        & 63.7       & 92.7  & 38.7 & 77.4  & 2.0       & \textbf{98.3} & 58.5 & 82.9 & 60.8         & 0.1   & 13.1   & \textbf{97.7} & 74.9    \\
          & GCB (ours)  & 70.1       & 96.7  & 57.4 & \textbf{96.7} & 25.3 & 67.1  & 63.7        & \textbf{95.7} & 30.4 & 72.8  & 54.2   & 86.1  & \textbf{85.9}    
\end{tblr}
\end{small}
\caption{Comparison of different attack methods against common image corruptions.}
\vskip -0.0in
\label{tab:corruption_comparison}
\end{table*}

\section{\methodabb{}'s Resilience}

\subsection{\methodabb{}'s Resilience on Clean-Data-Based Defenses}
In Table \ref{tab:defenses} in our paper, we already show poison-data-based backdoor defenses. There is also one branch of backdoor defense called clean-data-based backdoor defenses, where they assume the defender to have additional private in-distribution clean data (typically 5\%~10\% of the whole dataset). To evaluate these methods, we show the results in Table \ref{tab:defense_effectiveness_cifar10_clean_data_based}. As shown, our \methodabb{} is defended by FT-SAM and SAU, which are very recent defenses in 2023 and 2024, respectively. As a result, the conclusion here is: although not designed to be, our method can have good resilience to most of the defense methods. However, new-emerging defenses, especially clean-data-based defenses, are effective in defending our attacks.

\begin{table*}
\centering
\begin{small}
\begin{tblr}{
  width = 0.75\textwidth,
  rowsep = 0.2pt,
  colsep = 0.2pt,
  colspec = {Q[133]Q[62]Q[62]Q[62]Q[62]Q[62]Q[62]Q[62]Q[62]Q[62]Q[62]Q[62]Q[62]Q[62]Q[62]Q[63]},
  cells = {c},
  hline{2} = {3,5,7,9,11,13,15}{r},
  hline{2} = {2,4,6,8,10,12,14,16}{l},
  hline{1,12} = {-}{0.1em},
  hline{3,11} = {-}{0.05em},
  cell{1}{1} = {r=2}{},
  cell{1}{2} = {c=2}{0.1\linewidth},
  cell{1}{4} = {c=2}{0.1\linewidth},
  cell{1}{6} = {c=2}{0.1\linewidth},
  cell{1}{8} = {c=2}{0.1\linewidth},
  cell{1}{10} = {c=2}{0.1\linewidth},
  cell{1}{12} = {c=2}{0.1\linewidth},
  cell{1}{14} = {c=2}{0.1\linewidth}, 
  cell{3}{3,5,7,9,11,13,15} = {bg=customgreen},
  cell{5}{13,15} = {bg=customgreen},
  cell{6}{9,11,13,15} = {bg=customgreen},
  cell{7}{9,13,15} = {bg=customgreen},
  cell{8}{3,5,7,9,11,13,15} = {bg=customgreen},
  cell{9}{9,13,15} = {bg=customgreen},
  cell{10}{7,9,11,13,15} = {bg=customgreen},
  cell{11}{11,13} = {bg=customgreen},
  % 红色背景 (ASR >= 20%)
  cell{4}{3,5,7,9,11,13,15} = {bg=customred},
  cell{5}{3,5,7,9,11} = {bg=customred},
  cell{6}{3,5,7} = {bg=customred},
  cell{7}{3,5,7,11} = {bg=customred},
  cell{9}{3,5,7,11} = {bg=customred},
  cell{10}{3,5} = {bg=customred},
  cell{11}{3,5,7,9,15} = {bg=customred},
}
Defense→   & DeepSweep \cite{qiu2021deepsweep}   &      & BNP \cite{zheng2022bnpep}   &      & MCR \cite{zhao2020mcr}   &      & NPD \cite{zhu2024npd}   &      & FT-SAM \cite{zhu2023ftsam}   &      & SAU \cite{wei2024sau}   &      & PGBD \cite{amula2025prototype}  &      & Avg. \\
Attack↓    & CA   & ASR  & CA   & ASR  & CA   & ASR  & CA   & ASR  & CA   & ASR  & CA   & ASR  & CA   & ASR  & ASR  \\
BadNet     & 85.3 & 1.9  & 93.2 & 15.2 & 90.5 & 2.0  & 91.1 & 0.9  & 92.8 & 1.7  & 90.6 & 2.2  & 90.7 & 0.7  & 3.5  \\
Blended    & 70.9 & 65.5 & 93.8 & 93.0 & 91.4 & 41.7 & 91.5 & 74.2 & 93.2 & 51.8 & 91.2 & \textbf{32.2} & 87.1 & \textbf{24.3} & 54.7 \\
SIG        & 84.5 & 41.6 & 93.7 & 80.6 & 90.9 & 31.8 & 91.3 & 63.6 & 92.9 & 49.5 & 85.8 & 0.8  & 86.8 & 0.5  & 38.3 \\
IA         & 87.6 & 75.9 & 90.6 & 79.2 & 93.4 & 80.7 & 85.5 & 2.6  & 93.4 & 5.4  & 91.2 & 2.8  & 89.6 & 2.6  & 35.6 \\
SSBA       & 71.8 & 81.2 & 93.3 & 99.7 & 90.8 & 39.3 & 91.2 & 8.8  & 92.8 & \textbf{60.3} & 86.7 & 2.6  & 88.4 & 5.2  & 42.4 \\
WaNet      & 92.6 & 4.7  & 55.5 & 12.8 & 93.4 & 1.7  & 90.9 & 0.9  & 93.5 & 0.9  & 90.9 & 0.6  & 88.7 & 2.4  & 3.4  \\
BPP        & 90.0 & 32.5 & 91.4 & 79.6 & 93.5 & \textbf{83.9} & 53.0 & 0.0  & 93.7 & 49.0 & 91.6 & 4.4  & 87.3 & 6.7  & 36.6 \\
FLIP       & 70.7 & 26.9 & 92.0 & 85.9 & 90.2 & 0.4  & 90.1 & 0.0  & 93.0 & 0.5  & 91.2 & 0.5  & 84.1 & 10.8 & 17.9 \\
GCB (ours) & 77.9 & \textbf{93.2} & 92.3 & \textbf{100.0}& 90.8 & 78.0 & 90.6 & \textbf{97.4} & 92.7 & 1.8  & 90.6 & 5.4  & 85.5 & 21.1 & \textbf{56.7} \\
\end{tblr}
\end{small}
\caption{Comparison of different attack methods against \textbf{clean-data-based} defenses on CIFAR-10 with 2500 (5\%) clean image-label pairs.}
\vskip -0.0in
\label{tab:defense_effectiveness_cifar10_clean_data_based}
\end{table*}

\subsection{\methodabb{}'s Resilience on CIFAR-100}
As shown in Table \ref{tab:defense_effectiveness_cifar100}, our method shows the highest effectiveness against all the testing defenses except DBD.
Among the defenses evaluated in this experiment, AC and ABL do not reduce ASR, whereas DBD substantially reduces it. A plausible explanation is that DBD's self-supervised stage does not use the poisoned labels, but the experiment does not establish that self-supervision is the only effective defense mechanism. The method of \citet{xu2025clip} is effective for all attacks tested here, while requiring an additional pretrained CLIP model~\citep{radford2021learning}.

\begin{table*}
\centering
\begin{small}
\begin{tblr}{
  width = 0.7\textwidth,
  rowsep = 0.2pt,
  colsep = 0.2pt,
  colspec = {Q[133]Q[62]Q[62]Q[62]Q[62]Q[62]Q[62]Q[62]Q[62]Q[62]Q[62]Q[62]Q[62]Q[62]Q[62]Q[63]},
  cells = {c},
  hline{2} = {3,5,7,9,11,13,15}{r},
  hline{2} = {2,4,6,8,10,12,14,16}{l},
  hline{1,12} = {-}{0.1em},
  hline{3,11} = {-}{0.05em},
  cell{1}{1} = {r=2}{},
  cell{1}{2} = {c=2}{0.09\linewidth},
  cell{1}{4} = {c=2}{0.09\linewidth},
  cell{1}{6} = {c=2}{0.09\linewidth},
  cell{1}{8} = {c=2}{0.09\linewidth},
  cell{1}{10} = {c=2}{0.09\linewidth},
  cell{1}{12} = {c=2}{0.09\linewidth},
  cell{1}{14} = {c=2}{0.09\linewidth},
  % ===== 原有配色保持不变（仅覆盖到列 13）；若需也高亮 CGD 的 ASR（列 15），可仿照添加 =====
  cell{3-8,11}{3} = {bg=customred},
  cell{9-10}{3} = {bg=customgreen},
  cell{3-8,10-11}{5} = {bg=customred},
  cell{9}{5} = {bg=customgreen},
  cell{6,11}{7} = {bg=customred},
  cell{3-5,7-10}{7} = {bg=customgreen},
  cell{4-5}{9} = {bg=customred},
  cell{3,6-11}{9} = {bg=customgreen},
  cell{3-5,7-8,11}{11} = {bg=customred},
  cell{6,9-10}{11} = {bg=customgreen},
  cell{3-5,7-8,11}{13} = {bg=customred},
  cell{6,9-10}{13} = {bg=customgreen},
}
Defense→   & AC \cite{chen2018activation}   &      & SS \cite{tran2018spectral}   &      & ABL \cite{li2021anti}  &      & DBD \cite{huang2022decouple}  &      & CLP \cite{zheng2022channel}  &      & EP \cite{zheng2022bnpep}   &      & CGD \cite{xu2025clip} &      & Avg. \\
Attack↓    & CA   & ASR  & CA   & ASR  & CA   & ASR  & CA   & ASR  & CA   & ASR  & CA   & ASR  & CA  & ASR  & ASR  \\
BadNets    & 60.4 & 36.5 & 66.6 & 42.2 & 46.7 & 0.8  & 61.0 & 0.2  & 61.8 & 23.5 & 66.7 & 24.2 & 69.4 & 0.0 & 18.2 \\
Blended    & 60.2 & 81.7 & 67.7 & 91.1 & 49.9 & 0.0  & 61.5 & \textbf{97.3} & 63.0 & 52.0 & 66.8 & 82.7 & 69.7 & 0.0 & 57.8 \\
SIG        & 61.0 & 72.1 & 65.8 & 71.3 & 51.4 & 0.0  & 62.4 & 92.2 & 65.9 & 81.3 & 67.8 & 78.3 & 70.1 & 0.4 & 56.5 \\
IA         & 60.6 & 63.3 & 67.0 & 69.1 & 61.1 & 63.3 & 61.7 & 0.1  & 64.2 & 1.1  & 62.7 & 0.6  & 70.3 & 0.0 & 28.2 \\
LF         & 60.8 & 35.5 & 66.4 & 45.6 & 61.5 & 4.6  & 60.9 & 0.4  & 69.0 & 29.0 & 66.6 & 47.0 & 70.0 & 0.2 & 23.2 \\
SSBA       & 61.1 & 94.5 & 67.1 & \textbf{98.5} & 50.9 & 0.0  & 62.0 & 0.4  & 69.9 & \textbf{99.2} & 68.6 & \textbf{98.9} & 69.9 & 0.4 & 56.0 \\
WaNet      & 59.9 & 4.5  & 66.7 & 10.0 & 56.8 & 4.7  & 63.3 & 0.2  & 62.2 & 1.2  & 61.8 & 16.0 & 70.2 & \textbf{0.5} & 5.3  \\
BPP        & 60.3 & 6.2  & 67.1 & 24.8 & 53.2 & 13.5 & 60.5 & 0.2  & 59.4 & 0.2  & 62.8 & 0.1  & 71.0 & 0.1 & 6.4  \\
GCB (ours) & 60.3 & \textbf{95.2} & 67.1 & 97.6 & 60.1 & \textbf{96.2} & 62.1 & 1.3  & 68.4 & 95.9 & 65.9 & 94.3 & 68.6 & 0.0 & \textbf{68.6}
\end{tblr}
\end{small}
\caption{Comparison of different attack methods against other defenses on CIFAR-100 (with CGD added).}
\vskip -0.0in
\label{tab:defense_effectiveness_cifar100}
\end{table*}

\subsection{\methodabb{}'s Resilience compared with FLIP on Noisy training mitigation.}
\methodabb{} demonstrates substantially higher resilience against noisy training mitigation techniques compared to FLIP-opt, as shown in Table~\ref{tab:flip_noisy}. While several noise-robust methods such as SPL and BootStrap significantly reduce FLIP-opt’s attack success rate (ASR), they fail to mitigate \methodabb{}’s effectiveness, which consistently maintains a near-100\% ASR across all defenses. Moreover, \methodabb{} preserves competitive CA, often outperforming FLIP-opt even under strong defenses. This indicates that \methodabb{}’s gradient-consistent backdoor mechanism is more robust to label noise and regularization-based filtering, allowing it to persist where traditional feature-aligned poisoning (e.g., FLIP-opt) is largely neutralized.

\section{Additional Ablation Study on Label Preservation}
\label{sec:label_condition}

\begin{figure}[ht]
\centering
\includegraphics[width=0.7\columnwidth]{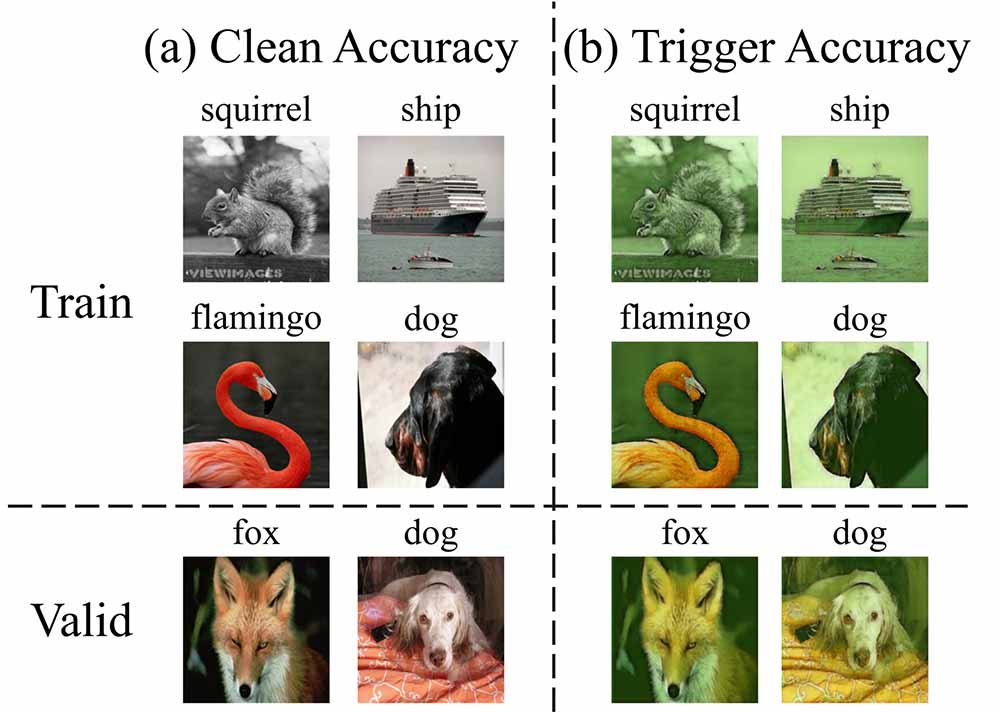}
\vskip -0.0in
\caption{Metric design for label preservation. A small gap between Triggered Accuracy (TA) and Clean Accuracy (CA) indicates that label-relevant information remains learnable after transformation. It is not a test of statistical independence.}
\label{fig:irrelavancy}
\vskip -0.00in
\end{figure}

In this section, we evaluate whether the trigger transformation preserves information needed for the benign classification task. This operational property is weaker than statistical independence between trigger and class features, which our experiment does not claim to establish.

\textbf{Verification of Label Preservation:}
We compare accuracy on clean images with accuracy on label-preserving transformed images. Let $h_0$ be a model trained and evaluated on clean data, and let $h_T$ be a model trained and evaluated on data transformed by $T$ while retaining the original labels. We report
\[
\mathrm{CA}=\Pr[h_0(X)=Y],
\qquad
\mathrm{TA}=\Pr[h_T(T(X,Y))=Y].
\]
A small CA--TA gap indicates that the transformation retains label information that is learnable by the evaluated architecture. Because the models are trained separately, this metric does not establish classifier invariance or independence between the trigger and class-discriminative features.

\textbf{Experimental Design:}

\begin{enumerate}
    \item \textbf{Clean Accuracy (CA):} Accuracy of $h_0$ on clean test inputs.
    \item \textbf{Triggered Accuracy (TA):} Accuracy of $h_T$ on transformed test inputs with their original labels.
\end{enumerate}

We transformed all images in both the training and testing datasets while preserving the original labels. Models with the same architecture as those used in the benign setting were trained, and each experiment was replicated five times to assess run-to-run variability.

\begin{table}[ht]
    \centering
    \begin{small}
    \begin{tabular}{lcc}
        \toprule
        & \textbf{CIFAR-10} & \textbf{CIFAR-100} \\
        \midrule
        \textbf{CA} & $93.9\%$ & $71.0\%$ \\
        \textbf{TA without $LC$} & $14.3\%$ & $3.8\%$ \\
        \textbf{TA (Our Method)} & $91.2\%$ & $67.4\%$ \\
        \bottomrule
    \end{tabular}
    \end{small}
        \caption{Performance on CIFAR-10 and CIFAR-100. TA drops substantially without \textit{LC}, indicating that the corresponding transformation removes label-relevant information under this evaluation and is consistent with possible mode collapse.}
    \vskip -0.0in
    \label{tab:irrelevance_results}
\end{table}

The results in Table~\ref{tab:irrelevance_results} show that \textbf{without label conditioning (LC)}, TA decreases substantially, suggesting loss of label-relevant information. With label conditioning, TA remains closer to CA, providing empirical evidence that the transformation better preserves label information. These results do not prove statistical independence between trigger and class features.

\begin{figure}[ht]
\begin{center}
\centerline{\includegraphics[width=\columnwidth]{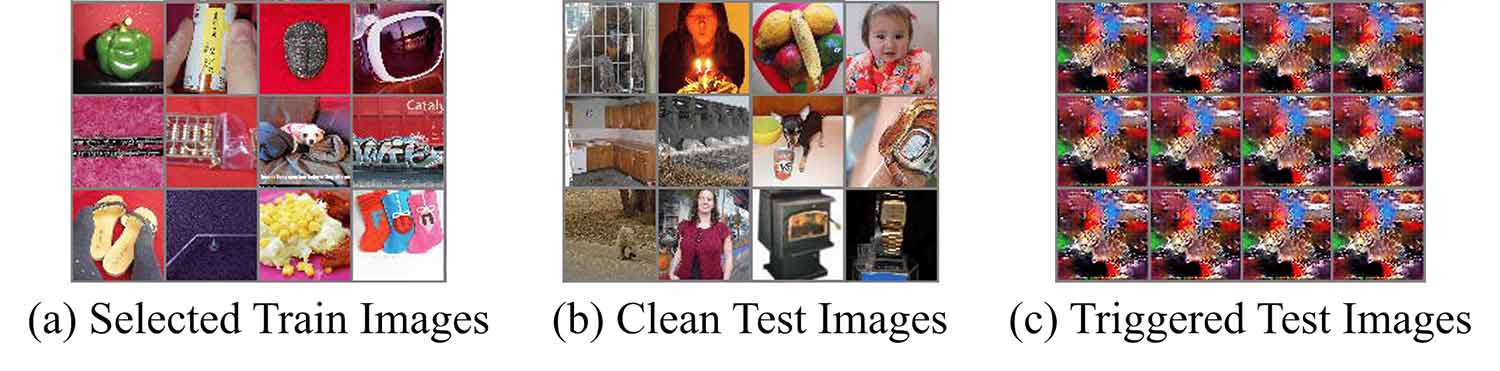}}
\vskip -0.0in
\caption{One example of \methodabb{} without Label Condition (LC). The triggered outputs collapse toward one pattern. This variant can still achieve high ASR but has low triggered accuracy on the original labels.}
\label{fig:collapse}
\end{center}
\vskip -0.0in
\end{figure}

Upon further examination of images without Label Conditioning (LC), we observed outputs concentrated around patterns resembling a single class, consistent with mode collapse in Generative Adversarial Networks. This behavior provides a plausible explanation for the low TA in Fig.~\ref{fig:collapse}. The collapsed outputs can still yield high ASR, but they preserve substantially less information about the original labels.

\section{\methodabb{}'s Resilience to Abnormal Sample Detection.}

\begin{figure*}[t]
\centering

\begin{minipage}[t]{0.18\textwidth}
  \centering
  \includegraphics[width=1.0\textwidth]{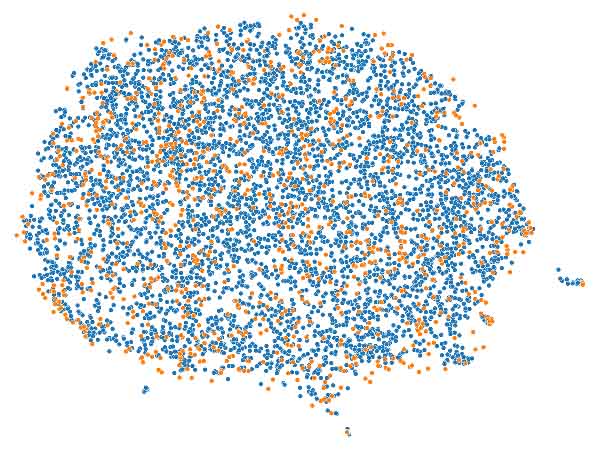}
  \caption*{(a) No Poison}
  \label{fig:clean}
\end{minipage}
\hfill
\begin{minipage}[t]{0.18\textwidth}
  \centering
  \includegraphics[width=1.0\textwidth]{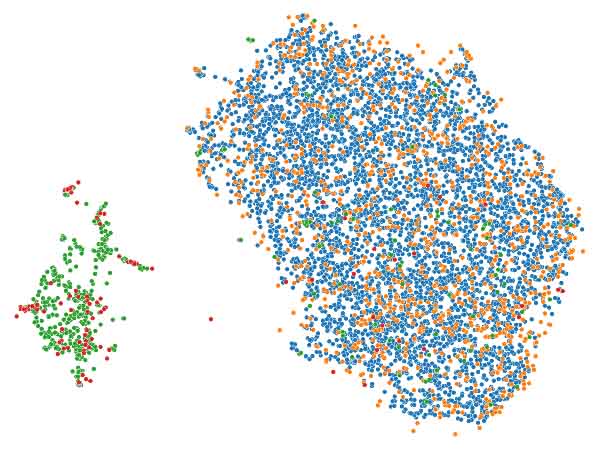}
  \caption*{(b) BadNets}
  \label{fig:bad}
\end{minipage}
\hfill
\begin{minipage}[t]{0.18\textwidth}
  \centering
  \includegraphics[width=1.0\textwidth]{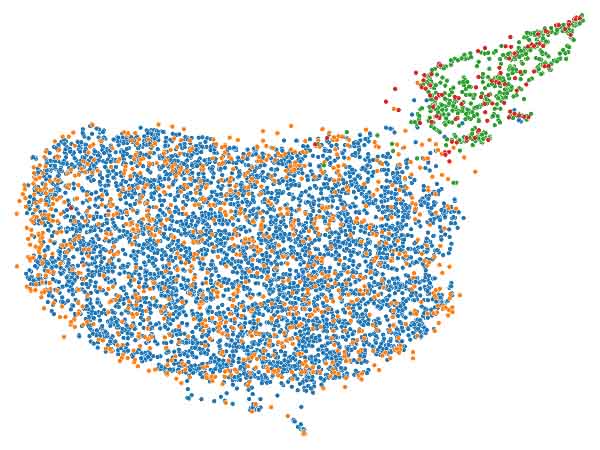}
  \caption*{(c) Blend}
  \label{fig:blend}
\end{minipage}
\hfill
\begin{minipage}[t]{0.18\textwidth}
  \centering
  \includegraphics[width=1.0\textwidth]{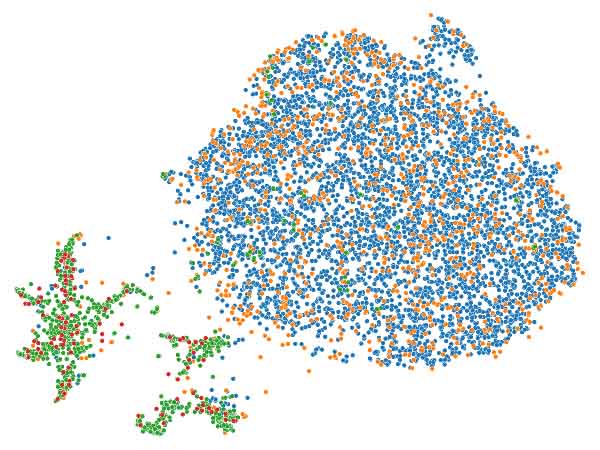}
  \caption*{(d) Input-Aware}
  \label{fig:ia}
\end{minipage}
\hfill
\begin{minipage}[t]{0.18\textwidth}
  \centering
  \includegraphics[width=1.0\textwidth]{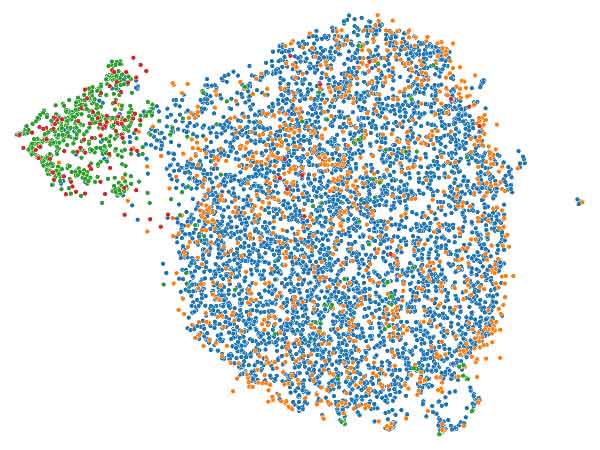}
  \caption*{(e) Low-Frequency}
  \label{fig:lf}
\end{minipage}
\hfill
\begin{minipage}[t]{0.18\textwidth}
  \centering
  \includegraphics[width=1.0\textwidth]{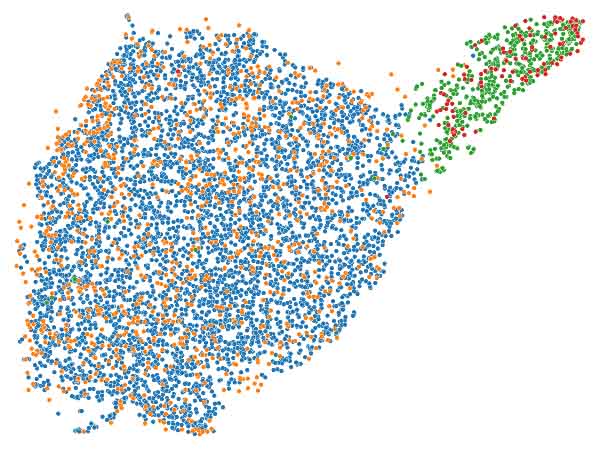}
  \caption*{(f) SSBA}
  \label{fig:ssba}
\end{minipage}
\hfill
\begin{minipage}[t]{0.18\textwidth}
  \centering
  \includegraphics[width=1.0\textwidth]{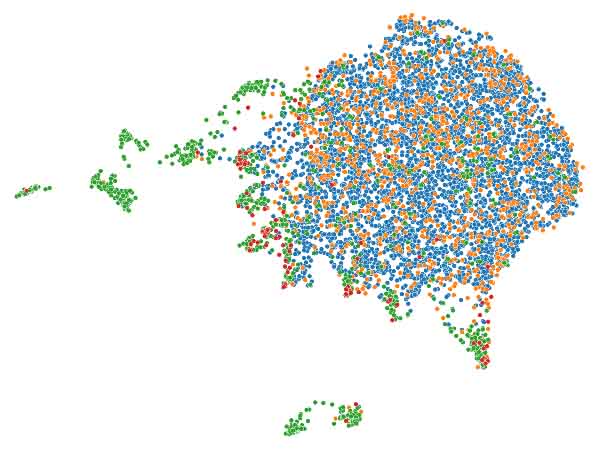}
  \caption*{(g) WaNet}
  \label{fig:wanet}
\end{minipage}
\hfill
\begin{minipage}[t]{0.18\textwidth}
  \centering
  \includegraphics[width=1.0\textwidth]{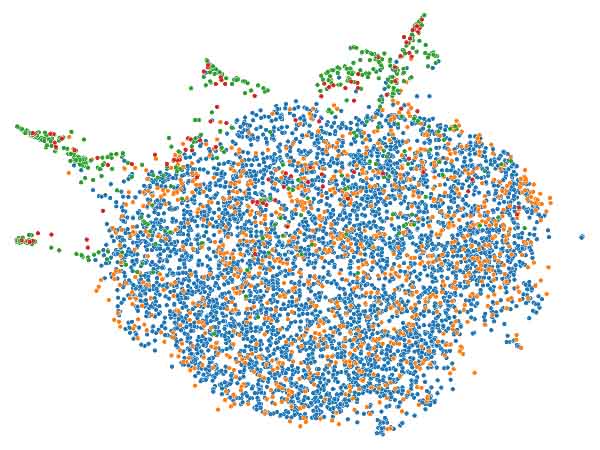}
  \caption*{(h) FLIP}
  \label{fig:flip}
\end{minipage}
\hfill
\begin{minipage}[t]{0.18\textwidth}
  \centering
  \includegraphics[width=1.0\textwidth]{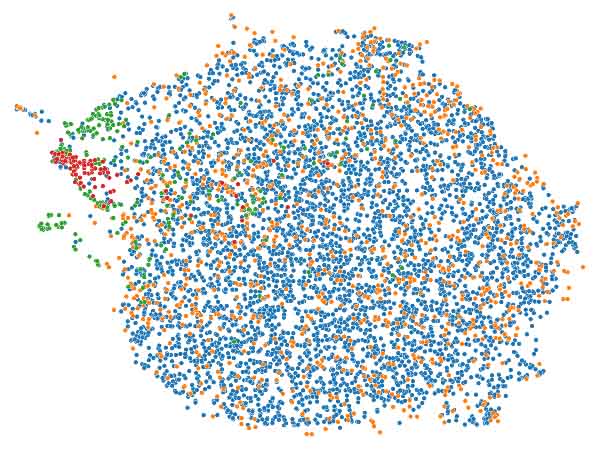}
  \caption*{(i) \textbf{\methodabb{} (Ours)}}
  \label{fig:ours}
\end{minipage}
\hfill
\begin{minipage}[t]{0.18\textwidth}
  \centering
  \includegraphics[width=1.0\textwidth]{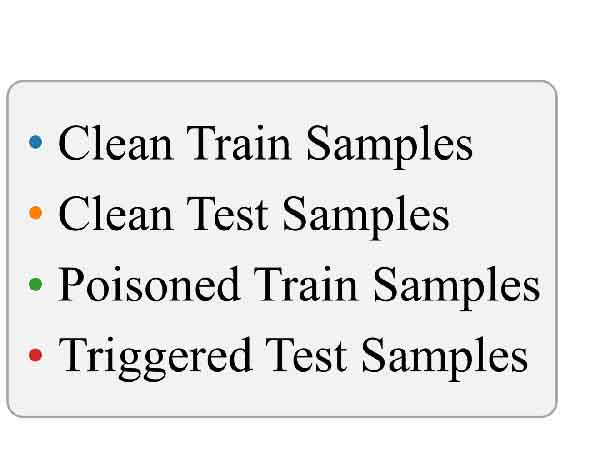}
  \caption*{(j) Legend}
  \label{fig:legends}
\end{minipage}

\vskip -0.00in
\caption{UMAP Visualization of different backdoor attack methods in the CIFAR-10 dataset.}
\label{fig:tsne_method}
\end{figure*}

\begin{figure*}[t]
\centering

\begin{minipage}[t]{0.18\textwidth}
  \centering
  \includegraphics[width=1.0\textwidth]{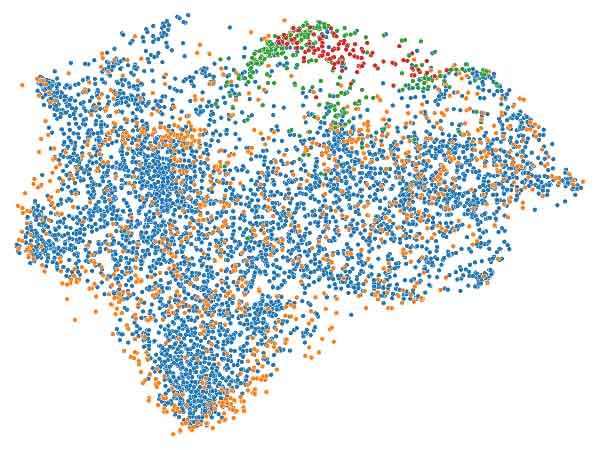}
  \caption*{(a) Layer 1}
  \label{fig:layer1}
\end{minipage}
\hfill
\begin{minipage}[t]{0.18\textwidth}
  \centering
  \includegraphics[width=1.0\textwidth]{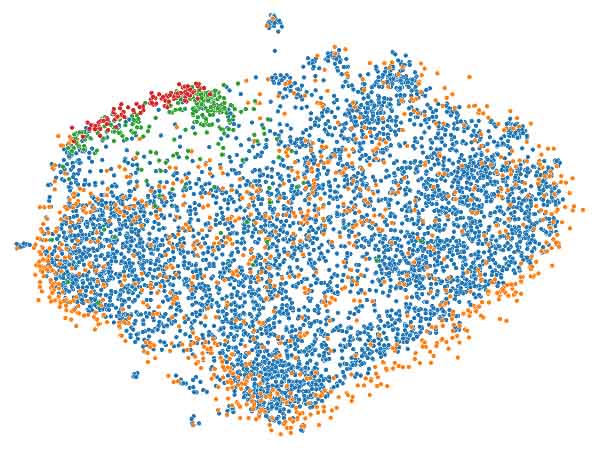}
  \caption*{(b) Layer 2}
  \label{fig:layer2}
\end{minipage}
\hfill
\begin{minipage}[t]{0.18\textwidth}
  \centering
  \includegraphics[width=1.0\textwidth]{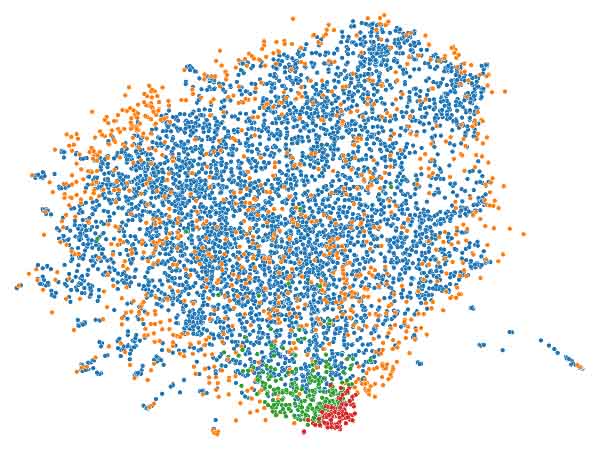}
  \caption*{(c) Layer 3}
  \label{fig:layer3}
\end{minipage}
\hfill
\begin{minipage}[t]{0.18\textwidth}
  \centering
  \includegraphics[width=1.0\textwidth]{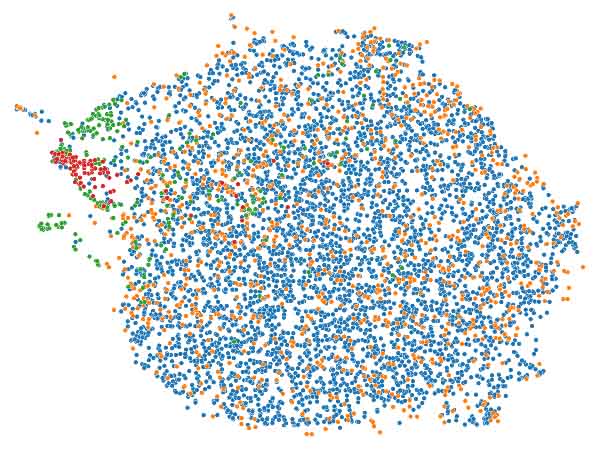}
  \caption*{(d) Layer 4}
  \label{fig:layer4}
\end{minipage}

\vskip -0.00in
\caption{UMAP Visualization of different layers on PreActResNet in \methodabb{}.}
\label{fig:tsne_layer}
\end{figure*}

Since the trigger of \methodabb{} is GAN-generated, there may be a significant distributional gap between the generated trigger and real image data. Consequently, it is essential to analyze the method’s resistance to defense strategies based on abnormal sample detection. Specifically, we employ Uniform Manifold Approximation and Projection (UMAP) to visualize the distribution of intermediate features in the victim model, a standard approach in backdoor detection research~\citep{qi2022revisiting, wu2024backdoorbench}. We investigate two key aspects:

\subsection{Detectability of Poisoned Training Samples}

We assess whether the selected, relabeled training samples can be detected as outliers when compared with the remaining clean samples. The training images are not generated or modified; only their labels are changed. In the UMAP projection in Figure~\ref{fig:tsne_method}, the selected samples are not visibly separated from the remaining samples. This is qualitative evidence about the displayed projection, not proof that the underlying feature distributions are identical.

One possible explanation is that the score selects samples according to features already present in unmodified training images. Under the displayed UMAP configuration, simple visual separation is therefore difficult. Similar qualitative observations for natural-trigger backdoors have been reported in previous studies~\citep{qi2022revisiting}.

\subsection{Detectability of Triggered Test Samples}

We also evaluate whether triggered test samples appear as outliers during inference. Figure~\ref{fig:tsne_method} shows that their two-dimensional UMAP projections overlap substantially with those of clean samples at the examined layers. This observation is specific to the tested victim model, layers, and projection settings.

The observed overlap is consistent with the adversarial objective encouraging in-distribution outputs. It suggests that the evaluated projection-based inspection may have limited discriminative power in this setting, but it does not guarantee evasion of general anomaly detectors.

\subsection{Impact of Network Layers on UMAP Visualization}

We further explore the impact of different network layers on the UMAP visualization of our method. As shown in Figure~\ref{fig:tsne_layer}, we visualize feature projections at Layer1--Layer4 of PreActResNet-18. Across these projections, selected training samples and triggered test samples show substantial visual overlap with clean samples. This supports the narrower claim that simple inspection of these particular projections does not reveal a clear separation.

\section{Hyperparameter Analysis}

To validate the hyperparameter sensitivity of our method, we conducted experiments on two key parameters: the learning rate and the weight factor of the information loss $\lambda$. These two terms are also considered crucial in the original InfoGAN paper~\citep{chen2016infogan}. We evaluated the training outcomes based on three aspects: (1) ASR, (2) visualization of triggered test images, and (3) visualization of selected training images.

\paragraph{(a) Effect of Learning Rate}
We tested five different learning rates: $1 \times 10^{-5}$, $3 \times 10^{-5}$, $1 \times 10^{-4}$, $3 \times 10^{-4}$, and $1 \times 10^{-3}$. We found that the ASR remained high (over 90\%) across all learning rates. However, when the learning rate was very low or very high ($1 \times 10^{-5}$ or $1 \times 10^{-3}$), strong artifacts were observed in the triggered test images, making these samples easier to detect at test time. Interestingly, we also found that different learning rates sometimes converged to different trigger patterns. At a learning rate of $3 \times 10^{-4}$, the trigger became a frame around the image, while other learning rates resulted in triggers with special colors. This finding—that different learning rates result in different patterns—was also observed in the original InfoGAN~\citep{chen2016infogan}.

\begin{figure}[ht]
\centering
\includegraphics[width=1.0\linewidth]{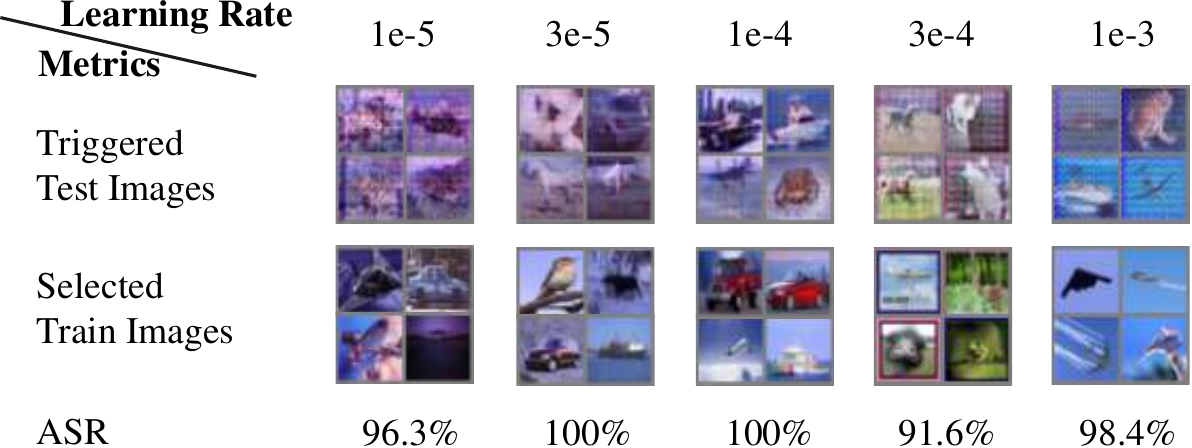}
\caption{Effect of learning rate on the trigger patterns and artifacts in the generated images. Each column corresponds to a different learning rate.}
\label{fig:learning_rate}
\end{figure}

\paragraph{(b) Effect of Weight Factor $\lambda$}
We tested five values for the information-loss weight $\lambda$: 0.05, 0.1, 0.25, 0.5, and 1.0. ASR drops substantially at lower weights (0.05 and 0.1), consistent with insufficient emphasis on code predictability. At high weights, the joint generator/recognition objective places greater emphasis on making $c$ predictable relative to the adversarial realism objective; empirically, this setting produces more noticeable artifacts and a larger measured gap between real and generated images. This interpretation is observational rather than a convergence guarantee.

\begin{figure}[ht]
\centering
\includegraphics[width=1.0\linewidth]{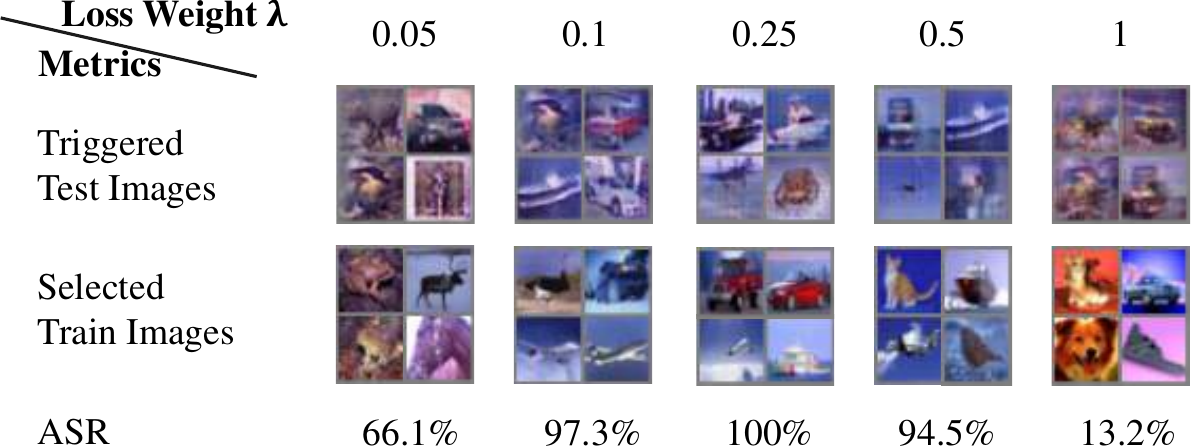}
\caption{Effect of the information loss weight factor $\lambda$ on ASR and image quality. Each column corresponds to a different value of $\lambda$.}
\label{fig:loss_weight}
\end{figure}

In conclusion, both the learning rate and the weight factor $\lambda$ are robust within a certain range. However, when these parameters become too high or too low, their effects differ. The learning rate affects the amount of artifacts in the generated images but does not significantly impact the ASR. On the other hand, the weight factor $\lambda$ has a large impact on the ASR.

\end{document}